\documentclass[review]{elsarticle}
\graphicspath{ {./figures/} }
\usepackage{hyperref}
\usepackage{float}
\usepackage{verbatim} 
\usepackage{apalike}
\usepackage{natbib}
\usepackage{xcolor}
\usepackage{graphicx}  
\usepackage{multirow}  
\usepackage{lscape}    
\usepackage{array}     
\usepackage{amsmath}       
\usepackage{amssymb}       
\usepackage{amsfonts}      
\usepackage[linesnumbered,ruled,vlined]{algorithm2e} 
\usepackage[numbers]{natbib} 
\usepackage{multirow} 
\usepackage[table,xcdraw]{xcolor}
\usepackage{xcolor}        
\usepackage{adjustbox} 
\usepackage[ruled,vlined]{algorithm2e}  
\restylefloat{figure}
\floatstyle{plaintop} 
\restylefloat{table}

\journal{Expert Systems with Applications}

\biboptions{authoryear}

\begin{document}
\begin{frontmatter}


\begin{titlepage}
\begin{center}
\vspace*{1cm}

\textbf{Taylor-Series Expanded Kolmogorov-Arnold Network for Medical Imaging Classification}

\vspace{1.5cm}

Kaniz Fatema$^{a}$ (fate2180@mylaurier.ca), Emad A. Mohammed$^a$ (emohammed@wlu.ca), Sukhjit Singh Sehra$^a$ (ssehra@wlu.ca) \\

\hspace{10pt}

\begin{flushleft}
\small  
$^a$ Department of Physics and Computer Science, \\ 
Wilfrid Laurier University, Waterloo, Canada

\vspace{1cm}
\textbf{Corresponding author at:} \\ 
Kaniz Fatema \\ 
Department of Physics and Computer Science, \\ 
Wilfrid Laurier University, Waterloo, Canada \\ 
Email: fate2180@mylaurier.ca

\end{flushleft}        
\end{center}
\end{titlepage}

\title{Taylor-Series Expanded Kolmogorov-Arnold Network for Medical Imaging Classification}

\author[label1]{Kaniz Fatema}
\ead{fate2180@mylaurier.ca}

\author[label1]{Emad A. Mohammed}
\ead{emohammed@wlu.ca}

\author[label1]{Sukhjit Singh Sehra}
\ead{ssehra@wlu.ca}

\address[label1]{Department of Physics and Computer Science, Wilfrid Laurier University, Waterloo, Canada}

\begin{abstract}
Effective and interpretable classification of medical images remains a key challenge in computer-aided diagnosis, particularly in data-scarce and resource-constrained clinical settings. This study introduces spline-based three Kolmogorov–Arnold Networks (KANs) for accurate medical image classification on limited and heterogeneous datasets: SBTAYLOR-KAN integrates B-splines with Taylor series, SBRBF-KAN combines B-splines with Radial Basis Functions, and SBWAVELET-KAN embeds B-splines within Morlet wavelet transforms. These architectures leverage spline-driven function approximation to efficiently capture both local and global nonlinearities.

The models were evaluated on diverse datasets, including brain MRI, chest X-rays, tuberculosis (TB) X-rays, and skin lesion images, without any image preprocessing application, demonstrating the ability to learn directly from unprocessed raw data. Comprehensive experiments with cross-dataset validation, robustness checks, and data reduction analysis confirmed strong generalization and stability. SBTAYLOR-KAN achieved an accuracy of up to 98.93\%, demonstrated a well-balanced F1-score, and maintained over 86\% accuracy using only 30\% of the training data across three datasets. Despite the significant class imbalance in the skin cancer dataset, we generated a balanced version of the data and conducted experiments with both the imbalanced and balanced datasets. In these experiments, SBTAYLOR-KAN outperformed the other models, achieving an accuracy of 68.22\% in classification.

Unlike conventional CNNs that demand tens to hundreds of millions of parameters—for instance, ResNet50 ($\sim$24.18M) and VGG16 ($\sim$14.78M) etc. SBTAYLOR-KAN achieves comparable functionality with just 2,872 trainable parameters, enabling efficient deployment in constrained medical environments. To improve interpretability, Gradient-weighted Class Activation Mapping (Grad-CAM) was employed to generate visual explanations of class-relevant regions in medical images. Overall, the proposed framework offers a lightweight, interpretable, and generalizable solution for medical image classification. It specifically addresses the challenges of limited datasets in the medical field and data-scarce scenarios, enabling real-world clinical AI integration.
The source code for all proposed models is publicly available at \citep{brishty2025taylor}.

\end{abstract}


\end{frontmatter}

\section{Introduction}
\label{introduction}
Artificial intelligence (AI) has emerged as a transformative force in medical imaging, enabling advancements in diagnostic accuracy, automated image segmentation, and disease classification. These improvements have the potential to reduce clinician workload and improve patient outcomes. However, most high-performing AI models, particularly in the field of deep learning, are heavily reliant on large-scale labelled datasets and substantial computational resources \citep{Siva2024}. This dependence poses a significant barrier in clinical environments, where data acquisition is constrained by privacy regulations, ethical concerns, and the considerable time and cost required for expert annotation \citep{shickel2017deep,chexpert2019}.

In many medical applications, especially those involving rare diseases, working with limited datasets is a common challenge \citep{Li2020}. These datasets not only tend to be small but are also characterized by high heterogeneity, class imbalance, and the need for precise labelling \citep{Siva2024}, often by radiologists or other domain experts. Despite their limited size, such datasets frequently contain crucial clinical patterns essential for accurate diagnosis. This underscores the need for models that can effectively learn from limited data, while still being interpretable and robust. Large over-parameterized models often fall short in this context, as they tend to overfit and are challenging to interpret in clinical decision-making \citep{Li2020}. 

While deep learning models have shown strong performance in many areas, they present significant challenges when applied to medical data. These include a high risk of overfitting in data-limited environments, difficulty interpreting results, and high computational costs \citep{rice2020overfitting,samek2017explainable}. In healthcare settings with limited resources, like rural hospitals or on-site treatment centers, these challenges make it harder to rely on AI models for real-time diagnostics \citep{iqbal2023ldmres}. Additionally, these models often struggle to generalize when trained on data that lacks diversity, a common issue in small, highly imbalanced medical datasets. In addition, deep learning models trained on small datasets also face difficulty in capturing the complex, multi-scale features needed for accurate disease detection, especially in complicated medical images \citep{najafabadi2015deep}. When trained on limited data, these models are prone to overfitting, learning patterns that don't reflect real spatial features \citep{sufi2024addressing}. Furthermore, with rare disease diagnoses, insufficient data can prevent deep learning models from understanding critical spatial characteristics, leading to misclassifications or low accuracy \citep{Salehi2023}.

Thus, obtaining optimal accuracy by utilizing a small dataset becomes the most significant limitation in the medical field for deep learning models. In response to these challenges, the research community is increasingly exploring lightweight, data-efficient models that require fewer parameters while retaining the ability to capture essential features. This hypothesis overcomes limitations by leveraging the Kolmogorov-Arnold theorem within deep learning frameworks, forming Kolmogorov-Arnold Networks (KANs). These networks provide a systematic method for constructing neural architectures capable of effectively approximating multivariate functions through combinations of univariate functions \citep{KAN2024,KARTtheory3}. This architectural innovation has shown the potential to enhance the accuracy and convergence of neural networks \citep{XuKAN2024}, particularly in applications that involve complex mathematical operations and scientific computation \citep{KAGNN2024}. 
Its novel design uses learnable univariate functions instead of linear weights. Traditional neural networks utilize fixed activation functions at nodes, while KAN uses dynamically changeable activation functions on network edges. Each weight parameter is represented by a univariate function, commonly parameterized using spline functions. This architecture increases the network’s flexibility, allowing it to better react to changing data patterns \citep{Guo2025KAN}. Moreover,
traditional neural networks generally optimize parameters using set weights and biases. KAN broadens the area of optimization by including learnable activation functions. This means that the parameters of the activation functions are altered throughout the optimization process, in addition to the network’s weights and biases. Although this makes training more complicated, it also gives KAN more flexibility, which enables it to capture complex data correlations more successfully \citep{KARTtheory3}. Additionally, traditional neural networks frequently necessitate a substantial number of parameters and computational resources, particularly in the case of large CNNs. This is a significant constraint of environments that are resource-constrained. In contrast, by applying learnable univariate functions, KAN decreases parameters dramatically. For instance, convolutional KANs (Conv-KAN) reduce computational expenses by achieving good accuracy with fewer parameters than CNNs \citep{KARTtheory3}. This decrease boosts computational efficiency and scalability, improving the model for resource-constrained contexts \citep{Guo2025KAN}. Moreover, KAN provides distinct benefits in managing high-dimensional data. It employs learnable univariate functions rather than static weights, allowing it more adaptability to the intricate structures of high-dimensional data. This feature makes KAN particularly effective in processing large, high-dimensional data \citep{KANsurvey}.

This study investigates KANs as an efficient learning architecture that can be optimally configured to extract clinically relevant patterns from limited medical imaging datasets. The primary objective is to identify optimal model configurations that balance parameter efficiency, accurate spatial feature representation, interpretability, and generalization to medical data. By exploring these configurations, we aim to build scalable, data-efficient AI models that reliably extract clinically meaningful patterns under limited data. Such models support robust, practical deployment in resource-constrained healthcare environments, bridging the gap between advanced machine learning capabilities and real-world clinical applicability.
The key contributions of our approach are outlined below.
\begin{enumerate}
    \item Multiple diverse medical imaging datasets, including Brain Tumor MRI, COVID-19 X-rays, TB X-rays, Skin Cancer, and MedMNIST, were used without any image pre-processing or enhancement, allowing the models to learn directly from unprocessed raw image data.
    
    \item Three novel spline-based KAN models were developed to classify different medical images: SBTAYLOR-KAN, combining B-splines with Taylor series to capture global and local nonlinearities; SBRBF-KAN, integrating Radial Basis Functions (RBF) with B-splines for localized nonlinear encoding; and SBWAVELET-KAN, fusing Wavelet transforms with B-splines to model both oscillatory and smooth spatial patterns. 

    \item Extensive experiments were conducted, including cross-dataset validation, evaluation on unseen datasets, robustness testing under data reduction, and statistical analyses using the kappa coefficient (KC), the Matthews correlation coefficient (MCC), error rate, and p-values.

    \item Explainable AI (XAI) was incorporated using Gradient-weighted Class Activation Mapping (Grad-CAM) to generate visual heatmaps, highlighting the critical regions that influenced the model’s classification decisions.

    \item The proposed models are lightweight and parameter-efficient, delivering high accuracy with a minimal number of trainable parameters, making them well-suited for deployment in resource-constrained medical environments.
\end{enumerate}

\section{Related Work}
\label{related_work}
In this section, we reviewed previous studies that used similar datasets for image classification with deep learning models. Notably, studies that applied KAN models to these specific dataset types are limited. Consequently, our review encompasses all classification tasks in which KAN models have been utilized. Our work aims to address this research gap, and a comprehensive description of our study is presented below.

\subsection{Deep Learning in Image Classification}
 \citep{mathivanan2024employing} employed a deep transfer learning approach on the Kaggle Brain Tumor MRI dataset (7,023 images; 5,618 training and 1,405 testing). After preprocessing and augmentation, four transfer learning architectures—ResNet152, VGG19, DenseNet169, and MobileNetv3—were fine‑tuned with an additional fully connected layer for four‑class classification. MobileNetv3 achieved the highest accuracy (99.75\%) and strong F1‑score (0.96) while remaining relatively lightweight, with an estimated 5.4M parameters and 53 effective layers. \citep{disci2025advanced} used the same dataset (7,023 images; 5,712 training and 1,311 testing) with six pre‑trained models—Xception, MobileNetV2, InceptionV3, ResNet50, VGG16, and DenseNet121. Following grayscale conversion, ROI cropping, resizing to 128×128, and augmentation (brightness/contrast, normalization), the models were fine‑tuned for four‑class classification. Xception achieved the best accuracy (95.27\%) and F1‑score (95.29\%), while MobileNetV2, with an estimated 3.4M parameters, also performed competitively.  \citep{amarnath2024transfer} evaluated five pre‑trained CNNs—ResNet50, Xception, EfficientNetV2‑S, ResNet152V2, and VGG16—on the same MRI dataset. These models had estimated depths ranging from 16 to 152 layers, with parameter counts from approximately 14M (VGG16) to 60M (ResNet152V2). Xception achieved the highest performance (98.17\% accuracy; 0.9817 F1‑score), followed by EfficientNetV2‑S (96.19\%; 0.9629 F1‑score). In the meantime, \citep{jeyapriya2025advanced} proposed a custom CNN with four convolutional layers, three MaxPooling2D layers, one 512‑unit dense layer, and a softmax layer, comprising an estimated 496K parameters. With augmentation and Adam optimization, the model reached 99.4\% accuracy and a 0.9951 weighted F1‑score, demonstrating the reliability of CNNs with proper preprocessing.  In another study, \citep{prasad2024mri} fine‑tuned EfficientNetB3 on a hybrid MRI dataset of 6,328 preprocessed and augmented images from two Kaggle sources. EfficientNetB3, featuring an estimated 26 convolutional blocks, additional dense layers, and around 12M parameters, achieved 98.13\% accuracy with a macro F1‑score of ~0.98 in four‑class tumor classification.
To classify COVID‑19, pneumonia, and normal chest X‑ray images, ~\citep{tengjongdee2024comparative} implemented DenseNet201, a deep learning CNN model. The dataset comprised 10,530 augmented images from the Kaggle Chest X‑ray datasets, evenly distributed across three classes. DenseNet201, a 201‑layer architecture with approximately 20 million parameters, achieved the highest performance with an accuracy of 94.75\% and an F1‑score of 97.57\%. Grad‑CAM visualization was leveraged to enhance interpretability by identifying pathological lung regions. DenseNet201 outperformed ResNet and MobileNet variants across all evaluation metrics.

Similarly, ~\citep{shin2024enhancing} developed a lightweight CNN to classify COVID‑19, pneumonia, and normal chest X‑ray images using 6,432 Kaggle images (5,147 for training) across three categories. Their 141,827‑parameter model, composed of four convolutional and two max‑pooling layers, was optimized for memory‑constrained deployment. Preprocessing techniques including basic, HOG, and CLAHE yielded comparable results, with peak performance reaching 95.06\% accuracy and an F1‑score of approximately 0.95, highlighting the efficiency of the model and the influence of preprocessing in early lung disease detection.~\citep{chakraborty2022transfer} employed VGG‑19, a deep transfer learning architecture, to classify COVID‑19, pneumonia, and healthy chest X‑ray images. The dataset included 3,797 images (2,733 for training) from the Kaggle COVID‑19 Radiography Database. VGG‑19, with 19 layers and roughly 20.1 million parameters, utilized pre‑trained convolutional blocks as frozen feature extractors with a trainable dense output layer. The model achieved 97.11\% accuracy and a macro F1‑score of 0.97, demonstrating strong effectiveness in lung pathology detection. In addition, ~\citep{rahman2021exploring} also investigated CheXNet, a DenseNet‑121‑based model with approximately 8.6 million parameters and 121 layers, for COVID‑19 detection from chest X‑ray images in the COVID dataset (18,479 images: 8,851 normal, 6,012 non‑COVID lung opacity, 3,616 COVID‑19). Fine‑tuned on gamma‑corrected images, CheXNet achieved 96.29\% accuracy and an F1‑score of 0.9628. The study emphasized the role of image enhancement techniques (HE, CLAHE, Complement, Gamma, BCET), identifying gamma correction as the most effective for improving COVID‑19 detection performance.

 For classifying normal and TB cases from chest X‑ray images, ~\citep{haque2025optimized} proposed a VGG‑19 transfer learning model leveraging the Tuberculosis Chest X‑ray Database (Kaggle), which contains 4,200 images comprising 3,500 normal and 700 TB‑positive cases. The model architecture consists of 19 layers with an estimated 20 million parameters, integrating 2D mean pooling, global average pooling, batch normalization, and dropout layers to enhance feature extraction and reduce overfitting. After hyperparameter tuning (tanh activation, SGD optimizer, 0.3 dropout), the model achieved 98\% classification accuracy. ~\citep{mirugwe2025improving} subsequently evaluated multiple CNN architectures—VGG16, VGG19, ResNet50, ResNet101, ResNet152, and Inception‑ResNet‑V2—on the same dataset using transfer learning. Their models incorporated additional dense, dropout, and softmax layers, and were trained on approximately 3,360 images. Among all tested architectures, VGG16 achieved the highest performance, reaching 99.4\% accuracy and an F1‑score of 98.3\%, despite its moderate 16‑layer design and approximately 14.7 million parameters. Similarly, ~\citep{wajgi2024optimized} focused on VGG19 to develop an optimized TB classification model using transfer learning and hyperparameter tuning. This 19-layer network with 20,157,250 parameters (\(\approx 20 \, \text{million}\)) integrated average and global average pooling, batch normalization, dropout, and dense layers for robust feature extraction.
 Their model achieved 98.11\% accuracy and an F1‑score of 0.9808. Moreover, ~\citep{jain2024tuberculosis} explored a deeper ResNet50‑based architecture consisting of 50 layers and 90,844,994 parameters (\(\approx 90.8 \, \text{million}\)), fine‑tuned with additional dense layers for binary classification. While this model demonstrated strong recall for TB detection, its overall performance was comparatively lower, with an accuracy of 91\% and an F1‑score of 0.857.


For cancer disease classification, \citep{uliana2025diffusion} implemented a Diffusion Model (DiffMIC) to classify both skin and oral cancers. The study utilized the PAD‑UFES‑20 and P‑NDB‑UFES datasets, comprising 3,763 histopathology image patches of oral lesions categorized into three classes—OSCC, leukoplakia with dysplasia, and leukoplakia without dysplasia—and employed K‑Fold cross‑validation for evaluation. The DiffMIC model uses a UNet backbone with ResNet18 encoders, resulting in unknown total parameters (not directly reported, standard diffusion + UNet architectures vary) and a multistage diffusion pipeline for classification. It achieved 64.57\% balanced accuracy (BACC) with 0.7153 F1‑score on PAD‑UFES‑20 (6‑class) and 90.50\% BACC with 0.9144 F1‑score on P‑NDB‑UFES. In another study, researchers \citep{azeem2023skinlesnet} proposed SkinLesNet, a multi‑layer deep convolutional neural network (CNN) for classifying skin lesions and detecting melanoma using the modified PAD‑UFES‑20 Modified dataset, which consists of 1,314 smartphone images of three classes (seborrheic keratosis, nevus, melanoma). SkinLesNet is a 4‑layer CNN model with additional dense, dropout, and softmax layers. Although the exact parameter count is not provided, it is a custom CNN (estimated $<1$ million parameters) based on its shallow architecture. The model achieved 96\% accuracy and an F1‑score of 92\%, significantly outperforming ResNet50 (82\%) and VGG16 (79\%) on the same dataset. \citep{yin2022study} also developed a DenseNet‑169‑based MD‑Net model for skin tumor classification by fusing dermoscopic images with patient metadata. The study used two datasets: PAD‑UFES‑20 and ISIC 2019. The DenseNet‑169 model has 169 layers, fine‑tuned with MetaNet and MetaBlock modules to enhance feature extraction from images and metadata. The model achieved 81.4\% balanced accuracy (BACC) on PAD‑UFES‑20 and 85.6\% BACC on ISIC 2019, demonstrating high classification performance across multiple skin tumor classes.
 
\subsection{KAN in Image Classification}
\citep{yang2025medkan} proposed MedKAN, KAN-based framework for multi-modal medical image classification. The study employed nine public MedMNIST datasets: BloodMNIST, BreastMNIST (780, 2 classes), DermaMNIST, OCTMNIST, PneumoniaMNIST, TissueMNIST, OrganAMNIST, OrganCMNIST, and OrganSMNIST. All datasets were pre-split for training, validation, and testing according to MedMNISTv2 standards, with images resized to 224$\times$224 pixels. 
Among the three variants (MedKAN-S, MedKAN-B, MedKAN-L), MedKAN-B achieved the best performance, with approximate 24.6 million parameters, 24 layers (stacked LIK and GIK blocks), 98. 6\% accuracy in BloodMNIST and an F1-score of approximately 0.986 (macro). The key contribution is the fusion of Local Information KAN (LIK) and Global Information KAN (GIK) modules, enabling the capture of fine-grained textures and global contextual features with high computational efficiency. MedKAN-B is the best-performing model in this study.
In another study, ~\citep{lv2025poolkannext} proposed PoolKANNeXt, a lightweight KAN-inspired CNN for biomedical and general image classification. The study utilized six datasets: peripheral blood cells (17,092 images, 8 classes; 12,837 train), urine sediment cells (6,687 images, 7 classes; 5,015 train), colorectal cancer histology (100,000 images, 9 classes; 75,013 train), a hybrid multi-cancer MRI/histology dataset (60,000 images, 12 classes; 48,000 train), retinal OCT8 (29,200 images, 8 classes; 26,400 train), and CIFAR-10 (60,000 images, 10 classes; 50,000 train). PoolKANNeXt consists of 106 layers and approximately 3.8 million parameters, leveraging dual-activation KAN blocks (GELU and Swish) with pooling-based feature mixing to improve efficiency. It achieved 99.43\% accuracy and 99.43\% geometric mean ($\approx$ F1‑score) on CIFAR‑10, with all biomedical datasets exceeding 93\% accuracy. Its main contribution is combining KAN-inspired dual activations with pooling-based feature fusion to deliver high accuracy with minimal parameters, and PoolKANNeXt is the best-performing model in this study. Additionally, ~\citep{vermavgg} proposed VGG‑KAN, a hybrid model combining VGG‑19 as a feature extractor with KAN layers for Alzheimer’s diagnosis using OASIS (6,488 MRI images; 4 classes; $\sim$5,171 train, 1,317 test) and ADNI (3,908 images; 2 classes; $\sim$3,160 train, 748 test). VGG‑KAN has $\sim$20M parameters with 16 convolutional, 3 fully connected, and 4 KAN layers (23 total). It achieved 95.67\% accuracy, F1-score $\approx$ 0.956 on OASIS and 97.86\% accuracy, F1 $\approx$ 0.9757 on ADNI, outperforming other CNNs and ensembles. The fusion of KAN layers with VGG‑19 captures complex nonlinear MRI features and enhances interpretability via Grad‑CAM, making VGG‑KAN the best-performing model in the study.
On the other hand, 
\citep{lu2025gdkansformer} proposed GDKansformer, which integrates Group-wise Dynamic KAN (GDKAN) with Multi-View Gated Attention (MVGA) to jointly capture fine-grained local features and global contextual dependencies for pathological image diagnosis. Three variants were developed: GDKansformer-S (4.74M parameters, 5 layers), GDKansformer-B (7.68M parameters, 8 layers), and GDKansformer-L (13.77M parameters, 11 layers). The study employed four publicly available pathological datasets: NCT-CRC-HE 100,000 training patches and 7,180 test patches for colorectal cancer tissue classification, BreaKHis (9,109 breast tumor images across four magnifications), BACH (400 microscopic H\&E breast tissue images, four classes), and the Chaoyang dataset (6,160 colon section images, four lesion types). All images were resized to 224$\times$224 pixels, and three-fold cross-validation was used to ensure fair evaluation. Experimental results showed that GDKansformer-L achieved the best overall performance, with 98.21\% accuracy and an F1-score of 87.54\% on NCT-CRC-HE, 97.85\% accuracy and F1-score of 89.62\% on BreaKHis, 87.93\% accuracy with F1-score of 75.60\% on BACH, and 93.64\% accuracy with specificity of 95.68\% on Chaoyang. By fusing dynamic KAN with MVGA, the model achieves better adaptability and efficiency, with GDKansformer-L showing the highest performance.

The literature shows that deep learning has achieved significant success in medical image classification across domains such as brain tumor MRI, chest X‑rays for COVID‑19 and tuberculosis, and skin cancer, predominantly using over‑parameterized CNNs or transfer learning models (e.g., ResNet, DenseNet, VGG, Xception, EfficientNet) with tens of millions of parameters. Although these models often achieve over 95\% accuracy, their reliance on extensive computational resources and large labelled datasets limits their applicability in resource-constrained settings, particularly for small datasets prone to overfitting, poor generalization, and low interpretability. In contrast, KAN has emerged as a lightweight, parameter‑efficient alternative that captures local and global features through learnable univariate functions, reducing model size without losing flexibility. 

However, existing studies focus mainly on synthetic or benchmark data sets (e.g., MedMNIST, CIFAR10, and various large-size histopathology image datasets), with no direct applications to small, real-world medical imaging data sets such as brain MRI, chest radiographs, or skin lesions. Moreover, the potential of KANs to achieve high diagnostic accuracy under limited data and computational resources, and the applicability of lightweight spline-based KAN variants in medical imaging, remain underexplored.

This pronounced research gap directly motivates our study, which introduces and rigorously evaluates three novel parameter-efficient spline-based KAN architectures, SBTAYLORKAN, SBRBFKAN, and SBWAVELETKAN, designed for small and heterogeneous medical imaging datasets. Unlike conventional methods that rely on over-parameterized CNNs and computationally expensive preprocessing steps, our proposed models are designed to operate directly on raw image data. This approach significantly reduces the number of trainable parameters while maintaining strong performance, particularly in data-scarce environments. Furthermore, integrating GradCAM provides transparent and clinically interpretable decisions. This bridges high-accuracy deep learning with real-world clinical use and supports the development of lightweight, resource-efficient, and reliable medical AI.

\section{Methodology}
\label{methodology}
 This section demonstrates the proposed KAN-based classification process for various medical image datasets. In the first steps, the datasets were collected, and without employing any preprocessing, they were split into training, validation, and testing sets. Subsequently, three optimized  KAN models, SBTAYLOR, SBRBF, and SBWAVELET, were developed, each applying different nonlinear feature transformations for effective representation learning. These models were trained and evaluated step-by-step under multiple test scenarios: (1) testing on the original datasets, (2) cross-dataset evaluation with related medical images, and (3) external validation on MedMNIST subsets. A quantitative analysis was also performed to evaluate the effect of reducing the number of training images.
Model performance was assessed using accuracy, precision, recall, F1-score, and area under the curve (AUC), along with statistical measures such as the KC, MCC, error rate, and p-values to ensure reliability. Finally, comparative experiments with existing models were conducted, and GradCAM-based XAI was applied to visualize critical regions, enhancing interpretability for medical decision-making. All the processes are demonstrated below in sequence.

\subsection{Kolmogorov–Arnold Formula for Multivariable Functions}

KANs are grounded in the Kolmogorov–Arnold representation theorem (KART) \citep{KARTtheory1}, which is also referred to as the superposition theorem. This theorem is a fundamental result in approximation theory \citep{KARTtheory2}.

The theorem states that any continuous multivariate function defined on a bounded domain can be expressed as a composition of a finite number of continuous univariate functions and a set of linear operations, specifically additions. More formally, for any smooth function \( f : [0,1]^{p_0} \to \mathbb{R}^{p_L} \), the KART guarantees the existence of continuous univariate functions \( \psi_k \) and \( \phi_{j,k} \) such that:

\begin{equation}
f(x) = \sum_{k=1}^{2p_0+1} \psi_k \left( \sum_{j=1}^{p_0} \phi_{j,k}(x_j) \right)
\label{eq:1}
\end{equation}

Equation~\ref{eq:1}, \( x = (x_1, \dots, x_{p_0}) \) represents the input vector consisting of \( p_0 \) variables, and the multivariate continuous function \( f(x) \) is expressed. The formula includes two levels of summation: an outer sum over \( 2p_0 + 1 \) terms of \( \psi_k : \mathbb{R} \to \mathbb{R} \), and for each \( k \), an inner sum over \( p_0 \) terms of \( \phi_{j,k} : [0,1] \to \mathbb{R} \), each acting on an individual input variable \( x_j \).

This decomposition illustrates the core principle behind the KAN architecture, where each transformation is built from univariate operations along the network edges followed by additions. The original Kolmogorov–Arnold representation corresponds to a two-layer KAN with the architectural shape \([p_0, 2p_0+1, 1]\), which can be viewed as a specific instance of the broader KAN framework \citep{KAN2024}.

\subsection{Kolmogorov–Arnold Network}
The KART states that any continuous multivariate function, defined on a bounded domain, can be represented as a combination of continuous univariate functions and additions. There are no strict limitations on choosing these univariate functions, as long as they are continuous. However, practical considerations may guide the selection of these functions depending on the specific problem and the desired properties. In the case of KANs implemented by \citep{KAN2024}, each KAN layer consists of a combination of spline functions and the sigmoid linear unit (SiLU) activation function, which have learnable coefficients.

One challenge splines face is the curse of dimensionality (COD), which arises because splines do not take full advantage of compositional structures. On the other hand, multi-layer perceptrons (MLPs) are less affected by COD due to their ability to learn features, but they tend to be less precise than splines in lower-dimensional settings. This is primarily because MLPs are not as efficient when it comes to optimizing univariate functions. A good model needs to successfully capture compositional structures (external degrees of freedom) and be able to approximate univariate functions (internal degrees of freedom) to learn the function accurately. For a more detailed discussion of KANs, refer to the work by \citep{KAN2024}. Below, we provide a brief summary of the KAN model.

\begin{figure}[!t]
    \centering
    \includegraphics[width=0.9\textwidth]{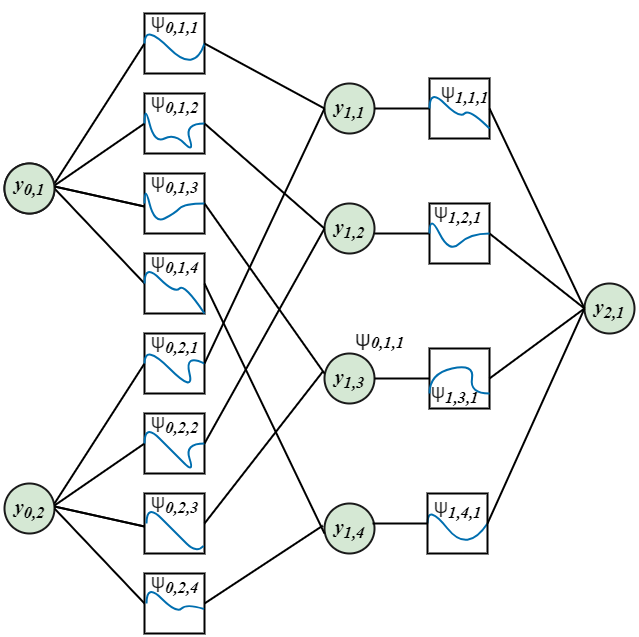}  
    \caption{Visualization of KAN model activation functions propagating through the network.}
    \label{fig:kan_diagram}
\end{figure}

A KAN is composed of several layers, as shown in Fig. \ref{fig:kan_diagram}, which displays a KAN with two layers. A KAN layer, with \( p_{\text{in}} \)-dimensional inputs and \( p_{\text{out}} \)-dimensional outputs, can be represented by a matrix of 1D functions, denoted as:
\begin{equation}
\Phi = \{ \psi_{k,m,n} \}, \quad n = 1, 2, \dots, p_{\text{in}}, \quad m = 1, 2, \dots, p_{\text{out}},
\end{equation}
where \( \psi_{m,n} \) are the learnable functions. The KAN architecture is described by an integer array \([p_0, p_1, \dots, p_L]\), where \( p_l \) indicates the number of neurons (or nodes) in the \( l \)-th layer (illustrated in Fig. \ref{fig:kan_diagram}). The \( j \)-th neuron in the \( l \)-th layer is labeled as \( (l, j) \), and the activation value of this neuron is represented by \( y_{l,j} \). Between any two consecutive layers \( l \) and \( l+1 \), there are \( p_l p_{l+1} \) activation functions. The activation function connecting the neurons \( (l, j) \) and \( (l+1, k) \) is represented by:
\begin{equation}
\psi_{l,j,k}, \quad l = 0, \dots, L-1, \quad j = 1, \dots, p_l, \quad k = 1, \dots, p_{l+1}.
\end{equation}
As shown in Fig. \ref{fig:kan_diagram}, the activation functions are applied to the edges between neurons, rather than at the nodes themselves, distinguishing KANs from MLPs. Specifically, the pre-activation of \( \psi_{l,j,k} \) is \( y_{l,j} \), while its post-activation is denoted by \( \tilde{y}_{l,j,k} \). These post-activation values \( \tilde{y}_{l,j,k} \) are then summed to calculate the activation value \( y_{l+1,k} \) for the \( (l+1,k) \)-th neuron:
\begin{equation}
y_{l+1,k} = \sum_{j=1}^{p_l} \tilde{y}_{l,j,k} = \sum_{j=1}^{p_l} \psi_{l,j,k}(y_{l,j}).
\end{equation}
This process can also be represented in matrix form as:
\begin{equation}
y_{l+1} = \begin{pmatrix}
\psi_{l,1,1}(\cdot) & \psi_{l,1,2}(\cdot) & \dots & \psi_{l,1,p_l}(\cdot) \\
\psi_{l,2,1}(\cdot) & \psi_{l,2,2}(\cdot) & \dots & \psi_{l,2,p_l}(\cdot) \\
\vdots & \vdots & \ddots & \vdots \\
\psi_{l,p_l+1,1}(\cdot) & \psi_{l,p_l+1,2}(\cdot) & \dots & \psi_{l,p_l+1,p_l}(\cdot)
\end{pmatrix}
\Phi_l y_l,
\end{equation}
where \( \Phi_l \) is the function matrix corresponding to the \( l \)-th KAN layer. The overall KAN is expressed as a composition of \( L \) layers:
\begin{equation}
\text{KAN}(y) = (\Phi_{L-1} \circ \Phi_{L-2} \circ \dots \circ \Phi_1 \circ \Phi_0) y.
\end{equation}
The original Kolmogorov-Arnold representation (See Equation~\ref{eq:1}) is a special case of the KAN, which consists of two layers with a shape of \([p_0, 2p_0+1, 1]\). On the other hand, the commonly known MLP alternates between linear transformations \( W \) and nonlinear activation functions \( \sigma \):
\begin{equation}
\text{MLP}(y) = (W_{L-1} \circ \sigma \circ W_{L-2} \circ \sigma \circ \dots \circ W_1 \circ \sigma \circ W_0) y.
\end{equation}
In MLPs, the linear transformations \( W \) and the nonlinear activation functions \( \sigma \) are applied separately, whereas in KANs, both the linear and nonlinear operations are combined within \( \Phi \).

\subsection{Traditional Spline-based Basis Function}
As discussed, traditional MLP structures utilize fixed activation functions, such as Rectified Linear Unit (ReLU), LeakyReLU, PReLU, Sigmoid, and others, applied to nodes (neurons). However, KAN shifts the trainable spline-based activation functions to the individual edges (weights), making each spline function parameterized \citep{KAN2024}.

This enables the approximation of a complex function \( f(x) \) through linear combinations and transformations between individual parameterized spline functions \( \phi_{j,k} \). The Kolmogorov–Arnold representation can be formulated as:
\begin{equation}
f(x) = \sum_{k=1}^{2p_0+1} \psi_k \left( \sum_{j=1}^{p_0} \phi_{j,k}(x_j) \right)
\end{equation}
where \( x = (x_1, \dots, x_{p_0}) \) is the input vector, \( \phi_{j,k}(x_j) \) represents a parameterized activation function (spline function) applied to individual input components, and \( \psi_k \) is the outer univariate function acting on their sum.

In KAN, B-spline functions serve as the inner activation functions \( \phi_{j,k} \). These spline functions are piecewise polynomials, determined by control points and knots. The B-spline function is particularly well-suited for KAN due to its adaptability in modelling complex nonlinear relationships. A typical inner function \( \phi_{j,k} \) can be described as:
\begin{equation}
\phi_{j,k}(x_j) = \gamma \left( d(x_j) + \text{spline}(x_j) \right)
\end{equation}
where \( \gamma \) is a trainable weight and \( d(x_j) \) is a base function, implemented here by the Sigmoid Linear Unit (SiLU), which is given by:
\begin{equation}
d(x_j) = \text{SiLU}(x_j) = \frac{x_j}{1 + e^{-x_j}}
\end{equation}
The spline term is expressed as:
\begin{equation}
\text{spline}(x_j) = \sum_{r} e_r C_r(x_j)
\end{equation}
where \( e_r \) are the coefficients learned through training, and \( C_r(x_j) \) is the B-spline basis function—a piecewise polynomial used to construct complex curves. These basis functions are employed to model intricate patterns or distributions along each input dimension.

\subsection{SBRBF-KAN (Spline-Based Radial Basis Function Kolmogorov Arnold Network)}
This study introduces the spline-based RBF KAN model, which utilizes Radial Basis Functions (RBFs) and B-spline basis functions. Generally, an RBF is a real-valued function that depends solely on the distance from a center point. Specifically, the Gaussian RBF is well known as a variant within the family of radial basis functions \citep{RBF1, RBF2}.
 
\subsubsection{RBF Transformation for Input Features}
The transformation of each input feature \( x_i \) in the BSRBF\_KANLayer using a Gaussian Radial Basis Function (RBF) is given by:
\begin{equation}
R_l(x_i, g_j) = \exp \left( - \frac{(x_i - g_j)^2}{\beta^2} \right), \quad \substack{i = 1, 2, \dots, n \\ j = 1, 2, \dots, m}
\end{equation}
where \( n \) is the input dimension to layer \( l \), and \( m \) represents the number of grid points or RBF centers. The superscript \( l \) denotes the layer number, with \( l = 1, 2, \dots, L \), where \( L \) is the total number of layers in the model. The grid points \( g_j \) are the \( j \)-th centers within the grid \( G_l \), where \( G_l = \{ g_1, g_2, \dots, g_m \} \) is the set of grid centers for layer \( l \). These grid centers \( g_j \) can either be learnable parameters or fixed points initialized at the start of the model's construction.

\subsubsection{Scaling Factor and Smoothness of RBF}
The scaling factor \( \beta \) controls the width of the Gaussian function, and it is defined as:
\begin{equation}
\beta = \frac{g_{\text{max}} - g_{\text{min}}}{m - 1}
\end{equation}
where \( g_{\text{max}} \) and \( g_{\text{min}} \) are the maximum and minimum values of the grid, respectively. The value of \( \beta \) dictates the smoothness and locality of the RBF transformation, and it is crucial for determining how much influence each grid center has over the input features.

\subsubsection{Feature Transformation with RBF}
After applying the RBF transformation to the input feature \( x_i \), the resulting features are collected into a vector, which encodes the proximity of the input to each grid center \( g_j \). These transformed features are then linearly combined using a learnable weight matrix \( W_l \), producing the output \( x_{l+1} \) of the layer:
\begin{equation}
x_{l+1} = W_l R_l(x^l, G_l)
\end{equation}
where \( x_{l+1} \) is the output of the \( (l+1) \)-th layer, \( x^l \) is the input to the current layer, and \( R_l(x^l, G_l) \) is the vector of RBF-transformed features for the input \( x^l \) with respect to the grid \( G_l \). The weight matrix \( W_l \) learns the optimal linear combination of these RBF-transformed features to produce the final output. The Gaussian RBF transformation creates a smooth, localized response for each input, capturing non-linear relationships within the data. This ability to model complex patterns enhances the model's representational power, enabling it to capture intricate dependencies within the feature space.

\subsubsection{Incorporation of B-splines into the Model}
In addition to the RBF transformation, the layer also incorporates B-splines. The B-splines are applied to the input features, and their values are computed based on the input's distance from the grid points. The resulting transformed features from both the RBF and B-splines are then concatenated and passed through a linear transformation using a learnable weight matrix \( W_l \) to compute the final output:
\begin{equation}
x_{l+1} = W_l \left[ B_s(x^l) \mid R_l(x^l, G_l) \right]
\end{equation}
where \( B_s(x^l) \) represents the output of the B-spline transformation, and \( R_l(x^l, G_l) \) is the output of the RBF transformation for the input \( x^l \) with respect to the grid \( G_l \). The combined output is then linearly mapped using the weight matrix \( W_l \) to form the output \( x_{l+1} \). This combination of RBF and B-splines enables the model to capture smooth and localized non-linear patterns in the data.

\subsubsection{Gradient Calculation and Backpropagation in the Model}
During training, the model minimizes a loss function \( L \) by adjusting the learnable weights \( W_l \) using backpropagation. The gradients of the loss function with respect to the weights and grid centers are computed and used to update the network.

The gradient of the loss function \( L \) with respect to the weight matrix \( W_l \) is computed as:
\begin{equation}
\frac{\partial L}{\partial W_l} = \frac{\partial L}{\partial x_L} \cdot W_{L-1} \cdot \frac{\partial R_{L-1}}{\partial x_{L-1}} \cdot \dots \cdot W_l \cdot \frac{\partial R_l}{\partial G_l}
\end{equation}
Where:
- \( \frac{\partial L}{\partial W_l} \) is the gradient of the loss with respect to the weight matrix \( W_l \),
- \( \frac{\partial R_l}{\partial G_l} \) is the gradient of the RBF transformation with respect to the grid centers \( G_l \),
- \( \frac{\partial R_l}{\partial x_l} \) is the gradient of the RBF transformation with respect to the input \( x_l \).

\subsubsection{Gradient with Respect to Grid Centers}
Since the grid centers \( G_l \) are fixed in the model (i.e., they are not learnable), only the weight matrices \( W_l \) are updated during backpropagation. The gradient of the RBF transformation with respect to the grid centers is computed as:
\begin{equation}
\frac{\partial R_l}{\partial g_j} = 2 \left( x_l - g_j \right) \cdot \exp \left( - \frac{(x_l - g_j)^2}{\beta^2} \right)
\end{equation}
Where:
- \( \frac{\partial R_l}{\partial g_j} \) is the gradient of the RBF with respect to the grid center \( g_j \),
- \( \beta \) is the scaling factor of the RBF.

\subsubsection{Updating the Learnable Weights}
Finally, the learnable weights \( W_l \) are updated using an optimization algorithm based on the computed gradients.

Integrating RBFs and B-splines in the BSRBF-KANLayer enables it to transform input features and capture complex nonlinear patterns. These transformed features are linearly combined using a learnable weight matrix, with updates occurring during backpropagation based on the loss function's gradients. This method allows the model to effectively capture smooth and localized patterns, enhancing its ability to manage complex data structures.

\begin{algorithm}
\small
\caption{Forward Computation of the SBRBF-KAN Network}
\label{alg:bsrbf-kan}
\KwIn{$I \in \mathbb{R}^{3 \times H \times W}$}
\KwOut{$p \in \mathbb{R}^C$}
\BlankLine
\textbf{1. Convolutional Feature Extraction}\\
$X_1 \gets \mathrm{ReLU}\bigl(\mathrm{Conv}_1(I)\bigr)$\\
$X_2 \gets \mathrm{MaxPool}(X_1)$\\
$X_3 \gets \mathrm{ReLU}\bigl(\mathrm{Conv}_2(X_2)\bigr)$\\
$X_4 \gets \mathrm{MaxPool}(X_3)$\\
$x \gets \mathrm{Flatten}(X_4)$
\BlankLine
\textbf{2. BSRBF-KAN Layers}\\
\For{$i=1$ \KwTo $3$}{
  $v \gets$ input to layer $i$ (set $v = x$ for $i=1$)\\
  $v_n \gets \mathrm{LayerNorm}(v)$\\
  $b \gets W_b^{(i)}\,\phi(v_n)$\\
  $B \gets \mathrm{B\_Spline}(v_n)$\\
  $R \gets \mathrm{RBF}(v_n)$\\
  $s \gets W_s^{(i)}\,\bigl[B \,\Vert\, R\bigr]$\\
  $h_i \gets b + s$\\
  \If{$i<3$}{$h_i \gets \mathrm{ReLU}(h_i)$}
}
\BlankLine
\textbf{3. Softmax Output}\\
$p_c = \frac{\exp(h_3[c])}{\sum_j \exp(h_3[j])} \quad \forall c \in \{1,\dots,c\}$
\BlankLine
\Return $p$
\end{algorithm}

\vspace{0.3em}
\noindent\textbf{Notation for SBRBF-KAN Algorithm}\\[-0.3em]

\begin{tabular}{@{}p{2.3cm}p{6.0cm}@{}}
$I$ & Input tensor.\\
$\mathrm{Conv}_i$ & $i$-th convolution.\\
$\mathrm{MaxPool}$ & Max pooling operator.\\
$X_k$ & Intermediate feature maps.\\
$x$ & Flattened vector.\\
$v$ & Input to BSRBF-KAN layer.\\
$v_n$ & Layer-normalized vector.\\
$\phi(\cdot)$ & SiLU activation.\\
$W_b^{(i)}$ & Base weights of layer $i$.\\
$B$ & B-spline embedding.\\
$R$ & RBF embedding.\\
$[B \,\Vert\, R]$ & Concatenation of $B$ and $R$.\\
$W_s^{(i)}$ & Spline+RBF weights.\\
$b$ & Base output.\\
$s$ & Spline+RBF output.\\
$h_i$ & Hidden representation.\\
$p$ & Softmax probabilities.\\
$c$ & Number of classes.\\
\end{tabular}

\subsection{SBTAYLOR-KAN (Spline-Based Taylor Basis Kolmogorov Arnold Network)}
KAN is effective for function approximation \citep{KAN2024}, but its scalability is limited in high-dimensional settings \citep{SB-TYS1}. The reliance on B-splines makes the model sensitive to knot configuration and more complex as the number of knots increases with feature dimensionality. B-splines generally allow local control of the learned activation functions, helping the model adapt to complex, high-dimensional data through piecewise polynomial transformations. However, using only B-splines can make it challenging to scale and interpret the model effectively. To overcome this, the hybrid spline-based Taylor KAN model combines two powerful techniques for multivariate function approximation: B-splines and Taylor series.

Unlike B-splines, Taylor series do not require piecewise modelling or control point optimization, making them simpler for approximating local behaviour around a given expansion point. This eliminates the need for pre-processing steps like knot selection. Additionally, it uses a truncated Taylor series for local approximation. While both methods capture local behaviour, Taylor series offer a simpler representation \citep{SB-TYS2}.

In this hybrid model, B-splines handle the global behaviour, while Taylor series manage local approximations. The combination of B-splines’ flexibility and Taylor series’ efficiency results in a more scalable and interpretable approach compared to using B-splines alone.

In the TaylorKAN hybrid model, B-splines and Taylor series are used in sequence or parallel to model complex function behaviours locally.

\subsubsection{KAN with B-splines}
The network layers use B-splines to capture local features of the data:
\begin{equation}
y = f_{\text{KAN}}(x) = W_{\text{base}} \cdot \sigma(x) + W_{\text{spline}} \cdot \text{B-splines}(x)
\end{equation}
Where:
\begin{itemize}
    \item \( W_{\text{base}} \) is the weight matrix for the linear transformation based on the base activation (e.g., ReLU or SiLU).
    \item \( W_{\text{spline}} \) is the weight matrix for the B-spline transformation.
    \item \( \text{B-splines}(x) \) is the local basis function transformation of \( x \).
\end{itemize}

\subsubsection{Taylor Series Approximation}
The Taylor series expresses a function \( f(x) \) as an infinite sum of terms based on its derivatives \( f^{(n)} \) evaluated at a specific point \( a \) (see equation \ref{eq:taylor_series}). This enables the approximation of complex functions by capturing their local behaviour around the expansion point.

\begin{equation}
f(x) \approx \sum_{n=0}^{\infty} \frac{f^{(n)}(a)}{n!} (x - a)^n
\label{eq:taylor_series}
\end{equation}
Where:
\begin{itemize}
    \item \( f^{(n)}(a) \) is the \( n \)-th derivative of \( f(x) \) evaluated at the point \( a \),
    \item \( n! \) is the factorial of \( n \),
    \item \( (x - a)^n \) is the difference between \( x \) and the expansion point \( a \), raised to the power of \( n \).
\end{itemize}

\subsubsection{Integration of B-Spline \& Taylor Series Approximation in KAN}
The Spline-based TaylorKAN model combines these transformations to form the output:

\begin{equation}
y_{\text{final}} = f_{\text{KAN}}(x) + f_{\text{Taylor}}(x)
\end{equation}
Where:
\begin{itemize}
    \item \( f_{\text{KAN}}(x) \) is the B-spline-based KAN transformation.
    \item \( f_{\text{Taylor}}(x) \) is the Taylor series approximation.
\end{itemize}
However, the complete process of the proposed Spline-based Taylor KAN module is shown in Algorithm 2.

\begin{algorithm}
\caption{Forward Computation of the Proposed SBTAYLOR-KAN Network}
\KwIn{Input image tensor $I \in \mathbb{R}^{3 \times H \times W}$}
\KwOut{Class probabilities $p \in \mathbb{R}^C$}
\BlankLine
\textbf{1. Convolutional Feature Extraction}\\
$X_1 \gets \mathrm{ReLU}\bigl(\mathrm{Conv}_1(I)\bigr)$\\
$X_2 \gets \mathrm{MaxPool}(X_1)$\\
$X_3 \gets \mathrm{ReLU}\bigl(\mathrm{Conv}_2(X_2)\bigr)$\\
$X_4 \gets \mathrm{MaxPool}(X_3)$\\
$x \gets \mathrm{Flatten}(X_4)$
\BlankLine
\textbf{2. Taylor Series Approximation}\\
$x_T \gets \displaystyle \sum_{n=0}^{N-1} (-1)^n \frac{x^{2n+1}}{(2n+1)!}$
\BlankLine
\textbf{3. KANLinear Layers}\\
$h_1 \gets \mathrm{ReLU}\bigl(W_b^{(1)}\,\phi(x_T) + \sum_{k} S^{(1)}_k \, B_k(x_T)\bigr)$\\
$h_2 \gets \mathrm{ReLU}\bigl(W_b^{(2)}\,\phi(h_1) + \sum_{k} S^{(2)}_k \, B_k(h_1)\bigr)$\\
$h_3 \gets W_b^{(3)}\,\phi(h_2) + \sum_{k} S^{(3)}_k \, B_k(h_2)$
\BlankLine
\textbf{4. Softmax Output}\\
$p_c \gets \displaystyle \frac{\exp(h_3[c])}{\sum_j \exp(h_3[j])}, \quad \forall c \in \{1,\dots,c\}$
\BlankLine
\Return $p$
\end{algorithm}

\vspace{0.3em}
\noindent\textbf{Notation for SBTAYLOR-KAN Algorithm}\\[-0.3em]

\begin{tabular}{@{}p{2.8cm}p{6.8cm}@{}}
$x_T$ & \raggedright Taylor approximation of $\sin(x)$: 
\(
x_T = \sum_{n=0}^{N-1} (-1)^n \frac{x^{2n+1}}{(2n+1)!}
\)\tabularnewline
$S_k^{(i)}$ & \raggedright B-spline coefficients \\ in KANLinear layer $i$.\tabularnewline
$B_k(\cdot)$ & \raggedright $k$-th B-spline basis function.\tabularnewline
$h_j$ & \parbox[t]{6.8cm}{\raggedright Hidden representation after\\KANLinear layer $j$.}\tabularnewline
$p$ & Softmax output vector, $p \in \mathbb{R}^c$.\\
\end{tabular}

The notations $I$, $\mathrm{Conv}_i$, $\mathrm{MaxPool}$, $x$, $\phi(\cdot)$, $W_b^{(i)}$, and $c$ are described in Algorithm~\ref{alg:bsrbf-kan}.

\subsection{SBWAVELET-KAN (Spline-Based Wavelet Kolmogorov Arnold Network)}
Our approach is conceptually related to Wav-KAN \citep{WAVKAN1}, which introduced wavelet basis functions—primarily the Morlet wavelet—as adaptive nonlinear activations within KAN. While Wav-KAN demonstrates the benefits of wavelet activations in capturing localized frequency patterns, it is limited to wavelet-only representations. In contrast, our HybridKAN architecture generalizes this idea by incorporating an additional B-spline basis expansion alongside the wavelet transform within each layer. This hybrid formulation enables the network to simultaneously model smooth, piecewise-continuous variations (via B-splines) and oscillatory features (via wavelets), with a learnable softmax mechanism controlling the contribution of each component. To our knowledge, this integration of wavelet and B-spline embeddings within a unified KAN framework has not been previously explored and offers a more flexible and expressive class of representations.

\subsubsection{Integration of B-spline \& Wavelet Transformations in the HybridKAN Layer}
In the proposed architecture, each HybridKAN linear layer combines two complementary nonlinear transformations: a wavelet-inspired activation and a B-spline basis expansion. This design enhances the network’s ability to approximate complex functions by capturing both high-frequency oscillations and smooth local trends.

Let \(v \in \mathbb{R}^{d}\) denote the input to a HybridKAN layer. The wavelet transform output is computed by applying a mother wavelet function \(\psi(\cdot)\) to each input dimension after learnable scaling and translation:

\begin{equation}
x_i = \frac{v_i - t_i}{s_i},
\end{equation}
where:
\begin{itemize}
    \item \(v_i\) is the \(i\)-th component of the input vector,
    \item \(s_i > 0\) is the trainable scale parameter for dimension \(i\),
    \item \(t_i\) is the trainable translation parameter for dimension \(i\).
\end{itemize}

Next, the scaled input is passed through the mother wavelet:

\begin{equation}
w_i = \psi(x_i),
\end{equation}
where:
\begin{itemize}
    \item \(w_i\) is the wavelet activation for input dimension \(i\),
    \item \(\psi(\cdot)\) is the chosen wavelet function, such as the Morlet wavelet.
\end{itemize}

The outputs across all input dimensions are aggregated using learnable wavelet weights:

\begin{equation}
W = \sum_{i=1}^{d}\mathbf{W}_{\text{wavelet},i}\,w_i,
\end{equation}
where:
\begin{itemize}
    \item \(\mathbf{W}_{\text{wavelet},i}\) denotes the trainable weight applied to the wavelet activation of dimension \(i\).
\end{itemize}

In parallel, the input vector \(v\) is processed through a B-spline embedding. A uniform grid over a bounded interval is defined, and B-spline basis functions \(B_j(v_i)\) of order \(k\) are constructed:

\begin{equation}
B_i = \bigl[B_j(v_i)\bigr]_{j=1}^{M},
\end{equation}
where:
\begin{itemize}
    \item \(B_j(v_i)\) is the value of the \(j\)-th B-spline basis function evaluated at \(v_i\),
    \item \(M\) denotes the total number of grid points plus the spline order,
    \item \(k\) is the spline order (degree).
\end{itemize}

The B-spline outputs are linearly combined via learnable spline weights:

\begin{equation}
S = \sum_{i=1}^{d}\sum_{j=1}^{M}\mathbf{W}_{\text{spline},i,j}\,B_j(v_i),
\end{equation}
where:
\begin{itemize}
    \item \(\mathbf{W}_{\text{spline},i,j}\) is the trainable weight associated with the \(j\)-th B-spline basis of dimension \(i\).
\end{itemize}

To integrate these two transformations, the model uses a trainable combination weight vector \(\alpha \in \mathbb{R}^{2}\), normalized via a softmax function. The final output of the HybridKAN layer is:

\begin{equation}
h = \alpha_1\,W + \alpha_2\,S,
\end{equation}
where:
\begin{itemize}
    \item \(\alpha = [\alpha_1, \alpha_2]\) are the softmax-normalized coefficients controlling the contribution of the wavelet and B-spline outputs,
    \item \(h\) is the aggregated output of the HybridKAN layer, optionally passed through batch normalization and a nonlinearity.
\end{itemize}

\subsubsection{Choice of Mother Wavelet}
The function \(\psi(\cdot)\) serves as the mother wavelet, determining the functional shape of the wavelet transform. In this work, unless otherwise specified, we instantiate \(\psi\) as the Morlet wavelet:

\begin{equation}
\psi_{\text{Morlet}}(x) = \exp\bigl(-\frac{1}{2}\,x^2\bigr)\,\cos(\omega_0 x),
\end{equation}
where:
\begin{itemize}
    \item \(\omega_0 = 5\) controls the oscillation frequency.
\end{itemize}

This wavelet is widely used due to its strong localization and oscillatory behaviour. However, the complete forward computation procedure of our proposed Wavelet-BSpline-KAN network is outlined in Algorithm 3.

\begin{algorithm}
\small
\caption{Forward Computation of the SBWAVELET-KAN Network}
\KwIn{$I \in \mathbb{R}^{3 \times H \times W}$}
\KwOut{$p \in \mathbb{R}^C$}
\BlankLine
\textbf{1. Convolutional Feature Extraction}\\
$X_1 \gets \mathrm{ReLU}\bigl(\mathrm{Conv}_1(I)\bigr)$\\
$X_2 \gets \mathrm{MaxPool}(X_1)$\\
$X_3 \gets \mathrm{ReLU}\bigl(\mathrm{Conv}_2(X_2)\bigr)$\\
$X_4 \gets \mathrm{MaxPool}(X_3)$\\
$x \gets \mathrm{Flatten}(X_4)$
\BlankLine
\textbf{2. Wavelet-BSpline-KAN Layers}\\
\For{$i=1$ \KwTo $3$}{
  $v \gets$ input to layer $i$ (set $v = x$ for $i=1$)\\
  $v_n \gets v$\\
  \textit{// Wavelet Transform}\\
  $W \gets \mathrm{WaveletTransform}(v_n)$\\
  \textit{// B-spline Transform}\\
  $B \gets \mathrm{B\_Spline}(v_n)$\\
  \textit{// Compute weighted combination}\\
  $C \gets \mathrm{Softmax}\bigl(\mathrm{combine\_weight}^{(i)}\bigr)$\\
  $h_i \gets C_1 \cdot W + C_2 \cdot B$\\
  $h_i \gets \mathrm{BatchNorm}(h_i)$\\
  \If{$i<3$}{$h_i \gets \mathrm{ReLU}(h_i)$}
}
\BlankLine
\textbf{3. Softmax Output}\\
$p_c = \frac{\exp(h_3[c])}{\sum_j \exp(h_3[j])} \quad \forall c \in \{1,\dots,c\}$
\BlankLine
\Return $p$
\end{algorithm}

\vspace{0.5em}
\noindent\textbf{Notation for SBWAVELET-KAN Algorithm}
\vspace{0.3em}

\begin{tabular}{@{} p{2.5cm} p{6.0cm} @{}}
$v$ & Input to Wavelet-BSpline-KAN layer. \\
$v_n$ & Normalized input (identity in this case). \\
\texttt{B\_Spline} & B-spline embedding function. \\
\texttt{WaveletTransform} & Wavelet embedding function (e.g., Morlet). \\
$\mathrm{combine\_weight}^{(i)}$ & Learnable combination weight for layer $i$. \\
$C$ & Softmax-normalized combination weights. \\
\end{tabular}

The notations $I$, $\mathrm{Conv}_i$, $\mathrm{MaxPool}$, $X_k$, $x$, $h_i$, $p$, and $c$ are already described in Algorithm~\ref{alg:bsrbf-kan}.

\section{Experiment and Results}
\label{experiment_results}
This section presents the diverse datasets sourced from multiple repositories that were utilized in our experiments, along with a detailed description of the experimental setup. It also includes a comprehensive quantitative evaluation demonstrating the effectiveness and robustness of the proposed model architecture. Furthermore, a rigorous comparative analysis is conducted against existing approaches in the literature to highlight the model’s performance.

\subsection{Dataset Description}

\subsubsection{Category-1: Training, Validation and Testing}

In this study, four diverse medical imaging datasets are utilized for experimental analysis: the Brain Tumor Kaggle Dataset (comprising no tumor, glioma, meningioma, and pituitary classes), Chest X-ray Dataset-1 (containing Normal, COVID-19, and Pneumonia cases), Chest X-ray Dataset-2 (featuring Normal and Tuberculosis cases), and a Skin Cancer dataset of six classes. Comprehensive details of these datasets are presented in Table~\ref{tab:dataset_splits}.

\begin{figure}[!t]
    \centering
    \includegraphics[width=0.9\textwidth]{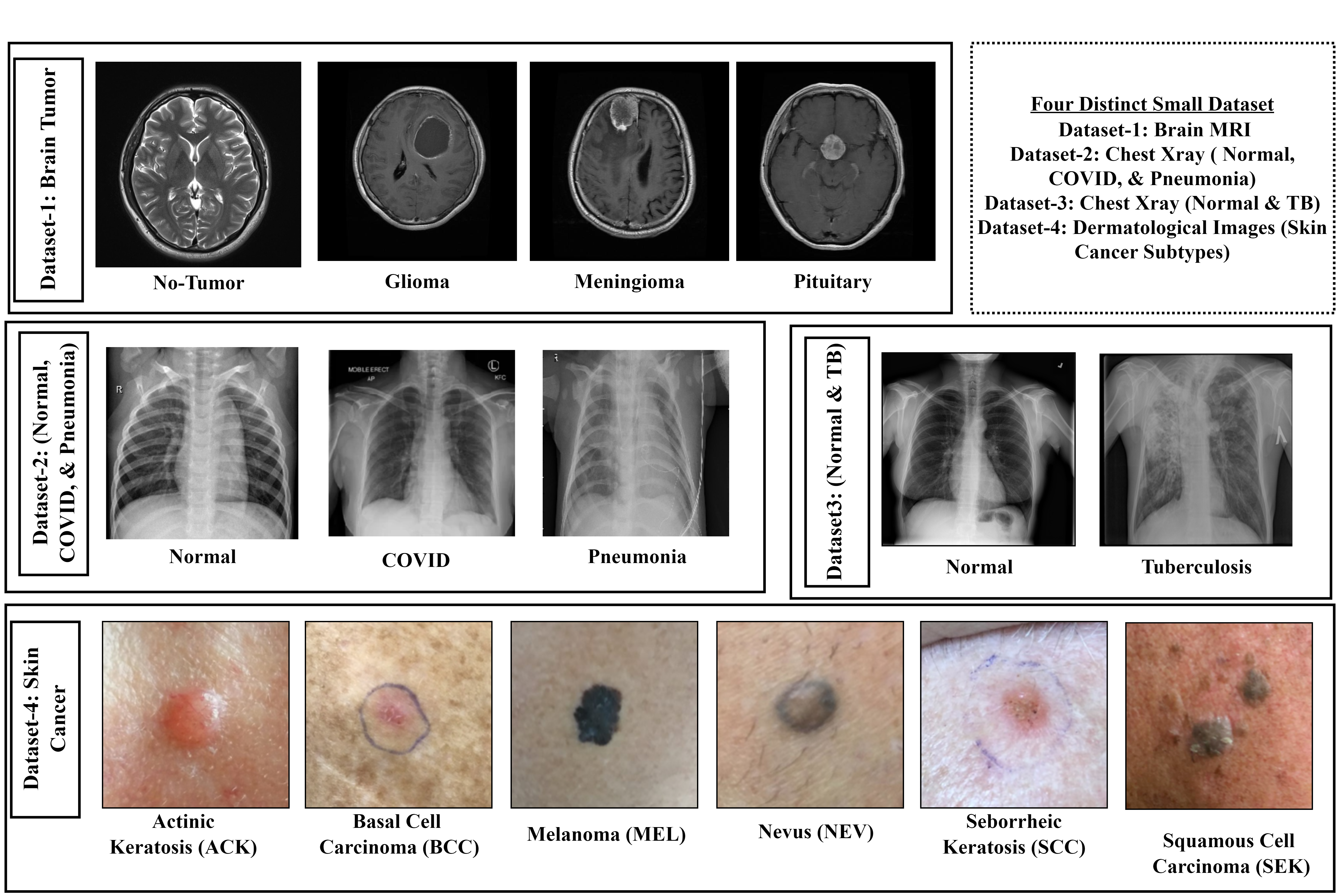}  
    \caption{Sample images from four medical imaging datasets.}
    \label{fig:dataset_images}
\end{figure}

\paragraph{\textbf{Dataset-1: Brain Tumor Kaggle Dataset}}  
The publicly accessible MRI dataset from the Kaggle repository was utilized for this study. This brain tumor MRI dataset~\citep{Nickparvar-BTMD} is a composite of three publicly accessible datasets: Figshare~\citep{brain-tumor-dataset}, SARTAJ~\citep{SB-BTCMRI}, and Br34H~\citep{BTD-AH}. This dataset contains 7,023 images across four classes—no tumor, glioma, meningioma, and pituitary —organized into two folders (training and testing). However, this study utilizes 5,712 images (no tumor 1,595 images, glioma 1,321 images, meningioma 1,339 images, and pituitary 1,457 images) from the training dataset for training, validation, and testing. Sample images from the dataset are shown in Fig.~\ref{fig:dataset_images} (Dataset-1).

\paragraph{\textbf{Dataset-2: COVID-19 Chest X-Ray Dataset}}  
This chest X-ray dataset is also available on Kaggle~\citep{SK-COVIDX}. It consists of 5,228 samples across three classes: Normal (1,802 images), Pneumonia (1,800 images), and COVID-19 (1,626 images). All images were resized to 256$\times$256 pixels in PNG format. Fig.~\ref{fig:dataset_images} (Dataset-2) presents sample images from this dataset.

\paragraph{\textbf{Dataset-3: TB Chest X-ray Database}}  
This dataset comprises 4,200 chest X-ray (CXR) images from a publicly available Kaggle repository ~\citep{TB-Kaggle}. It was curated through a collaborative effort among researchers from Qatar University (Doha, Qatar), the University of Dhaka (Bangladesh), and partners in Malaysia. These teams worked closely with medical professionals from Hamad Medical Corporation (Doha, Qatar) and healthcare institutions across Bangladesh. The dataset includes 700 CXR images exhibiting signs of tuberculosis (TB) and 3,500 normal cases. Each image has a resolution of 512$\times$512 pixels. This balanced and structured dataset provides a solid foundation for evaluating the proposed models in TB detection. A dataset sample is shown in Fig.~\ref{fig:dataset_images} (Dataset-3).

\paragraph{\textbf{Dataset-4: PAD-UFES-20 Dataset}}  
The PAD-UFES-20 dataset ~\citep{MendeleyDataset}, developed by the Dermatological and Surgical Assistance Program of the Federal University of Espírito Santo (UFES), Brazil, comprises 2,298 clinical images representing 1,641 skin lesions collected from 1,373 patients. These images were captured using various smartphone devices, introducing variability in image size, resolution, and lighting conditions. The skin lesions are categorized into six classes with the following distribution: Actinic Keratosis (ACK) – 730 samples, Basal Cell Carcinoma (BCC) – 845 samples, Melanoma (MEL) – 52 samples, Nevus (NEV) – 244 samples, Squamous Cell Carcinoma (SCC) – 192 samples, and Seborrheic Keratosis (SEK) – 235 samples. Fig.~\ref{fig:dataset_images} (Dataset-4) illustrates representative samples.

\begin{figure}[!t]
    \centering
    \includegraphics[width=0.9\textwidth]{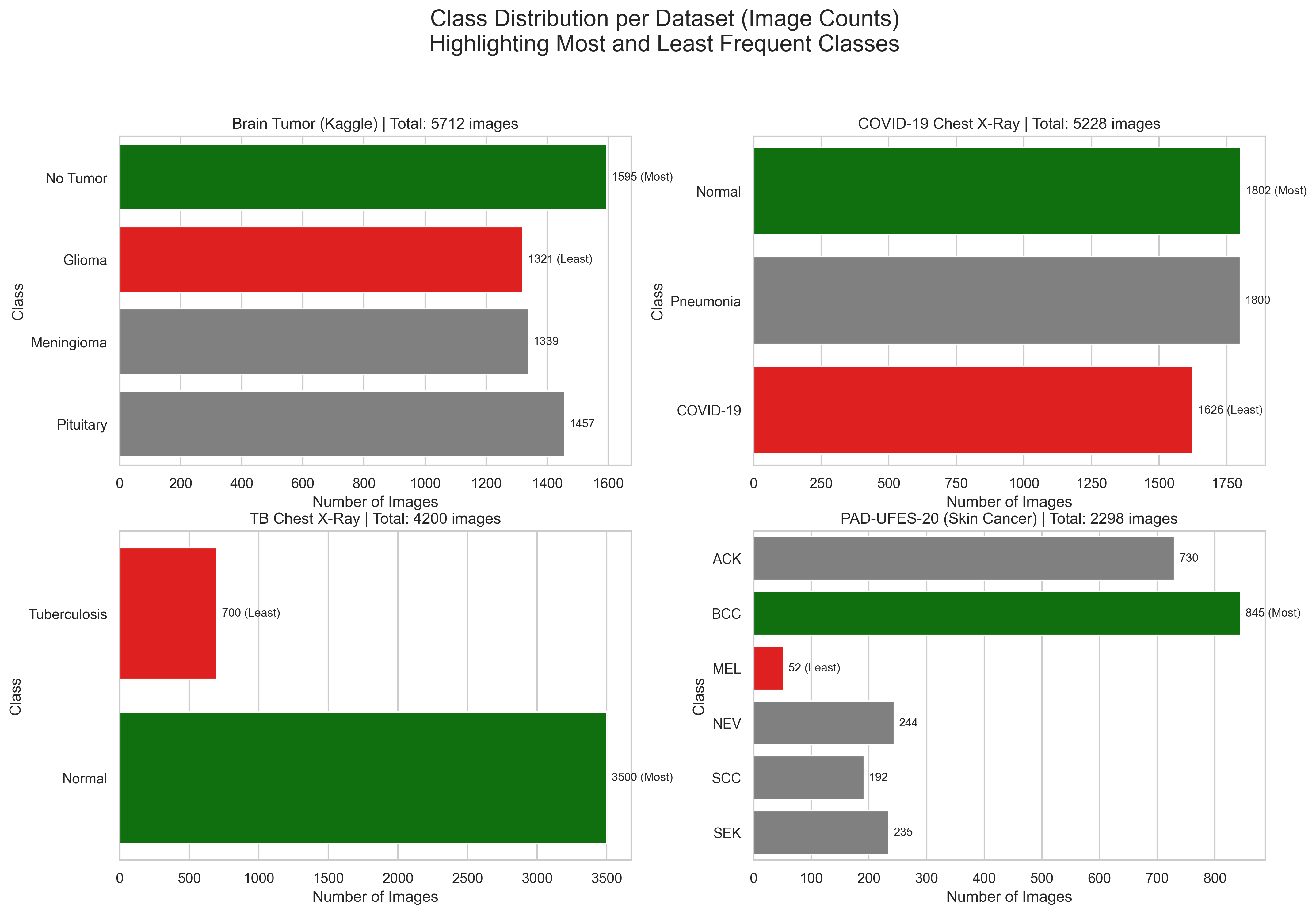}
    \caption{Class distribution across four medical imaging datasets, highlighting the most and least frequent classes.}
    \label{fig:class_distribution}
\end{figure}

Figure \ref{fig:class_distribution} shows the class distribution across four different medical imaging datasets. This visualization highlights how some classes are overrepresented while others are underrepresented, a common issue known as class imbalance. Addressing this imbalance is crucial for creating fair and effective machine learning models, especially for medical diagnostics.

The PAD-UFES-20 dataset is relatively small and exhibits a class imbalance, which can impact the performance of machine learning models. It contains six classes, with the ACK (0) and BCC (1) classes already having enough images. To address the class imbalance, we created a balanced dataset of 3,000 images, with 500 images from each class. For the ACK (0) and BCC (1) classes, which already had sufficient samples, we randomly selected 500 images from each. For the remaining classes, we employed augmentation techniques such as flipping, rotation, and adjustments to brightness and sharpness to generate additional images.

The primary objective of this experiment was to evaluate the performance of the model in both imbalanced and balanced data sets. This comparison is crucial for understanding how class imbalance influences model accuracy and generalization. In medical applications, where datasets are often limited and imbalanced, this insight is especially important. Accurate predictions are crucial for reliable diagnoses, and addressing class imbalance is key to improving the effectiveness of machine learning models in medical applications.

\subsubsection{Category-2: Evaluating Diverse Datasets to Assess Model Robustness}

To rigorously assess the robustness of the model, this study introduces four additional test datasets drawn from the same diagnostic category but independently sourced. These test sets, entirely unseen during training, are used solely to evaluate generalization and robustness. Unlike the original test set, they present the model with new and diverse scenarios more closely resembling real-world conditions. To ensure fair and meaningful comparisons, the design of each new test set was carefully controlled, closely matching the size of the original test set and maintaining balanced class distributions (e.g., an equal number of images per class). This thoughtful construction enables a more accurate and reliable assessment of the model’s stability in practical settings. The dataset details are provided below.

\paragraph{\textbf{New Testing Dataset-I (Brain Tumor Kaggle Dataset):}}  
As previously noted, a separate testing folder is dedicated to the publicly available Kaggle dataset~\citep{Nickparvar-BTMD}. This isolated dataset comprises 1,311 MRI images, including 405 no-tumor, 300 glioma, 306 meningioma, and 300 pituitary cases. For robustness evaluation, 100 images from each class were randomly selected, ensuring balanced representation across categories.

\paragraph{\textbf{New Testing Dataset-II (COVID-19 Radiography Dataset):}}  
A collaborative team of researchers from Qatar University, the University of Dhaka, and institutions in Pakistan and Malaysia, working alongside medical professionals, developed a comprehensive chest X-ray image database to support the research of COVID-19. This dataset includes~\citep{RahmanCOVID19} images of COVID-19 positive cases, as well as normal and viral pneumonia cases, and has been released in multiple stages. In the second updated version, the dataset expanded significantly by including 3,616 COVID-19, 10,192 normal, and 1,345 viral pneumonia images. In the dataset, pneumonia cases are labelled explicitly as viral pneumonia. Although this is a subset of the broader pneumonia category, it still reflects the core characteristics of pneumonia that are relevant to model evaluation. To ensure consistency, 600 test images are employed, comprising 200 images selected from each of three classes.

\paragraph{\textbf{New Testing Dataset-III (Shenzhen \& Montgomery CXR Dataset):}}  
To evaluate the robustness of the models, two separate datasets within the same category are utilized to construct the test set. The first dataset~\citep{TB-CXR-Shenzhen} includes 326 images labelled as normal and 336 labelled as TB. The second dataset~\citep{MC-CXR} comprises approximately 103 normal and 295 TB images. From these, a balanced test dataset is curated, consisting of around 200 images—100 from the normal class and 100 from the TB class.

\paragraph{\textbf{New Testing Dataset-IV (Fluorescence Skin Cancer (FLUO-SC) Dataset):}}  
The Fluorescence Skin Cancer (FLUO-SC) dataset~\citep{FLUO-SC2024} comprises a total of 1,563 clinical images paired with 1,563 corresponding fluorescence images, collected over the span of one year using the LINCE autofluorescence imaging device. This dataset was developed as part of the Dermatological and Surgical Assistance Program (PAD) at the Federal University of Espírito Santo (UFES), aimed at offering free treatment for skin lesions to underserved rural communities. The dataset also includes six categories of skin lesions: three malignant (BCC with 496 images, SCC with 78 images, and MEL with 19 images), and three benign (ACK with 560 images, SEK with 252 images, and NEV with 158 images). All images are stored in .jpg format and vary in size due to using different smartphone cameras for image acquisition. Ethical approval for the dataset was obtained from the UFES ethics committee and the Brazilian government. However, a new testing dataset has been introduced in this case, consisting of 297 images: 50 ACK, 50 BCC, 50 NEV, 50 SEK, 78 SCC, and 19 MEL images.

\subsubsection{Category-3: Evaluating MedMNIST Datasets to Assess Model Robustness}
In this phase, two MedMNIST datasets, including BreastMNIST and PneumoniaMNIST, are utilized to assess the model's robustness in terms of disease classification. The MedMNIST data set is divided into 70:20:10 in training, validation, and testing, respectively. As a result, the BreastMNIST and PneumoniaMNIST datasets have been pre-divided in the same ratio as used in our study. In this case, our proposed KAN models are retrained and tested with the testing dataset to evaluate the models' performance.

\paragraph{\textbf{BreastMNIST Dataset:}}
The BreastMNIST baseline dataset \citep{MedMNIST} consists of breast ultrasound images from 600 women, aged 25 to 75, collected in 2018. It includes 780 images across two classes: normal and abnormal (benign \& malignant), with each image averaging 500x500 pixels in PNG format. The dataset is split into training, validation, and testing sets containing 546, 156, and 78 images, respectively.

\paragraph{\textbf{PneumoniaMNIST Dataset:}} 
The PneumoniaMNIST dataset \citep{MedMNIST} contains 5,856 chest X-ray images (JPEG), divided into three folders: 4,708 images for training, 624 for testing, and 524 for validation. Anterior–posterior chest radiographs of pediatric patients aged 1–5 years, obtained during routine clinical care at Guangzhou Women and Children’s Medical Center, were classified as either Pneumonia or Normal. The images were quality-checked to remove poor scans, then graded by two expert physicians, with a third expert reviewing the evaluation set to correct any potential grading errors.

\subsection{Experimental Setup}
All experiments are conducted in Python 3.12.4 using the PyTorch framework for model development, training, and evaluation. GPU acceleration is enabled via NVIDIA CUDA 12.6 on an HPC cluster with NVIDIA GPUs. Each job runs on a single GPU with 32 GB system memory and 8 CPU cores, managed via the SLURM scheduler. The setup is compatible with NVIDIA A100-class devices.
Experiments are performed in an offline environment with all Python dependencies (matplotlib, numpy, scikit-learn, seaborn, torchsummary, Pillow, pygments) installed locally via pip using the --no-index flag to ensure reproducibility. Execution is orchestrated via SLURM batch scripts, with the primary training and evaluation handled through a Python .py script.

\subsection{Parameter Configuration and Model Differences}

The dataset is divided into 70\%, 20\%, and 10\% for training, validation, and testing, respectively, as shown in Table~\ref{tab:dataset_splits}. For the balanced PAD-UFES-20 dataset (dataset-4), this results in 2,100 images for training, 600 for validation, and 300 for testing, maintaining the same split ratio. After extensive experimentation with several configurations, the SBTAYLOR-KAN, SBWAVELET-KAN, and SBRBF-KAN models were developed using a common optimal architectural backbone composed of convolutional and customized KAN-based fully connected layers. Despite this shared structure, each model incorporates unique parametric components that distinguish their learning capabilities.

\paragraph{Common Hyperparameters:}
All three models employ the following consistent settings:
\begin{itemize}
    \item \texttt{grid\_size} = 5
    \item \texttt{spline\_order} = 3
    \item Learning Rate = 0.0001
    \item Batch Size = 32
    \item Weight Decay = $1 \times 10^{-4}$ (L2 regularization)
    \item Activation Function: Sigmoid Linear Unit (SiLU)
    \item Optimizer: Stochastic Gradient Descent (SGD)
    \item Standalone spline scaling: Enabled
\end{itemize}

\paragraph{Model-Specific Components}
\begin{enumerate}
    \item \textbf{SBTAYLOR-KAN:} 
    The Taylor KAN model incorporates a taylor series expansion module before the KAN layers. Specifically, it employs a 5-term approximation to transform the input features into a richer polynomial basis. This approach enhances the model's ability to represent smooth non-linear functions, such as sinusoidal patterns, by capturing higher-order input dynamics early in the processing pipeline. Including the taylor approximation serves as an implicit feature mapping that complements the subsequent spline-based transformations in the KAN layers.
   
    \item \textbf{SBWAVELET-KAN:} 
    The Wavelet KAN model integrates wavelet-based feature transformations—such as the Morlet wavelet—into the architecture alongside traditional B-spline representations. These components are fused through a learnable combine\_weight mechanism, allowing the model to adaptively balance wavelet frequency-localized features with smooth spatial representations from splines. Furthermore, "BatchNorm1d" is applied after each KAN layer to promote training stability and mitigate internal covariate shift.
    
    \item \textbf{SBRBF-KAN:} 
    The RBF KAN model incorporates RBF encodings as an alternative to wavelets, enabling localized non-linear transformations of the input space. These RBF features are concatenated with B-spline outputs and processed jointly through the KAN layers. The model leverages "LayerNorm" instead of batch normalization, offering consistent normalization across varying batch sizes. It also includes a dynamic mechanism to initialize and scale spline weights adaptively, enhancing its ability to learn expressive and flexible representations.
\end{enumerate}

These architectural and parametric differences critically shape how each model represents input features and generalizes across the classification task.

\begin{table}[!t]
\centering
\caption{Class-wise distribution and 70:20:10 split for training, validation, and test sets across all datasets. 
Class labels (0--5) are used for visualization in confusion matrices and figures.}
\vspace{1em} 
\label{tab:dataset_splits}
\scriptsize  

\renewcommand{\arraystretch}{1.0}  
\setlength{\tabcolsep}{3pt}  

\resizebox{\textwidth}{!}{  
\begin{tabular}{|l|l|r|r|r|r|}
\hline
\textbf{Dataset} & \textbf{Class} & \textbf{Total} & \textbf{Train} & \textbf{Val} & \textbf{Test} \\
\hline
\multirow{4}{*}{\shortstack[l]{\textbf{Dataset-1} \\ Brain Tumor}} 
 & Glioma  (0)     & 1321 & 925  & 264 & 132 \\
 & Meningioma (1)   & 1339 & 937  & 268 & 134 \\
 & No Tumor (2)    & 1595 & 1116 & 319 & 160 \\
 & Pituitary (3)    & 1457 & 1020 & 291 & 146 \\
\hline
 & \textit{Total} & \textbf{5712} & \textbf{3998} & \textbf{1142} & \textbf{572} \\
\hline
\multirow{3}{*}{\shortstack[l]{\textbf{Dataset-2} \\ COVID-19 X-ray}} 
 & COVID-19 (0) & 1626 & 1138 & 325 & 163 \\
 & Normal (1) & 1802 & 1261 & 360 & 181 \\
 & Pneumonia (2) & 1800 & 1260 & 360 & 180 \\
\hline
 & \textit{Total} & \textbf{5228} & \textbf{3660} & \textbf{1045} & \textbf{523} \\
\hline
\multirow{2}{*}{\shortstack[l]{\textbf{Dataset-3} \\ TB X-ray}} 
 & Normal (0) & 3500 & 2450 & 700 & 350 \\
 & TB (1) & 700  & 490  & 140 & 70  \\
\hline
 & \textit{Total} & \textbf{4200} & \textbf{2940} & \textbf{840} & \textbf{420} \\
\hline
\multirow{6}{*}{\shortstack[l]{\textbf{Dataset-4} \\ PAD-UFES-20}} 
 & ACK (0)          & 730  & 511  & 146 & 73  \\
 & BCC (1)         & 845  & 592  & 169 & 84  \\
 & MEL (2)         & 52   & 36   & 10  & 6   \\
 & NEV (3)        & 244  & 171  & 49  & 24  \\
 & SCC (4)         & 192  & 134  & 38  & 20  \\
 & SEK (5)        & 235  & 165  & 47  & 23  \\
\hline
 & \textit{Total} & \textbf{2298} & \textbf{1609} & \textbf{459} & \textbf{230} \\
\hline
\end{tabular}
}
\end{table}


\subsection{Evaluation Matrices}
The performance of the proposed KAN medical disease classification system is assessed using key outcome-based metrics: True Positive (TP), True Negative (TN), False Positive (FP), and False Negative (FN). These values from the confusion matrix reflect the model's accuracy in identifying normal and abnormal cases and its errors. TP represents correctly detected abnormalities, TN indicates correctly identified normal cases, FP denotes false alarms, and FN represents missed detections. All the values can be obtained from the confusion matrix.

Based on these values, core evaluation metrics — accuracy, precision, recall, and F1-score — are computed using Equations~\ref{eq:accuracy} to \ref{eq:f1score} as follows \citep{kar2024efficient}:

\begin{itemize}
    \item \textbf{Accuracy:} Measures the overall correctness of the model.
\end{itemize}
\begin{equation}
\text{Accuracy} = \frac{TP + TN}{TP + TN + FP + FN}
\label{eq:accuracy}
\end{equation}

\begin{itemize}
    \item \textbf{Precision:} Indicates how many of the predicted positive cases are truly positive.
\end{itemize}
\begin{equation}
\text{Precision} = \frac{TP}{TP + FP}
\label{eq:precision}
\end{equation}

\begin{itemize}
    \item \textbf{Recall (Sensitivity):} Measures how well the model captures positive cases.
\end{itemize}
\begin{equation}
\text{Recall} = \frac{TP}{TP + FN}
\label{eq:recall}
\end{equation}

\begin{itemize}
    \item \textbf{F1-Score:} Harmonic mean of precision and recall, offering a balance between the two.
\end{itemize}
\begin{equation}
\text{F1-Score} = \frac{2 \cdot \text{Precision} \cdot \text{Recall}}{\text{Precision} + \text{Recall}}
\label{eq:f1score}
\end{equation}

These metrics provide a clear view of the model’s diagnostic performance, helping clinicians assess its reliability, sensitivity, and safety. Reporting them ensures transparency and builds confidence in its clinical use for accurate and reproducible diagnoses.

Moreover, the receiver operating characteristic (ROC) curve was also employed to evaluate the robustness and discriminative ability of the models. This curve plots the TP rate against the FP rate, providing a visual assessment of classification performance across different thresholds. Each colored line in the ROC graph corresponds to a specific class, with a color legend displayed on the left side of the figure. The Area Under the Curve (AUC) for all classes is nearly 1.00, which is the highest achievable value, while the micro-averaged ROC curve also approaches 1.00. These results indicate that the models achieve excellent class-level and overall classification performance. In this study, all ROC curves were generated using the test sets of the respective datasets to evaluate the predictive capability of each model.

\subsection{Statistical Analysis}

The performance of the three proposed KAN models (SBTAYLOR, SBRBF, and SBWAVELET) was evaluated using several statistical metrics to assess their robustness across diverse medical image datasets \citep{luna2024statds}. Specifically, the KC, MCC, error rate (\(E_r\)), and p‑values were calculated to measure model reliability and statistical significance.

The KC is commonly employed as a reliability or validity measure, indicating how well predicted classes agree with the true classes beyond chance. It is mathematically expressed as:

\begin{equation}
KC = \frac{p_o - p_e}{1 - p_e},
\label{eq:kc}
\end{equation}

where \(p_o\) is the observed agreement and \(p_e\) is the expected agreement by chance.

The MCC is used to evaluate the quality of binary or multiclass classification. It is defined as:

\begin{equation}
MCC = \frac{ TP \cdot TN - FP \cdot FN }{\sqrt{(TP+FP)(TP+FN)(TN+FP)(TN+FN)}},
\label{eq:mcc}
\end{equation}

 Its range is \([-1,1]\), where \(+1\) indicates perfect prediction, \(0\) random prediction, and \(-1\) complete disagreement \citep{smyth2020gastric}.

The error rate (\(E_r\)) measures the probability of misclassification and is given by:

\begin{equation}
E_r = \frac{\text{Number of Misclassified Samples}}{\text{Total Number of Samples}} 
    = \frac{\sum_{i \neq j} C_{ij}}{\sum_{i,j} C_{ij}},
\label{eq:errorrate}
\end{equation}

where \(C_{ij}\) is the element of the confusion matrix corresponding to true class \(i\) and predicted class \(j\).

Lastly, the p‑value was obtained from the Chi‑square (\(\chi^2\)) test of independence to evaluate the statistical significance of the classification results. The test statistic is calculated as:

\begin{equation}
\chi^2 = \sum_{i,j} \frac{(O_{ij} - E_{ij})^2}{E_{ij}},
\label{eq:chisquare}
\end{equation}

where \(O_{ij}\) is the observed frequency in the confusion matrix and \(E_{ij} = \frac{\text{RowSum}_i \cdot \text{ColSum}_j}{N}\) is the expected frequency under the null hypothesis of independence. 
A p‑value less than or equal to \(0.05\) indicates statistical significance, leading to the rejection of the null hypothesis \citep{casella2024statistical, di2020statistical}.

\subsection{KAN Models' Performance Evaluation Using Evaluation Matrices}

The performance evaluation of SBTAYLOR-KAN, SBRBF-KAN, and SBWAVALET-KAN across the four datasets—Brain Tumor (Dataset-1), COVID-19 X-ray (Dataset-2), TB X-ray (Dataset-3), and PAD-UFES-20 (Dataset-4)—demonstrates distinct behaviours under varying class distributions and complexities, as summarized in Table~\ref{tab:kan_performance}. In Dataset-1 (Brain Tumor), which has a relatively balanced class distribution across four tumor types, all models achieve high performance, with SBTAYLOR-KAN slightly leading in overall accuracy (95.09\%) and maintaining a robust F1-score (93.75\%). Dataset-2 (COVID-19 X-ray) shows a similar pattern where SBTAYLOR-KAN consistently excels, achieving the highest accuracy (96.37\%) and F1-score (97.33\%), reflecting its strong ability to handle three-class medical image classification with moderate balance among classes. For Dataset-3 (TB X-ray), which has a simple binary classification but with a larger class imbalance favoring normal samples, all models surpass 98\% accuracy. Here, SBWAVALET-KAN and SBRBF-KAN achieve marginally higher F1-scores (97.50\%), but SBTAYLOR-KAN remains competitive with 96.50\%, showing its adaptability even in slightly skewed data.
The most challenging scenario arises with Dataset-4 (PAD-UFES-20), which is highly imbalanced across six skin lesion categories, including minority classes like MEL, NEV, and SEK. In the original imbalanced setting, all models exhibited a notable decline in performance. The SBRBF-KAN model demonstrated relatively superior results, achieving an accuracy of 56.67\% and an F1-score of 40.00\%. In contrast, the SB-TAYLOR-KAN model achieved a slightly lower accuracy of 54.70\% and an F1-score of 38.00\%. The SBWAVELET-KAN model, however, recorded the lowest performance, with an accuracy of 50.13\% and an F1-score of 28.33\%, particularly in the context of the highly imbalanced dataset.
After balancing the dataset with augmentation, all models show improvement, with SB-TAYLOR-KAN performing the best, achieving an accuracy of 68.22\% and an F1-score of 72.52\%. SBRBF-KAN follows with an accuracy of 61.52\% and an F1-score of 55.60\%, while SBWAVELET-KAN achieves an accuracy of 60.52\% and an F1-score of 55.39\%. This highlights SB-TAYLOR-KAN's superior resilience in handling class imbalance, even in the original, highly imbalanced setting. Consequently, additional experiments will be conducted using this balanced dataset.

\begin{figure}[!t]
    \centering
    \includegraphics[width=0.9\textwidth]{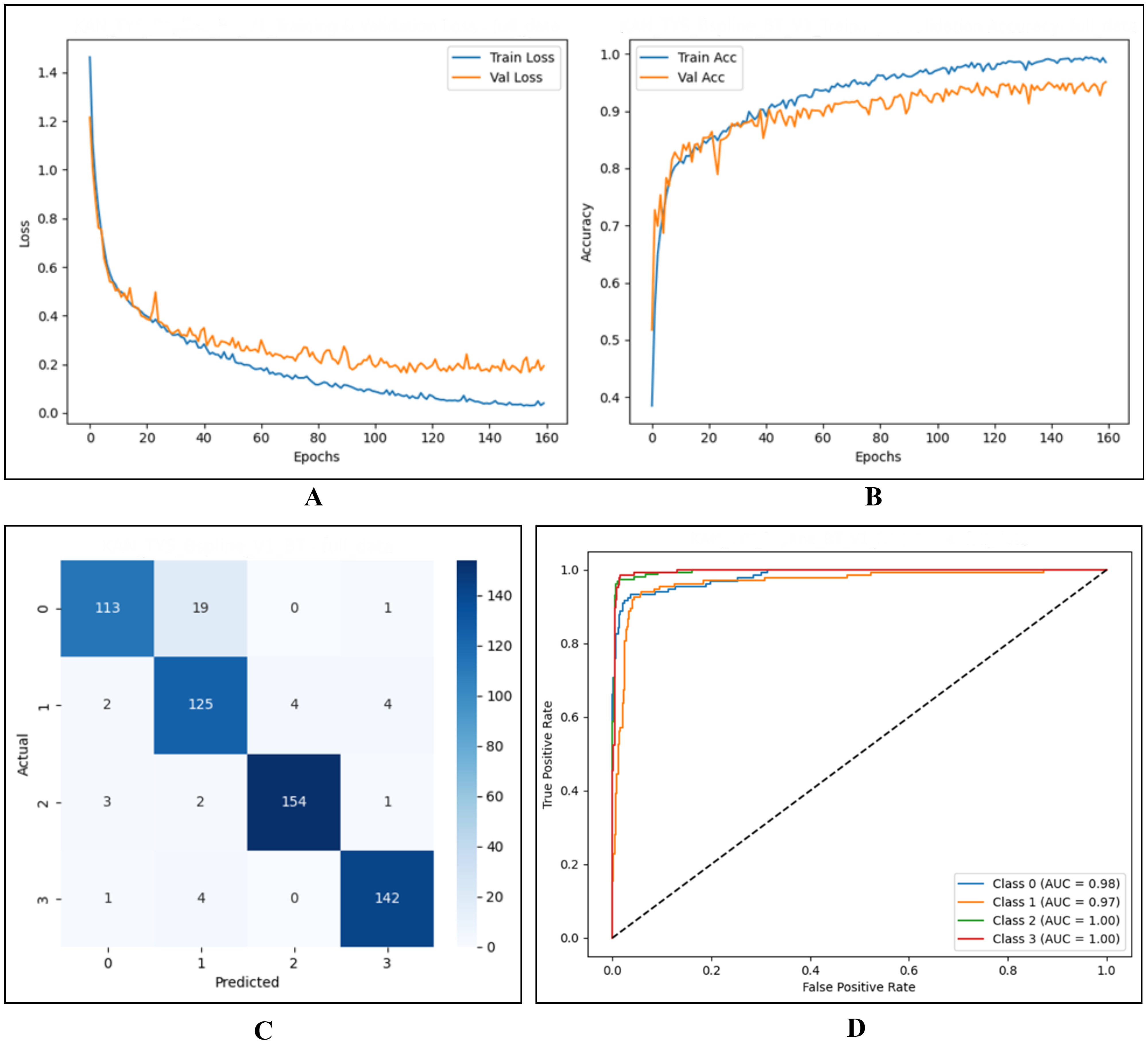}
    \caption{A. Training \& Validation Loss Curve \& B. Training \& Validation Accuracy Curve, C. Confusion Matrix, and D. ROC Curve for SBTAYLOR-KAN on Dataset-1 (Brain Tumor)}
    \label{fig:TYS_BT1}
\end{figure}

\begin{figure}[!t]
    \centering
    \includegraphics[width=0.9\textwidth]{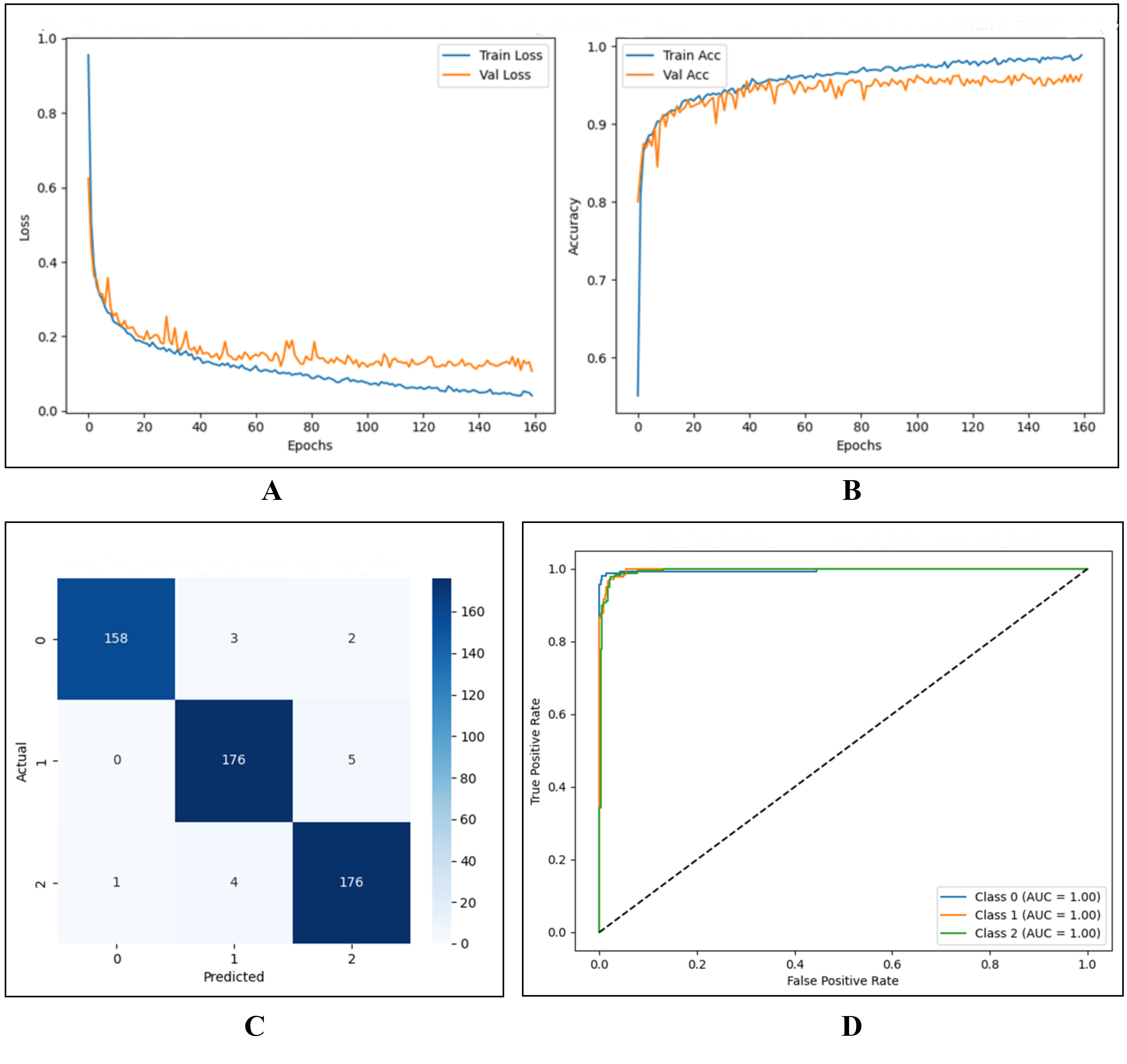}
    \caption{A. Training \& Validation Loss Curve \& B. Training \& Validation Accuracy Curve, C. Confusion Matrix, and D. ROC Curve for SBTAYLOR-KAN on Dataset-2 (COVID-19 X-ray)}
    \label{fig:TYS_ChestXray1}
\end{figure}

\begin{figure}[!t]
    \centering
    \includegraphics[width=0.9\textwidth]{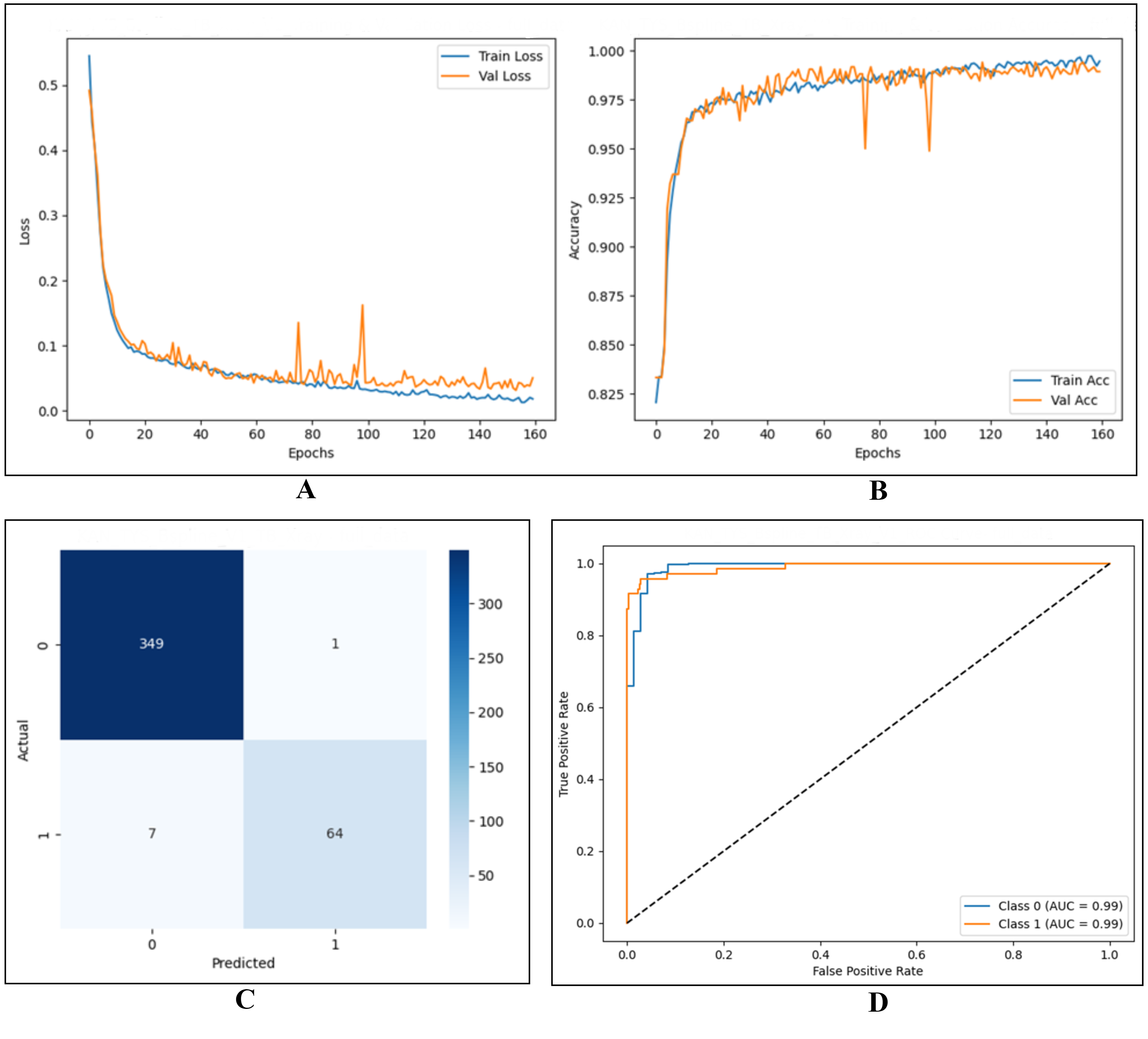}
    \caption{A. Training \& Validation Loss Curve \& B. Training \& Validation Accuracy Curve, C. Confusion Matrix, and D. ROC Curve for SBTAYLOR-KAN on Dataset-3 (TB X-ray)}
    \label{fig:TYS_TBXray1}
\end{figure}

\begin{figure}[!t]
    \centering
    \includegraphics[width=0.9\textwidth]{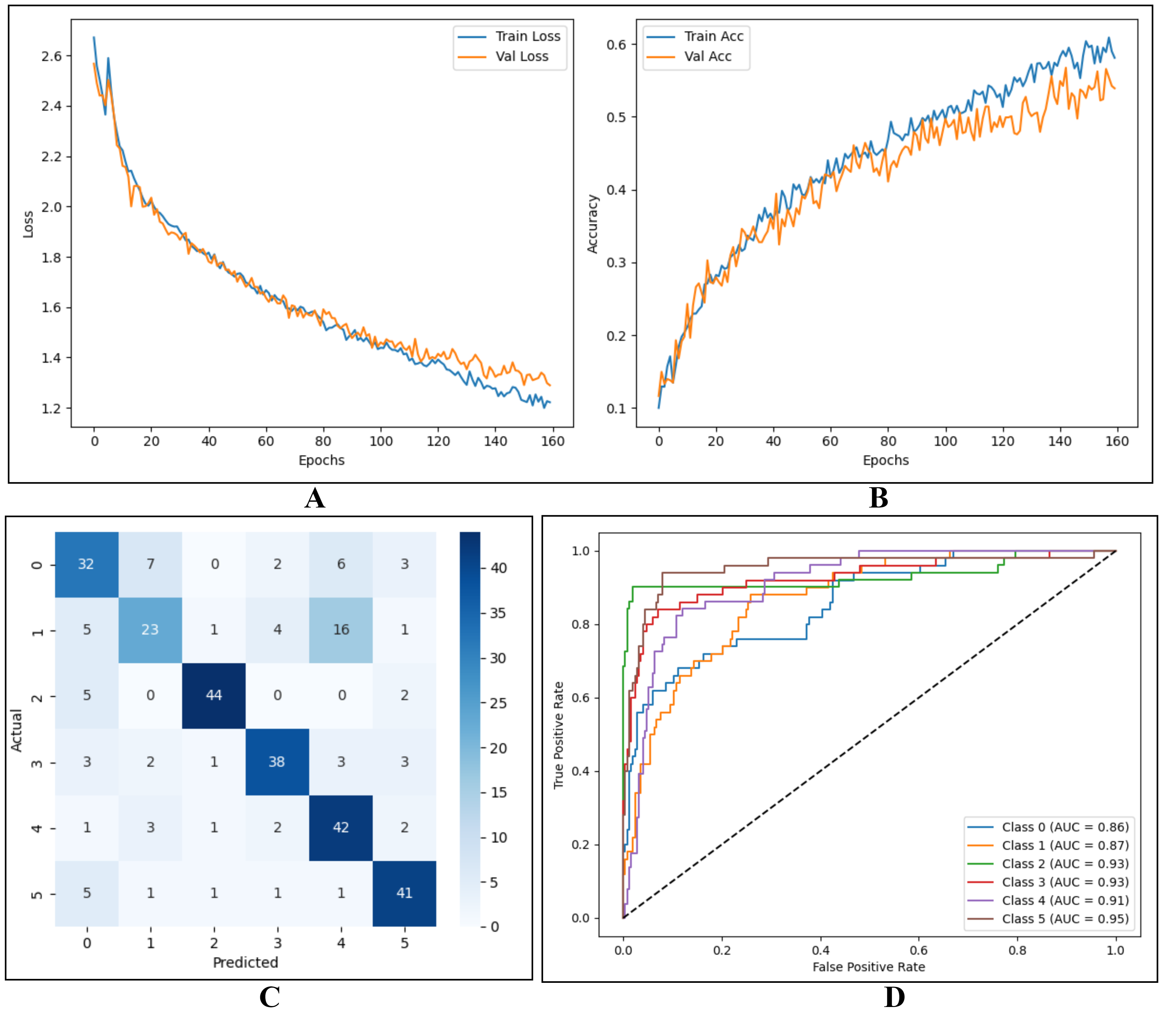}
    \caption{A. Training \& Validation Loss Curve \& B. Training \& Validation Accuracy Curve, C. Confusion Matrix, and D. ROC Curve for SBTAYLOR-KAN on Dataset-4 (PAD-UFES-20)}
    \label{fig:TYS_Skin1}
\end{figure}

\begin{figure}[!t]
    \centering
    \includegraphics[width=0.9\textwidth]{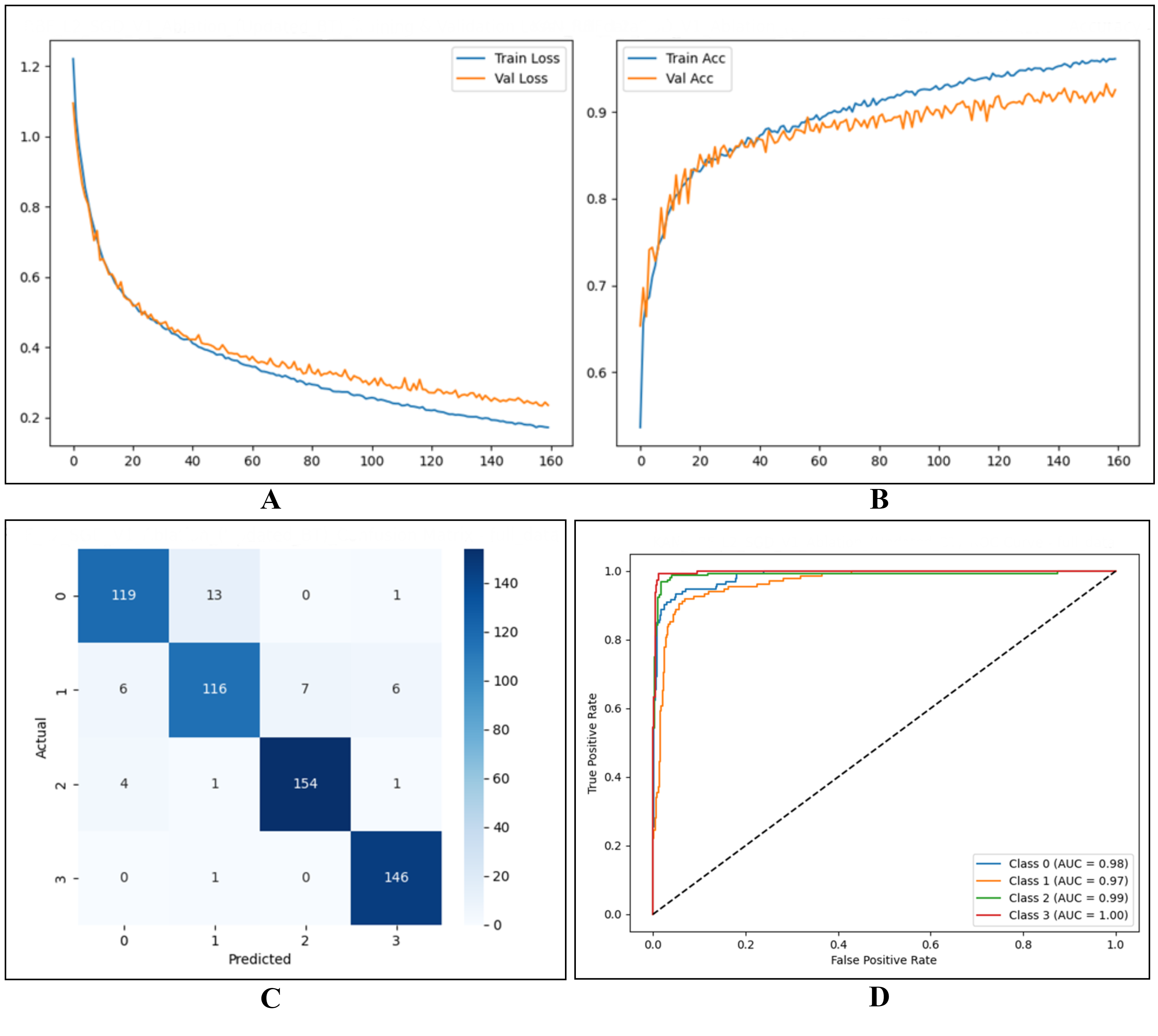}
    \caption{A. Training \& Validation Loss Curve \& B. Training \& Validation Accuracy Curve, C. Confusion Matrix, and D. ROC Curve for SBRBF-KAN on Dataset-1 (Brain Tumor)}
    \label{fig:RBF_BT1}
\end{figure}

\begin{figure}[!t]
    \centering
    \includegraphics[width=0.9\textwidth]{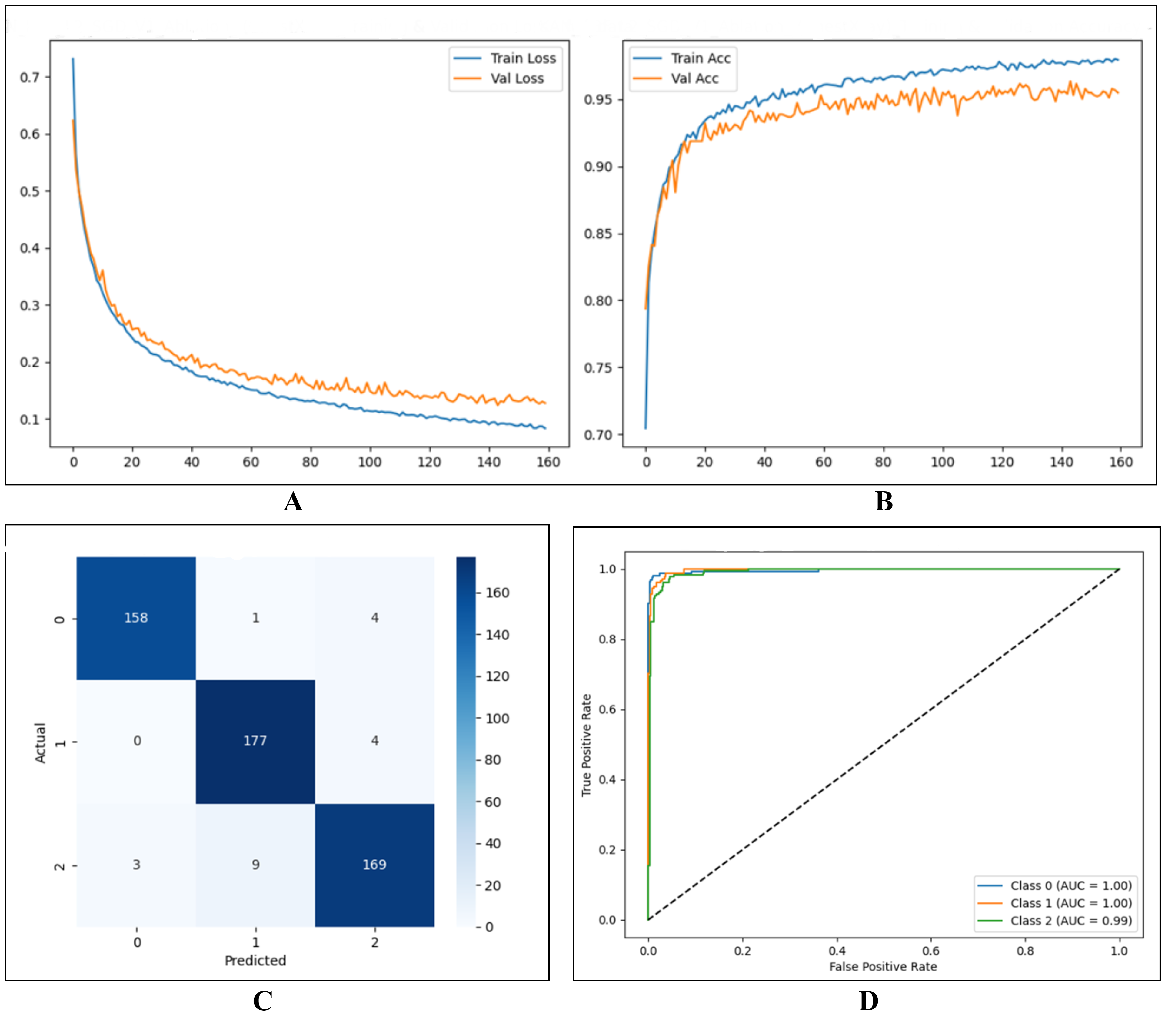}
    \caption{A. Training \& Validation Loss Curve \& B. Training \& Validation Accuracy Curve, C. Confusion Matrix, and D. ROC Curve for SBRBF-KAN on Dataset-2 (COVID-19 X-ray)}
    \label{fig:RBF_ChestXray1}
\end{figure}

\begin{figure}[!t]
    \centering
    \includegraphics[width=0.9\textwidth]{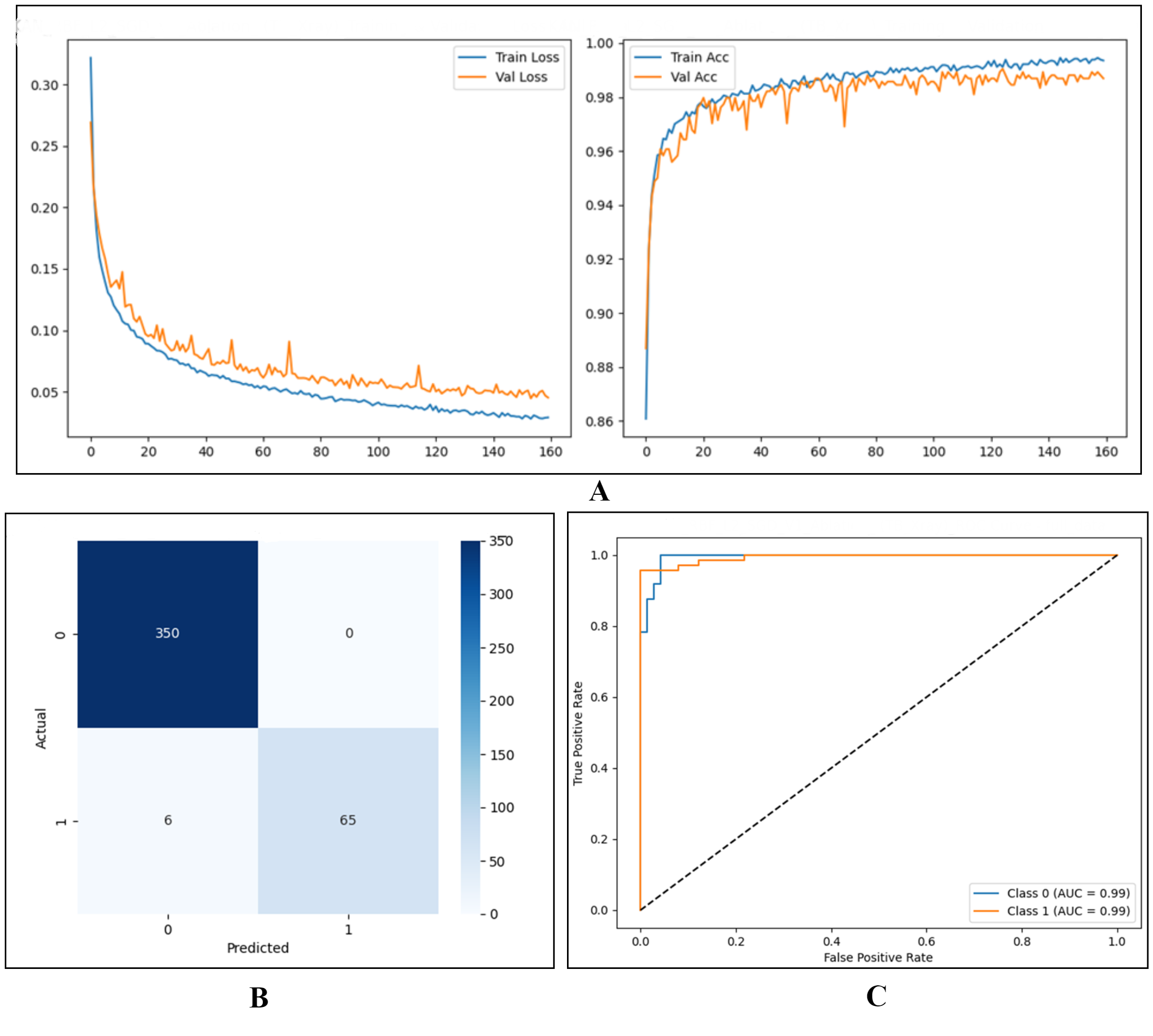}
    \caption{A. Training \& Validation Loss Curve \& B. Training \& Validation Accuracy Curve, C. Confusion Matrix, and D. ROC Curve for SBRBF-KAN on Dataset-3 (TB X-ray)}
    \label{fig:RBF_TBXray1}
\end{figure}

\begin{figure}[!t]
    \centering
    \includegraphics[width=0.9\textwidth]{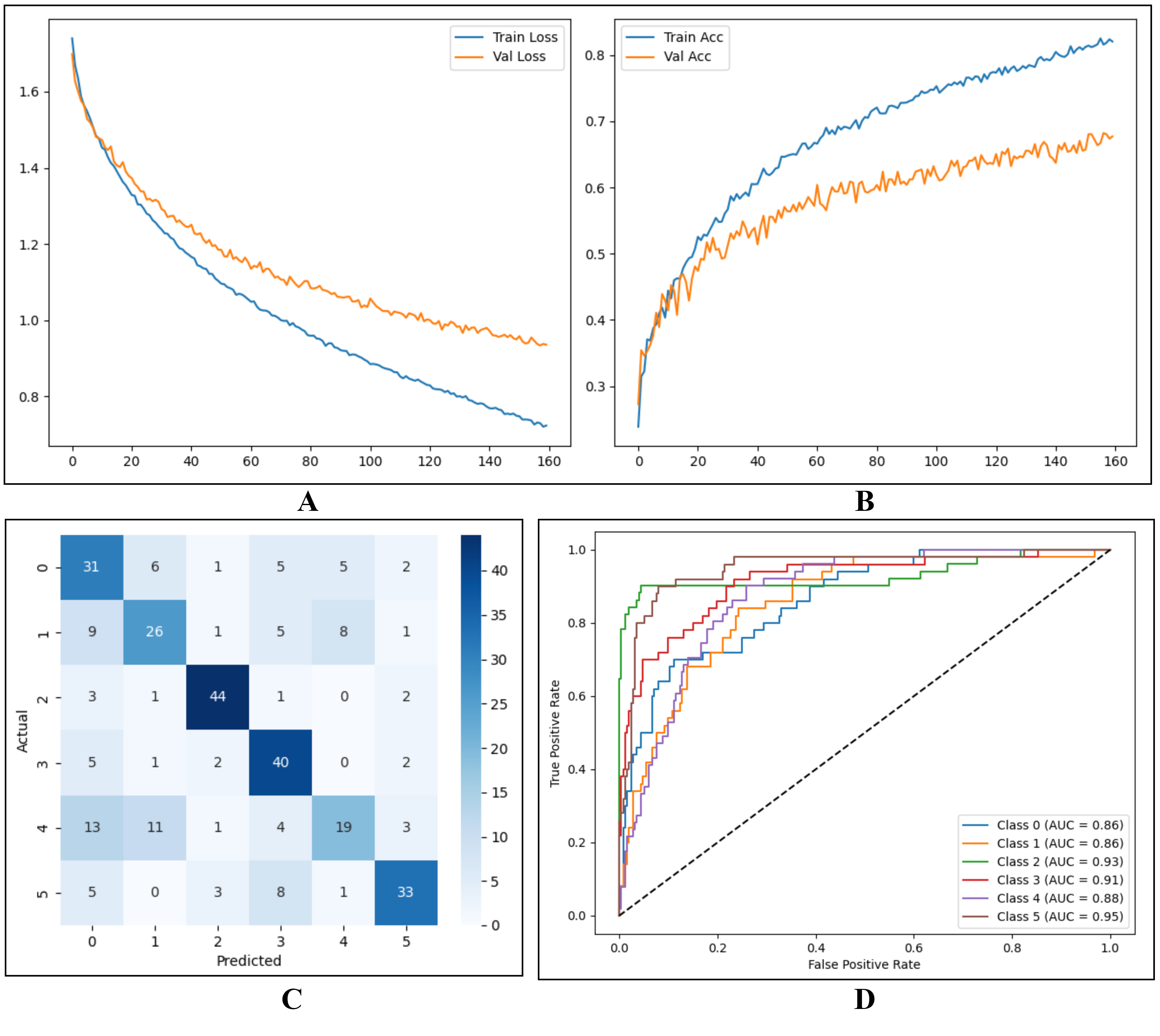}
    \caption{A. Training \& Validation Loss Curve \& B. Training \& Validation Accuracy Curve, C. Confusion Matrix, and D. ROC Curve for SBRBF-KAN on Dataset-4 (PAD-UFES-20)}
    \label{fig:RBF_Skin1}
\end{figure}

\begin{figure}[!t]
    \centering
    \includegraphics[width=0.9\textwidth]{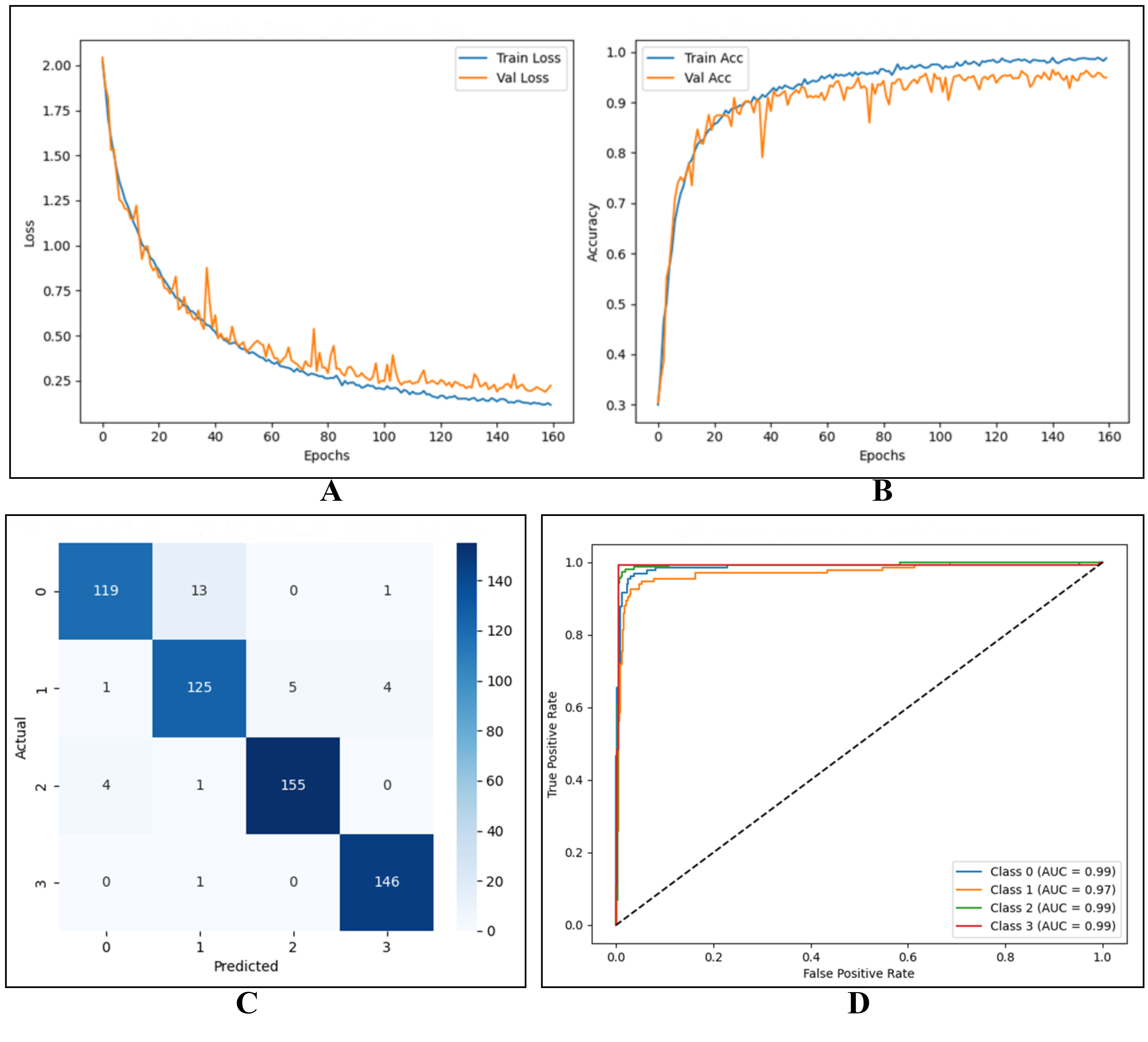}
    \caption{A. Training \& Validation Loss Curve \& B. Training \& Validation Accuracy Curve, C. Confusion Matrix, and D. ROC Curve for SBWAVELET-KAN on Dataset-1 (Brain Tumor)}
    \label{fig:WAVE_BT1}
\end{figure}

\begin{figure}[!t]
    \centering
    \includegraphics[width=0.9\textwidth]{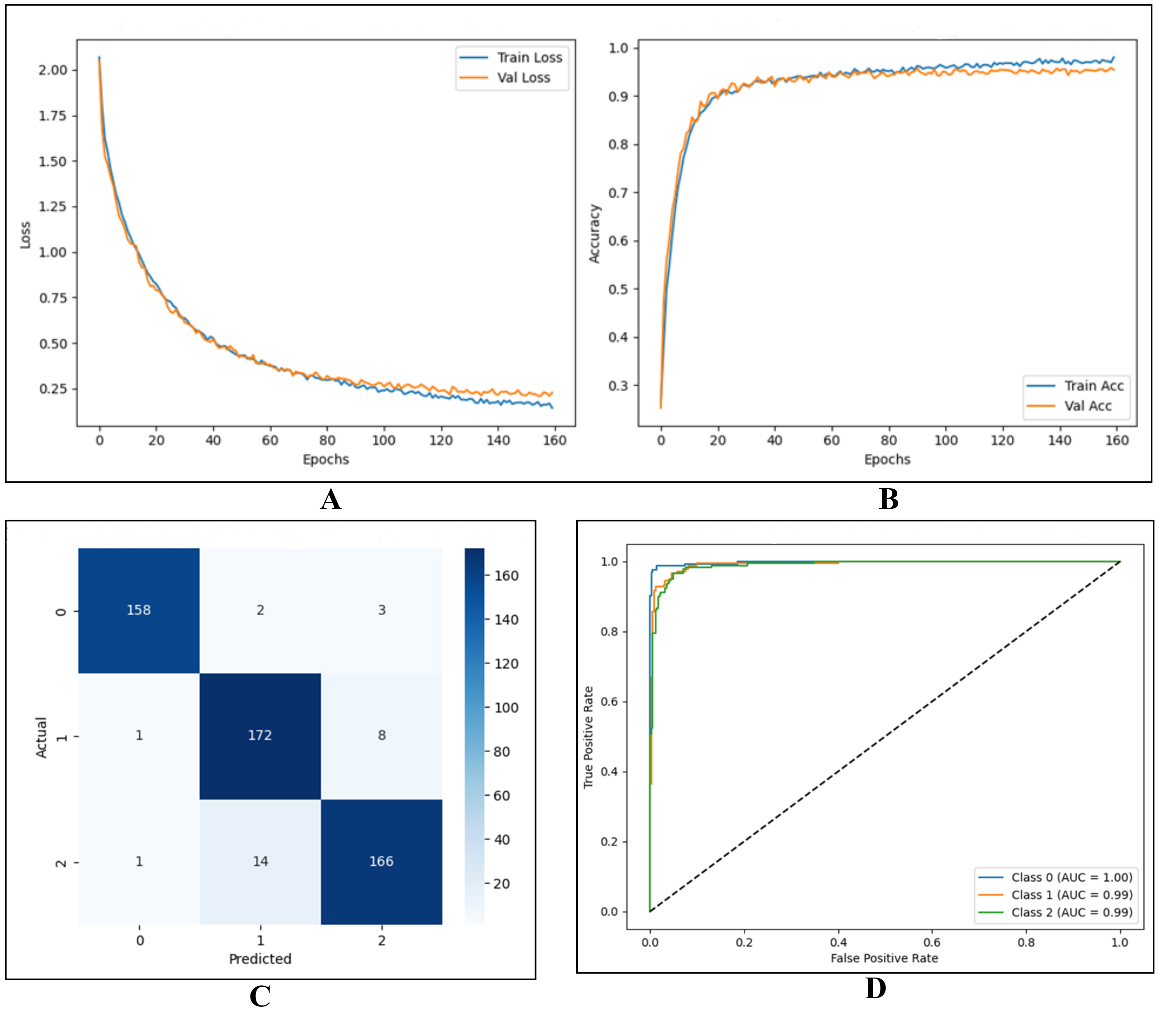}
    \caption{A. Training \& Validation Loss Curve \& B. Training \& Validation Accuracy Curve, C. Confusion Matrix, and D. ROC Curve for  SBWAVELET-KAN on Dataset-2 (COVID-19 X-ray)}
    \label{fig:WAVE_ChestXray1}
\end{figure}

\begin{figure}[!t]
    \centering
    \includegraphics[width=0.9\textwidth]{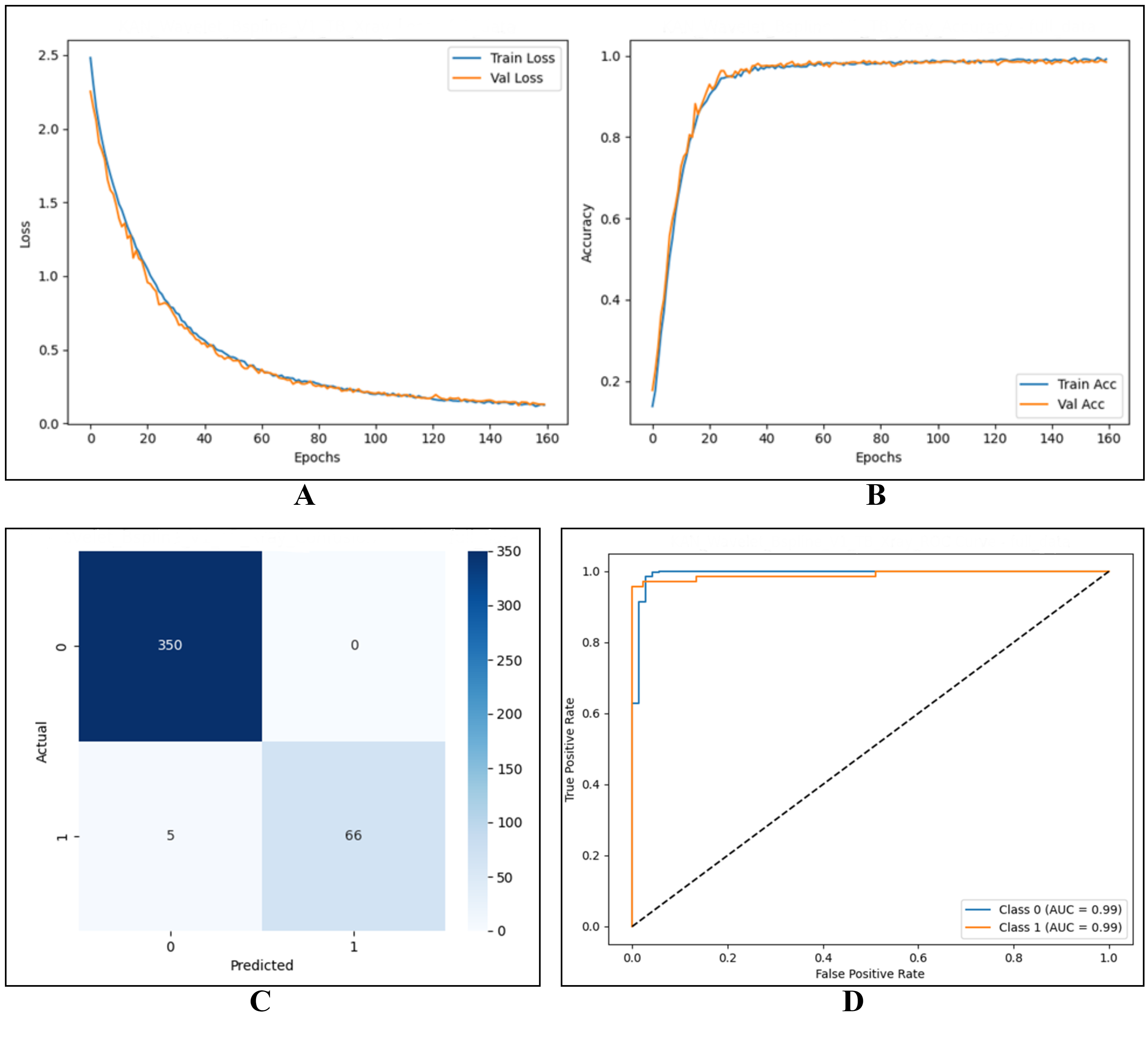}
    \caption{A. Training \& Validation Loss Curve \& B. Training \& Validation Accuracy Curve, C. Confusion Matrix, and D. ROC Curve for  SBWAVELET-KAN on Dataset-3 (TB X-ray)}
    \label{fig:WAVE_TBXray1}
\end{figure}

\begin{figure}[!t]
    \centering
    \includegraphics[width=0.9\textwidth]{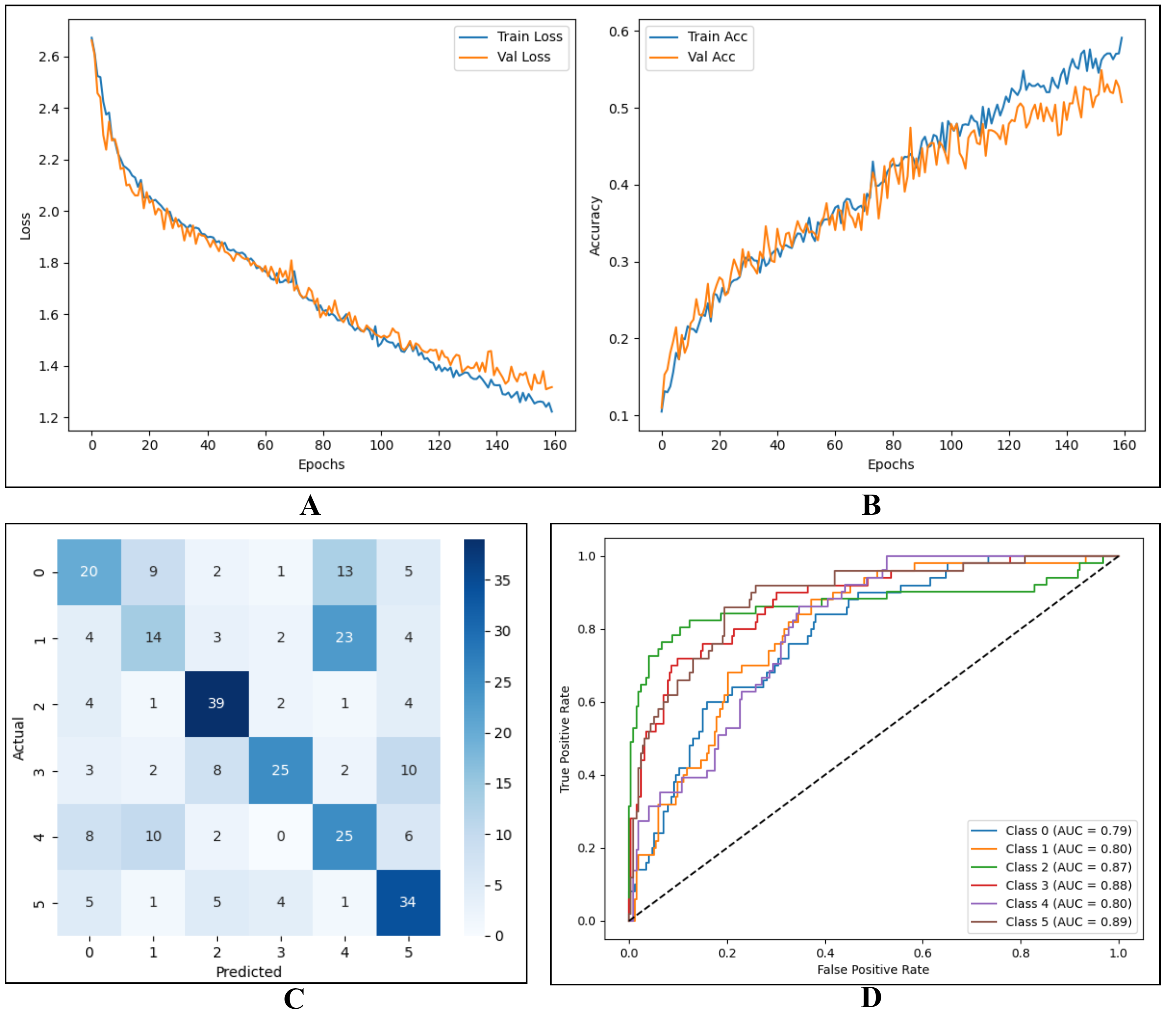}
    \caption{A. Training \& Validation Loss Curve \& B. Training \& Validation Accuracy Curve, C. Confusion Matrix, and D. ROC Curve for  SBWAVELET-KAN on Dataset-4 (PAD-UFES-20)}
    \label{fig:WAVE_Skin1}
\end{figure}

In conclusion, the analysis across all four datasets demonstrates the robustness and adaptability of SBTAYLOR-KAN. The model performs consistently well on balanced and moderately imbalanced datasets (Datasets 1–3) and remains competitive even under extreme imbalance (Dataset 4). Notably, when we experimented with a smaller but balanced version of the skin disease dataset (Dataset-4), the model's performance significantly improved, highlighting its ability to excel when class distributions are more balanced. This highlights SBTAYLOR-KAN's effectiveness in medical imaging tasks, where both class imbalance and data heterogeneity are frequent challenges. Our findings suggest that, when working with a smaller but balanced dataset, the model can perform even better, demonstrating its ability to handle more even class distributions. For a visual summary of model performance, including loss/accuracy curves, confusion matrices, and ROC curves, see Figures~\ref{fig:TYS_BT1}–\ref{fig:WAVE_Skin1}.

\begin{table}[htbp]
\centering
\caption{Performance Metrics for Three KAN Models on Four Distinct Datasets}
\vspace{1em} 
\label{tab:kan_performance}
\small  
\renewcommand{\arraystretch}{0.9}  

\resizebox{\textwidth}{!}{  
\begin{tabular}{|@{\hskip 5pt}l@{\hskip 5pt}|c|c|c|c|c|}
\hline
\textbf{Dataset} & \textbf{Model} & \textbf{ACC} & \textbf{Precision} & \textbf{Recall} & \textbf{F1-score} \\ 
\textbf{} & & \textbf{(\%)} & \textbf{(\%)} & \textbf{(\%)} & \textbf{(\%)} \\
\hline
\multirow{3}{*}{Dataset-1} & SBTAYLOR-KAN & 95.09 & 93.75 & 93.75 & 93.75 \\
& SBRBF-KAN & 93.68 & 93.00 & 93.00 & 93.00 \\
& SBWAVELET-KAN & 94.91 & 94.75 & 94.50 & 94.62 \\
\hline
\multirow{3}{*}{Dataset-2} & SBTAYLOR-KAN & 96.37 & 97.00 & 97.00 & 97.33 \\
& SBRBF-KAN & 95.51 & 96.00 & 96.00 & 96.00 \\
& SBWAVELET-KAN & 95.41 & 94.67 & 94.67 & 94.67 \\
\hline
\multirow{3}{*}{Dataset-3} & SBTAYLOR-KAN & 98.93 & 98.00 & 95.00 & 96.50 \\
& SBRBF-KAN & 98.69 & 99.00 & 96.00 & 97.50 \\
& SBWAVELET-KAN & 98.45 & 99.50 & 96.50 & 97.50 \\
\hline
\multirow{3}{*}{Dataset-4} & SBTAYLOR-KAN (Original) & 54.70 & 36.16 & 40.00 & 38.00 \\
& SBRBF-KAN (Original) & 56.67 & 38.00 & 41.50 & 40.00 \\
& SBWAVELET-KAN (Original) & 50.13 & 28.50 & 30.00 & 28.33 \\
& SBTAYLOR-KAN (Balanced) & 68.22 & 73.29 & 72.77 & 72.52 \\
& SBRBF-KAN (Balanced) & 67.72 & 64.84 & 65.25 & 64.64 \\
& SBWAVELET-KAN (Balanced) & 60.52 & 58.50 & 52.60 & 55.39 \\
\hline
\end{tabular}
}
\end{table}



\subsection{KAN Models' Performance Evaluation Using Statistical Matrices}

Table~\ref{tab:kc_analysis} represents the statistical comparison of the three KAN models—SBTAYLOR-KAN, SBRBF-KAN, and SBWAVALET-KAN—across four datasets, showing clear differences in their predictive performance and reliability.
For Dataset-1, SBWAVALET-KAN demonstrates the strongest performance, achieving the highest KC (0.9302) and MCC (0.9307), along with the lowest error rate (0.0521), suggesting superior classification accuracy. However, a shift is observed in Dataset-2, where SBTAYLOR-KAN outperforms the other models with a KC of 0.9570, MCC of 0.9571, and a minimal error rate of 0.0285, indicating robust generalization capability. In Dataset-3, all three models perform exceptionally well. Still, SBRBF-KAN slightly leads with a KC/MCC of 0.9830/0.9831 and the lowest error rate of 0.0084, while SBTAYLOR-KAN remains highly competitive with a similar error rate. Dataset-4 presents the most challenging scenario, with all models showing a noticeable drop in performance. SBTAYLOR-KAN leads with a KC of 0.5772, MCC of 0.5795, and an error rate of 0.3523, while SBRBF-KAN follows closely with a KC of 0.5043 and an error rate of 0.3711. SBWAVELET-KAN shows the lowest performance in this dataset, with a KC of 0.4237, MCC of 0.4251, and the highest error rate of 0.4801. Despite these lower results, all models show statistical significance ($p < 0.05$). Overall, SBTAYLOR-KAN delivers balanced and reliable performance across diverse datasets. Its low error rates, robustness, and adaptability make it a practical choice for applications requiring consistent results on heterogeneous data.

\begin{table*}[htbp]
\centering
\caption{Statistical Analysis of Kappa Coefficient (KC), Matthews Correlation Coefficient (MCC), Error Rate, and p-Values for Three KAN Models Across Four Datasets}
\vspace{1em} 
\label{tab:kc_analysis}
\resizebox{0.8\textwidth}{!}{
\begin{tabular}{|l|l|c|c|c|c|}
\hline
\textbf{Dataset} & \textbf{Model} & \textbf{KC} & \textbf{MCC} & \textbf{Error Rate} & \textbf{p-Value} \\
\hline

\multirow{3}{*}{Dataset-1} 
    & SBTAYLOR-KAN & 0.8335 & 0.8354 & 0.1245 & $<0.05$ \\
    & SBRBF-KAN & 0.9069 & 0.9077 & 0.0695 & $<0.05$ \\
    & SBWAVELET-KAN & 0.9302 & 0.9307 & 0.0521 & $<0.05$ \\
\hline

\multirow{3}{*}{Dataset-2} 
    & SBTAYLOR-KAN & 0.9570 & 0.9571 & 0.0285 & $<0.05$ \\
    & SBRBF-KAN & 0.8845 & 0.8850 & 0.0769 & $<0.05$ \\
    & SBWAVELET-KAN & 0.8429 & 0.8442 & 0.1046 & $<0.05$ \\
\hline

\multirow{3}{*}{Dataset-3} 
    & SBTAYLOR-KAN & 0.9773 & 0.9774 & 0.0113 & $<0.05$ \\
    & SBRBF-KAN & 0.9830 & 0.9831 & 0.0084 & $<0.05$ \\
    & SBWAVELET-KAN & 0.9564 & 0.9573 & 0.0118 & $<0.05$ \\
\hline

\multirow{3}{*}{Dataset-4} 
    & SBTAYLOR-KAN & 0.5772 & 0.5795 & 0.3523 & $<0.05$ \\
    & SBRBF-KAN & 0.5043 & 0.5065 & 0.3711 & $<0.05$ \\
    & SBWAVELET-KAN & 0.4237 & 0.4251 & 0.4801 & $<0.05$ \\
\hline
\end{tabular}
} 
\end{table*}

\subsection{ Quantitative Analysis of KAN Models}
To evaluate the three KAN models under different levels of data reduction, we conducted experiments on four medical imaging datasets: Brain Tumor, COVID-19 Chest X-Ray, TB X-Ray, and Skin Cancer (PAD-UFES-20). Tables~\ref{tab:brain_tumor_results}--\ref{tab:PAD-UFES-20_results} present the accuracy, precision, recall, and F1-score for each dataset at varying data retention levels.

\subsubsection{ Quantitative Analysis of KAN Models on Dataset-1 (Brain Tumor)}
Table \ref{tab:brain_tumor_results} presents the experimental outcomes of the KAN models on the brain tumor dataset, highlighting distinct performance patterns under various levels of data reduction.
The SBTAYLOR-KAN model consistently performs consistently, maintaining high accuracy alongside balanced precision, recall, and F1-scores across all data reduction levels. Its performance initiates at 95.09\% accuracy with the full dataset (100\%) and remains above 86\% accuracy even when the dataset is reduced to 30\%, signifying its strong resilience to data reduction in comparison to the other models.

Conversely, the SBRBF-KAN model shows a more gradual but consistent decline in performance as the data is reduced. In contrast, SBWAVELET-KAN exhibits highly fluctuating behavior, with a sharp decline at certain reduction points---most notably at 65\% of the data, where its accuracy drops to 72.28\%, which is significantly lower than SBTAYLOR-KAN at the same level. Interestingly, SBWAVELET-KAN demonstrates partial recovery at 60\% and 50\% data levels, highlighting its sensitivity to the composition of the remaining dataset.

Overall, the findings show that SBTAYLOR-KAN is the most robust and reliable model under progressive data reduction. Its optimal operational range lies between 70--90\% data usage, where it consistently maintains 92--94\% accuracy with stable F1-scores, ensuring dependable classification performance without notable degradation. When data usage drops below 50\%, performance variability increases across all models, with SBWAVELET-KAN being particularly unstable. This indicates that extremely low data usage is not suitable for maintaining reliable performance.

\subsubsection{ Quantitative Analysis of KAN Models on Dataset-2 (COVID-19 Xray)}
As summarized in Table~\ref{tab:COVID_19_Xray}, the SBTAYLOR-KAN model exhibits exceptional resilience to data reduction, maintaining high accuracy and balanced performance metrics even as the dataset size decreases. At full data availability (100\%), it achieves the highest F1-score of 97.33\%, slightly ahead of SBRBF-KAN and notably outperforming SBWAVELET-KAN. As data is gradually reduced, SBTAYLOR-KAN shows only a modest decline in accuracy, staying above 92\% until 30\% data usage. In contrast, SBWAVELET-KAN exhibits sharper drops, with accuracy falling to 91.59\% at 35\% data usage, indicating greater sensitivity to data scarcity. SBRBF-KAN occasionally surpasses SBTAYLOR-KAN at moderate reductions (95--85\%), but lacks the same stability across severe reductions. Notably, SBTAYLOR-KAN sustains F1-scores around 94--95\% within the 85--60\% range, which makes this interval the most reliable and cost-effective zone for training without significant loss of diagnostic accuracy. This consistent robustness suggests that SBTAYLOR-KAN is the preferred choice for scenarios where dataset size is constrained, offering both reliability and efficiency under limited data conditions.

\subsubsection{ Quantitative Analysis of KAN Models on Dataset-3 (TB X-ray)}
As illustrated in Table~\ref{tab:TB_Xray_results}, the SBTAYLOR-KAN model demonstrates remarkable robustness under progressive data reduction, maintaining accuracy above 96\% even when the dataset is reduced to 30\%. Compared to SBRBF-KAN and SBWAVELET-KAN, the TAYLOR variant exhibits a smoother performance decay, with minimal fluctuations in F1-score and recall. In contrast, SBWAVELET-KAN is highly sensitive to reduction, showing a sharp performance drop at 85\% data (F1-score $\approx$ 79.50\%), while SBRBF-KAN occasionally loses recall under aggressive reductions. Notably, SBTAYLOR-KAN sustains competitive performance across all metrics until approximately 30-40\% of the original images, indicating strong generalization under constrained data scenarios. Overall, for reliable clinical-grade outcomes, a reduction range between 50\% and 100\% is preferable, as all three models perform optimally; however, SBTAYLOR-KAN proves to be the most stable and trustworthy model when further reducing the dataset.

\subsubsection{ Quantitative Analysis of KAN Models on Dataset-4 (PAD-UFES-20)}
The quantitative analysis of the balanced PAD-UFES-20 dataset, presented in Table~\ref{tab:PAD-UFES-20_results}, shows that the SBTAYLOR-KAN model consistently performs well across various data reductions. At 100\% data, it achieves an accuracy of 68.22\%, precision of 73.29\%, recall of 72.77\%, and an F1-score of 72.51\%. Even with a reduction to 70\% (1,470 samples), it maintains 57.40\% accuracy and an F1-score of 57.42\%, demonstrating its stability. In contrast, the SBRBF-KAN model shows more variability, with a higher accuracy of 64.39\% at 95\% data, but less consistent precision and recall. SBWAVELET-KAN performs poorly under data reduction, especially at 20\%, where accuracy drops to 46.61\% and F1-score to 27.00\%. Overall, the SBTAYLOR-KAN model is the most reliable, particularly in the 55-65\% data range, offering an optimal balance between data efficiency and predictive performance.

\subsection{KAN Models' Performance Evaluation on Isolated and Distinct Data within the Same Category}
The performance of the proposed KAN models on distinct and isolated datasets is presented in Table~\ref{tab:model_performance}. Among the three models, SBTAYLOR-KAN consistently demonstrates stronger generalization, maintaining high test accuracy across both previous and new datasets. On Test Dataset-1, it achieved balanced results with 95.09\% (P) and 94.45\% (N), while on Dataset-2, it outperformed others with 96.37\% (P) and 90.00\% (N). For Test Dataset-3, it retained 89.65\% accuracy compared to sharper drops in SBRBF-KAN (85.00\%) and SBWAVELET-KAN (84.70\%). 
A significant decline in performance was observed across all models on New Test Dataset-4, likely due to its distinct sources and complex image patterns. However, SBTAYLOR-KAN outperformed the others with a new accuracy of 61.70\%, while SBWAVELET-KAN dropped sharply to 55.85\%. These results indicate that SBTAYLOR-KAN remains the most reliable and stable model when transitioning from known to unseen test data.

\begin{table}[htbp]
\centering
\caption{Model performance evaluation with new and unseen data across various datasets for different models (SBTAYLOR-KAN, SBRBF-KAN, SBWAVELET-KAN). The table shows both the previous (P) test accuracy and the new (N) testing accuracy.}
\vspace{1em} 
\label{tab:model_performance}
\begin{adjustbox}{valign=c, width=\textwidth}  
\begin{tabular}{|c|c|c|c|c|c|}
\hline
\textbf{Dataset} & \textbf{Model} & \multicolumn{1}{c|}{\textbf{Test ACC (P)}} & \multicolumn{1}{c|}{\textbf{Test ACC (N) }} \\ \hline
\multirow{3}{*}{Test Dataset-1} & SBTAYLOR-KAN & 95.09\% & 94.45\% \\
& SBRBF-KAN & 93.68\% & 93.00\% \\
& SBWAVELET-KAN & 94.91\% & 94.00\% \\ \hline
\multirow{3}{*}{Test Dataset-2} & SBTAYLOR-KAN & 96.37\% & 90.00\% \\
& SBRBF-KAN & 95.51\% & 88.00\% \\
& SBWAVELET-KAN & 95.41\% & 87.00\% \\ \hline
\multirow{3}{*}{Test Dataset-3} & SBTAYLOR-KAN & 98.93\% & 89.65\% \\
& SBRBF-KAN & 98.69\% & 85.00\% \\
& SBWAVELET-KAN & 98.45\% & 84.70\% \\ \hline
\multirow{3}{*}{Test Dataset-4} & SBTAYLOR-KAN & 68.22\% & 61.70\% \\
& SBRBF-KAN & 67.72\% & 60.65\% \\
& SBWAVELET-KAN & 60.52\% & 55.85\% \\ \hline
\end{tabular}
\end{adjustbox}
\end{table}

\subsection{Performance Comparison of the State-of-Art Models}
In this section, state-of-the-art models were evaluated on four datasets: brain tumor (Dataset-1), COVID-19 X-ray (Dataset-2), TB X-ray (Dataset-3), and the balanced PAD-UFES-20 (Dataset-4), utilizing the same model configurations (0.0001 learning rate, SGD Optimizer, SiLU activation function, and 32 batch size). The performance comparison, summarized in Table~\ref{tab:performance_metrics}, highlights the varying strengths of InceptionV3 \citep{szegedy2016rethinking}, ResNet50 \citep{he2016deep}, MobileNetV2 \citep{mobilenetv2}, and VGG16 \citep{simonyan2014very} across these datasets. 

 On Dataset-1, InceptionV3 achieved the best performance with an accuracy of 82.78\%, an F1-score of 82.00\%, and an AUC of 95.00\%, outperforming ResNet50 and significantly surpassing the lower-performing MobileNetV2 and VGG16. In Dataset-2, VGG16 led with the highest accuracy (94.56\%), F1-score (94.45\%), and AUC (95.78\%), followed closely by InceptionV3, while MobileNetV2 lagged. Dataset-3 results demonstrate that both VGG16 and InceptionV3 excelled, with VGG16 slightly outperforming in AUC (97.50\%) and InceptionV3 achieving a marginally higher F1-score (96.98\%). In contrast, Dataset-4 presented challenges for all models, showing markedly lower performance, where MobileNetV2 obtained the highest accuracy (58.24\%), F1-score (46.50\%), and the highest AUC (75.00\%). Overall, the findings suggest that model performance varies considerably across datasets, exhibiting inconsistent results.

\begin{table}[htbp]
\centering
\caption{Performance Metrics (ACC, F1-score, AUC) for Different State-of-Art Models on Four Datasets}
\vspace{1em} 
\label{tab:performance_metrics}
\small  
\begin{adjustbox}{valign=c, width=\textwidth}  
\begin{tabular}{|@{\hskip 2pt}l@{\hskip 2pt}|@{\hskip 2pt}c@{\hskip 2pt}|@{\hskip 2pt}c@{\hskip 2pt}|@{\hskip 2pt}c@{\hskip 2pt}|@{\hskip 2pt}c@{\hskip 2pt}|}
\hline
\textbf{Dataset} & \textbf{Model} & \textbf{ACC} (\%) & \textbf{F1-score} (\%) & \textbf{AUC} (\%) \\
\hline
\multirow{4}{*}{Dataset-1} & InceptionV3 & 82.78 & 82.00 & 95.00 \\
& ResNet50 & 75.99 & 75.00 & 93.00 \\
& MobileNetV2 & 58.24 & 63.38 & 88.25 \\
& VGG16 & 46.60 & 31.54 & 66.50 \\
\hline
\multirow{4}{*}{Dataset-2} & InceptionV3 & 91.04 & 91.00 & 94.33 \\
& ResNet50 & 85.67 & 83.29 & 85.87 \\
& MobileNetV2 & 75.34 & 73.06 & 78.68 \\
& VGG16 & 94.56 & 94.45 & 95.78 \\
\hline
\multirow{4}{*}{Dataset-3} & InceptionV3 & 96.78 & 96.98 & 96.00 \\
& ResNet50 & 87.41 & 74.00 & 73.50 \\
& MobileNetV2 & 83.21 & 45.65 & 50.00 \\
& VGG16 & 96.86 & 95.60 & 97.50 \\
\hline
\multirow{4}{*}{Dataset-4} & InceptionV3  & 52.55 & 41.38 & 64.16 \\
& ResNet50 & 53.82 & 41.83 & 65.50 \\
& MobileNetV2 & 58.24 & 46.50 & 75.00 \\
& VGG16 & 57.23 & 48.66 & 60.16 \\
\hline
\end{tabular}
\end{adjustbox}
\end{table}

\subsection{Evaluation on MedMNIST Datasets} 
The primary objective of this experiment was to evaluate how our proposed models perform on the MedMNIST dataset images and to compare their performance with existing and recent state-of-the-art methods. For a fair comparison, all of our models were retrained and tested using the official test image data. The results of this evaluation are summarized in Table~\ref{MedMNIST}.

The performance of our proposed models—SBTAYLOR-KAN, SBRBF-KAN, and SBWAVELET-KAN—was systematically evaluated on the BreastMNIST and PneumoniaMNIST datasets. On BreastMNIST, our SBTAYLOR-KAN model achieved a test accuracy of 89.38\% and an AUC of 94.00\%, closely matching the performance of leading models such as MedKAN-S (90.40\% ACC, 94.20\% AUC) and MedVII-S (89.70\% ACC, 93.80\% AUC). In contrast, SBRBF-KAN and SBWAVELET-KAN attained 83.80\%/89.90\% and 77.80\%/65.50\% (ACC/AUC), respectively, showing lower performance yet confirming the overall competitiveness of our KAN-based approaches.
On PneumoniaMNIST, SBTAYLOR-KAN demonstrated strong generalization, achieving 95.39\% accuracy and an impressive 99.89\% AUC, surpassing most existing models and approaching the performance of MedKAN-B (96.30\% ACC, 99.60\% AUC). Meanwhile, SBRBF-KAN achieved 95.06\% ACC and 98.00\% AUC, and SBWAVELET-KAN achieved 91.45\% ACC and 94.50\% AUC, both confirming the effectiveness of our KAN-based designs across different datasets.
In summary, these results clearly illustrate that our lightweight KAN models, particularly SBTAYLOR-KAN, can deliver near state-of-the-art performance while using a fraction of the parameters, highlighting their efficiency and potential for practical deployment in medical imaging tasks.

\begin{table}[htbp]
\centering
\caption{Comparison of Model Performance of Our Proposed KAN Models on BreastMNIST and PneumoniaMNIST Datasets}
\vspace{1em} 
\label{MedMNIST}
\fontsize{8}{10}\selectfont  
\renewcommand{\arraystretch}{0.8}

\resizebox{\textwidth}{!}{  
\begin{tabular}{|@{\hskip 2pt}c@{\hskip 2pt}|@{\hskip 2pt}c@{\hskip 2pt}|@{\hskip 2pt}c@{\hskip 2pt}|}
\hline
\multicolumn{3}{|c|}{\textbf{BreastMNIST Dataset}} \\
\hline
\textbf{Model Name} & \textbf{Test ACC (\%)} & \textbf{AUC (\%)} \\
\hline
SADAE \citep{ge2022self} & 85.90 & 89.70 \\
BP-CapsNet \citep{BPcapsnet2023} & 84.00 & 82.40 \\
MedViT-S \citep{manzari2023medvit} & 89.70 & 93.80 \\
EHDFL \citep{han2023ehdfl} & 89.70 & 89.40 \\
MedKAN-S \citep{yang2025medkan} & 90.40 & 94.20 \\
MedKAN-B \citep{yang2025medkan} & 89.50 & 93.70 \\
MedKAN-L \citep{yang2025medkan} & 88.50 & 85.90 \\
\hline
\textbf{SBTAYLOR-KAN} & \textbf{89.38} & \textbf{94.00} \\
\hline
SBRBF-KAN & 83.80 & 89.90 \\
SBWAVELET-KAN & 77.80 & 65.50 \\
\hline
\multicolumn{3}{|c|}{\textbf{PneumoniaMNIST Dataset}} \\
\hline
\textbf{Model Name} & \textbf{Test ACC (\%)} & \textbf{AUC (\%)} \\
\hline
AutoML \citep{bisong2019building} & 94.60 & 99.10 \\
SADAE \citep{ge2022self} & 90.10 & 98.30 \\
BP-CapsNet \citep{BPcapsnet2023} & 92.00 & 97.00 \\
MedViT-S \citep{manzari2023medvit} & 96.10 & 99.50 \\
MedKAN-S \citep{yang2025medkan} & 95.20 & 99.30 \\
MedKAN-B \citep{yang2025medkan} & 96.30 & 99.60 \\
MedKAN-L \citep{yang2025medkan} & 92.40 & 98.80 \\
\hline
\textbf{SBTAYLOR-KAN} & \textbf{95.39} & \textbf{99.89} \\
\hline
SBRBF-KAN & 95.06 & 98.00 \\
SBWAVELET-KAN & 91.45 & 94.50 \\
\hline
\end{tabular}
}
\end{table}

\subsection{Comparisons of Model Parameters (our Vs existing)}

Our network architecture builds upon the KAN  and leverages Kolmogorov's superposition theorem, which states that any multivariate continuous function can be represented as a finite composition of univariate continuous functions. Central to this design are KANLinear layers, which integrate basis functions—such as B-splines—through trainable weights. These weights are optimized during training to form structured compositions, enabling the network to approximate complex target functions with high fidelity. This formulation not only embodies the theoretical foundation of Kolmogorov’s principle but also aligns with the Universal Approximation Theorem~\citep{Hornik1989}, which asserts that neural networks with sufficient capacity can approximate any continuous function using trainable parameters alone. By relying exclusively on learnable components, the architecture avoids the need for non-trainable elements, maintaining both theoretical soundness and practical expressiveness in function approximation.

In this section, Table~\ref{tab:model_size_comparison} compares the size of the model parameters in large, medium, and small models, demonstrating a clear trade-off between complexity and model size. Among the large models, MedKAN-L exhibits the highest parameter count at 48~million, followed by MedKAN-B (24.6M) and traditional CNN architectures such as ResNet50 (24.19M), InceptionV3 (22.11M), and VGG16 (14.79M). MedKAN-S stands out as a more lightweight large model with 11.5~million parameters. The medium category contains only MobileNetV2, which is substantially smaller at 2.63~million parameters while maintaining a moderate network depth of 56 layers. In contrast, the proposed small models, including SBRBF-KAN, SBWAVELET-KAN, and SBTAYLOR-KAN, achieve ultra-compact configurations with only 93k, 3.3k, and 2.8k parameters, respectively, eliminating non-trainable weights. This highlights SBTAYLOR-KAN as the most parameter-efficient model, achieving strong performance despite its extremely small footprint, making it ideal for resource-constrained medical applications.

\begin{table*}[htbp]
\centering
\caption{Comparison of Model Parameter Sizes Across Experimented Medical Datasets, Highlighting the Superior Performance of SBTAYLOR-KAN in Small-Size Models. Unavailable values are marked with (--).}
\vspace{1em} 
\label{tab:model_size_comparison}
\resizebox{0.8\textwidth}{!}{  
\begin{tabular}{|@{\hskip 2pt}l@{\hskip 2pt}|@{\hskip 2pt}l@{\hskip 2pt}|@{\hskip 2pt}c@{\hskip 2pt}|@{\hskip 2pt}c@{\hskip 2pt}|@{\hskip 2pt}c@{\hskip 2pt}|@{\hskip 2pt}c@{\hskip 2pt}|}
\hline
\textbf{Size} & \textbf{Model} & \textbf{Layers} & \textbf{Trainable} & \textbf{Non-Trainable} & \textbf{Total} \\
\hline
\multirow{6}{*}{Large} 
 & MedKAN-L          & --  & --        & --         & 48{,}000{,}000 \\
 & MedKAN-B          & --  & --        & --         & 24{,}600{,}000 \\
 & ResNet50          & 158 & 602{,}118 & 23{,}587{,}712 & 24{,}189{,}830 \\
 & InceptionV3       & 166 & 307{,}206 & 21{,}802{,}784 & 22{,}109{,}990 \\
 & VGG16             & 20  & 75{,}267  & 14{,}714{,}688 & 14{,}789{,}955 \\
 & MedKAN-S          & --  & --        & --         & 11{,}500{,}000 \\
\hline
Medium 
 & MobileNetV2       & 56  & 376{,}326 & 2{,}257{,}984 & 2{,}634{,}310 \\
\hline
\multirow{3}{*}{Small} 
 & SBRBF-KAN         & 16  & 93{,}168  & 0           & 93{,}168 \\
 & SBWAVELET-KAN     & 10  & 3{,}300   & 0           & 3{,}300 \\
 & SBTAYLOR-KAN      & 10  & 2{,}872   & 0           & 2{,}872 \\
\hline
\end{tabular}
}  
\end{table*}

\subsection{Interpretability and Visual Validation Using Grad-CAM}
Grad-CAM \citep{GradCAM2017} is a widely used technique for enhancing the interpretability of deep neural networks, particularly in computer vision applications. It works by leveraging the gradients of a target class flowing into the final convolutional layer to generate coarse localization heatmaps. These heatmaps highlight the regions in an input image that most strongly influence the model’s prediction, effectively visualizing where the model’s attention is focused. It offers valuable insight into a model's decision-making by highlighting regions that influence predictions. This enhances interpretability, helps detect biases or failure modes, and improves model reliability. In high-stakes applications like medical imaging, such transparency is essential for building trust and validating models before deployment.

\begin{figure}[!t]
    \centering
    \includegraphics[width=0.9\textwidth]{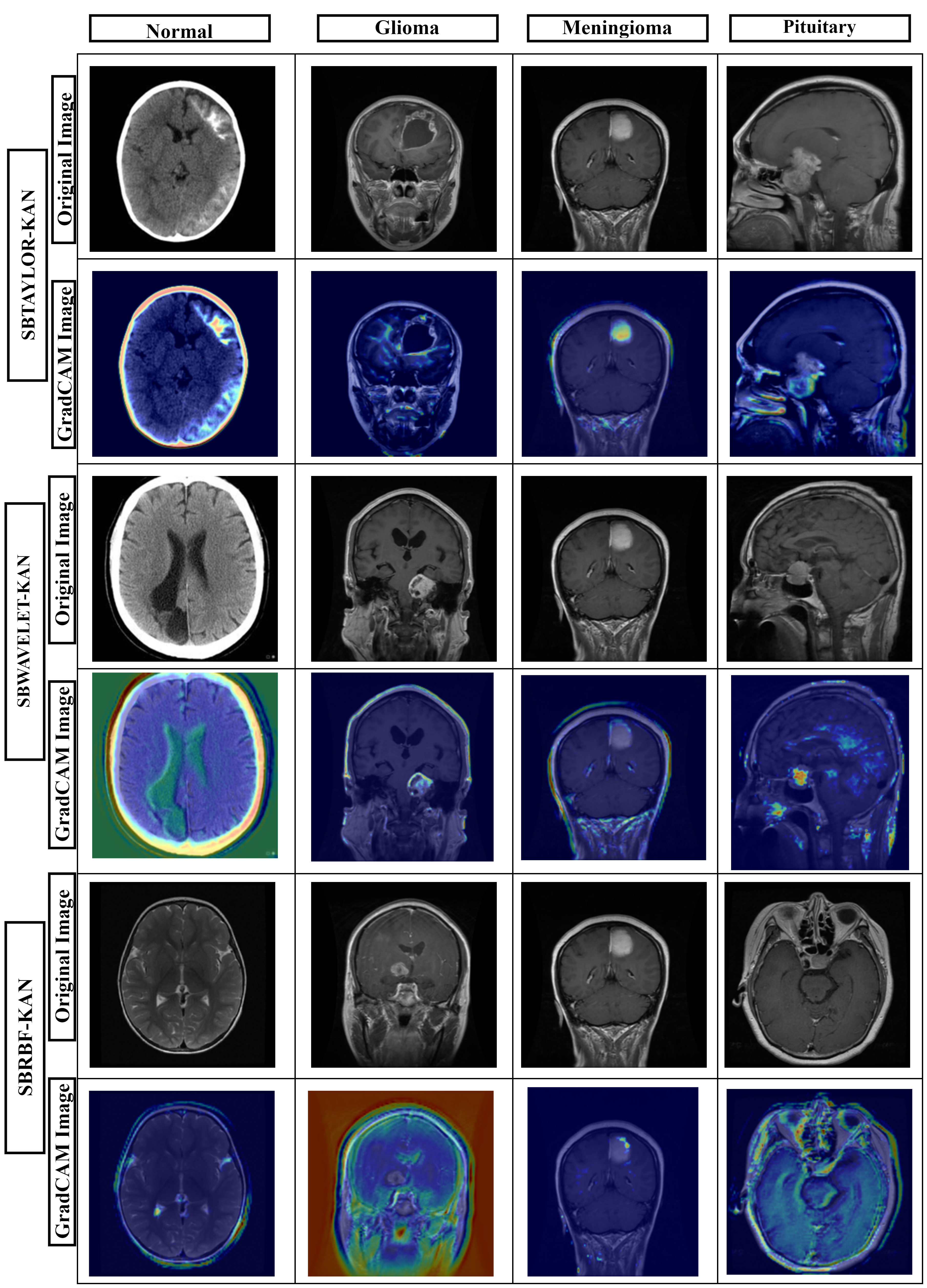}
    \caption{Grad-CAM outputs of three kernel-based models on distinct brain MRI images, highlighting attention differences in tumor localization.}
    \label{fig:gradcam_bt}
\end{figure}

\begin{figure}[!t]
    \centering
    \includegraphics[height=0.8\textheight]{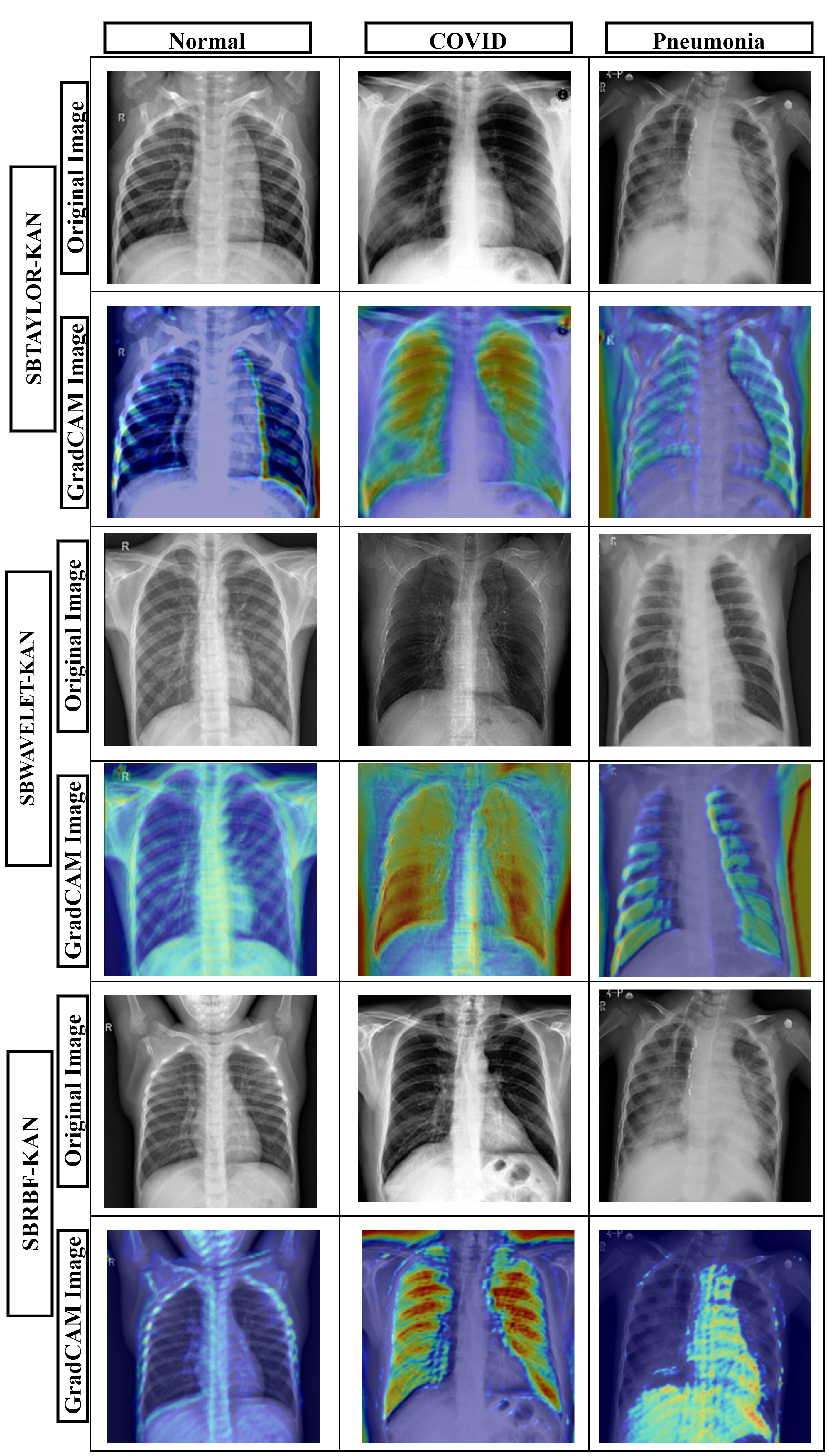}
    \caption{Grad-CAM outputs of three kernel-based models on distinct chest X-ray images, showing model-specific attention patterns for thoracic abnormalities.}
    \label{fig:gradcam_chestxray}
\end{figure}

\begin{figure}[!t]
    \centering
    \includegraphics[width=0.9\textwidth]{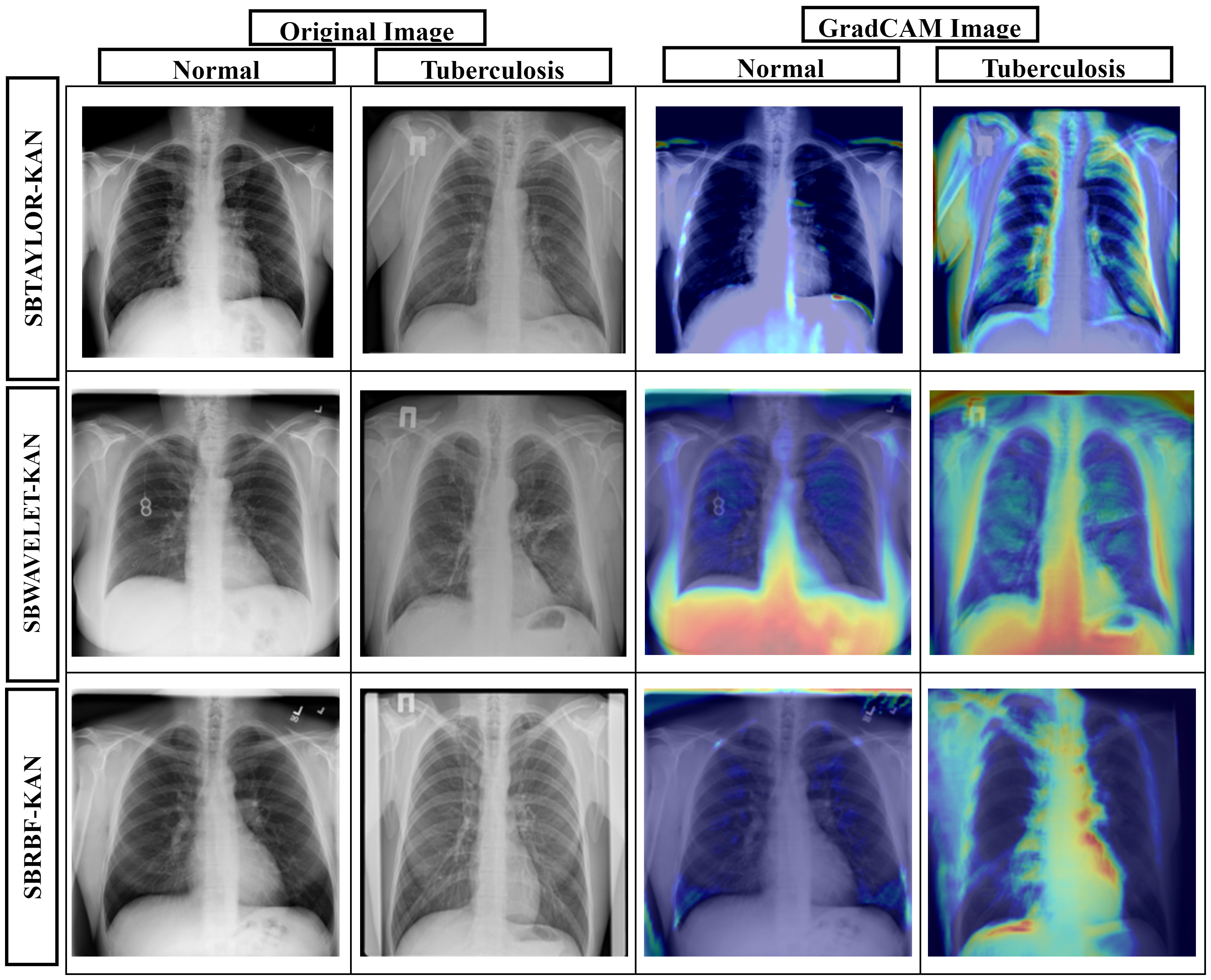}
    \caption{Grad-CAM outputs of three kernel-based models on normal and tuberculosis chest X-rays, illustrating attention shifts between pathological and healthy regions.}
    \label{fig:gradcam_tbxray}
\end{figure}

\begin{figure}[!t]
    \centering
    \includegraphics[width=0.9\textwidth]{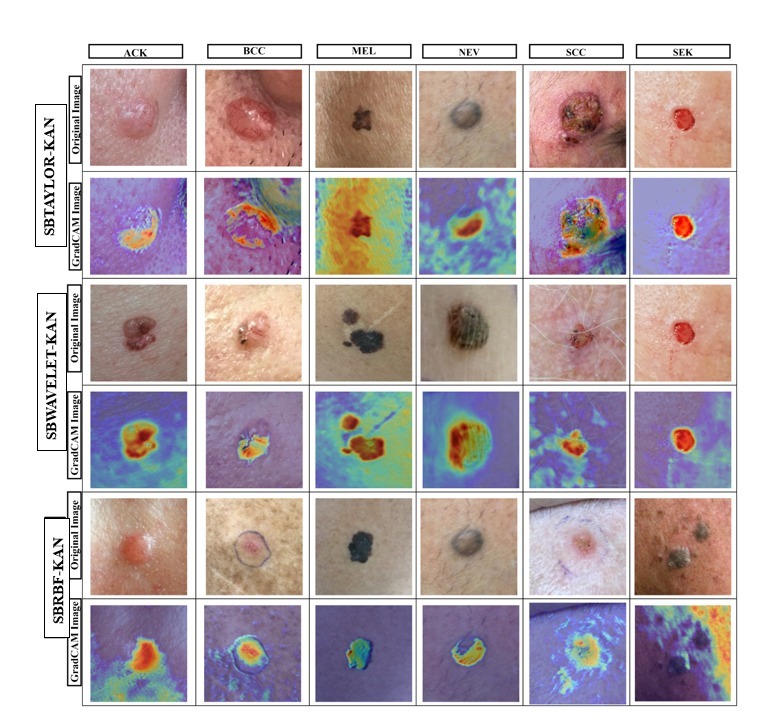}
    \caption{Grad-CAM outputs of three kernel-based models on distinct skin lesion images, showing attention patterns across lesion classes.}
    \label{fig:gradcam_skincancer}
\end{figure}

In this study, we applied Grad-CAM to the feature maps of our proposed KAN model to ensure that it consistently attends to clinically significant regions. This visual validation lets clinicians confirm that the model’s predictions are grounded in relevant medical features, thereby increasing trust in its outcomes. To evaluate and compare the interpretability of different KAN models, we selected separate input images from the same class across multiple datasets instead of relying on a single shared image. This choice better reflects real clinical scenarios, where medical images naturally vary due to differences in acquisition settings, lesion appearance, and dataset characteristics. This approach avoids biases from pixel-wise comparisons, revealing model-specific saliency patterns and enabling more robust, clinically meaningful evaluation of visual explanations where interpretability and generalization are essential. Figures~\ref{fig:gradcam_bt}–\ref{fig:gradcam_skincancer} show Grad-CAM visualizations from the KAN model, highlighting key regions that drive its predictions and demonstrating interpretability.

Across the different medical imaging datasets, each B-spline-based KAN model—TAYLOR-KAN, RBF-KAN, and WAVELET-KAN—exhibits its own characteristic performance. In the brain tumor dataset, WAVELET-KAN provides the clearest and most focused tumor highlights, RBF-KAN distributes its attention more broadly to include surrounding context. At the same time, TAYLOR-KAN demonstrates moderate yet reliable performance by capturing the tumors along with adjacent structures, which aids in understanding the overall brain context. In the chest X-ray datasets, including both COVID/Pneumonia and Tuberculosis cases, WAVELET-KAN consistently delivers highly localized and diagnostically precise attention on the diseased lung regions. RBF-KAN produces smoother and more diffused activation maps that cover larger thoracic areas, sometimes including irrelevant regions. In contrast, TAYLOR-KAN balances these behaviors by highlighting the key pathological regions while incorporating part of the surrounding tissue. This makes it practical for diagnostic scenarios where contextual information is beneficial. In the skin lesion dataset, WAVELET-KAN sharply isolates the lesions, RBF-KAN offers broad but less lesion-specific coverage. TAYLOR-KAN again exhibits a stable middle ground by identifying the lesion while including nearby skin patterns. 
Overall, WAVELET-KAN provides high pinpoint accuracy, while RBF-KAN offers broader contextual coverage. TAYLOR-KAN emerges as a reliable all-rounder, balancing precision and context by capturing the primary lesion along with relevant surrounding features. This makes it especially suitable for real-world diagnostic tasks, where both accuracy and contextual awareness are essential.

\section{Discussion}
\label{discussion}
The primary objective of this study was to investigate and identify the most effective spline-based KAN configuration for medical image classification tasks involving small-scale, heterogeneous, and rare disease datasets---a scenario often encountered in real-world clinical settings. To address this, we introduced and systematically evaluated three lightweight KAN variants: SBTAYLOR-KAN, SBRBF-KAN, and SBWAVELET-KAN. Each variant employs a distinct mathematical foundation for functional approximation named Taylor series, RBF, and wavelet transforms, respectively, while leveraging a shared B-spline structure to promote compact and interpretable model architectures. A central goal of this investigation was to identify the configuration that best balances accuracy, generalization, and computational efficiency, particularly in constrained or data-scarce environments.

Extensive experiments were conducted across four diverse medical imaging datasets, including Brain Tumor MRI, COVID-19 Chest X-rays, TB Chest X-rays, and PAD-UFES-20 Skin Cancer to assess the models' classification performance under real-world variability. Comparing Table \ref{tab:kan_performance} and Table \ref{tab:performance_metrics}, the results demonstrate that SBTAYLOR-KAN consistently outperformed both the other KAN variants and a range of deep learning baselines, including ResNet50 (24.19M parameters), InceptionV3 (22.10M), VGG16 (14.78M), and MobileNetV2 (2.63M), all of which are significantly larger in scale. For instance, on the Brain Tumor, COVID-19 chest X-ray, and TB-Xray datasets, SBTAYLOR-KAN achieved an accuracy of 95.09\%, 96.37\%, and 98.93\%  using only 2,872 trainable parameters, highlighting a dramatic improvement in parameter efficiency without compromising predictive performance. Likewise, despite the significant class imbalance in the skin cancer dataset, we created a balanced version of the data and performed experiments using both the imbalanced and balanced datasets. In these experiments, SBTAYLOR-KAN outperformed the other models, achieving an accuracy of 68.22\% when evaluated on the balanced dataset. These findings further highlight the model's robustness and efficiency in handling various medical imaging challenges.

In terms of generalization, SBTAYLOR-KAN also demonstrated superior robustness to unseen data. As shown in Table~\ref{tab:model_performance}, it maintained high accuracy and F1-scores on external test sets never seen during training, with only negligible performance drops. This confirms its capacity to retain reliable diagnostic accuracy across domain shifts, a critical need for medical AI applications. 

Supporting this further, statistical analysis across datasets (Table \ref{tab:kc_analysis}) revealed that TAYLOR-KAN consistently demonstrated reliable performance across all four medical imaging datasets. In Dataset-1 (Brain Tumor), it achieved strong agreement scores (KC = 0.8335, MCC = 0.8354), despite not being the top performer. In Dataset-2 (COVID-19), TAYLOR-KAN outperformed all other models with the highest KC (0.9570), MCC (0.9571), and the lowest error rate (0.0285), underscoring its robustness in critical diagnostic tasks. While RBF-KAN slightly led in Dataset-3 (TB X-ray) with a KC of 0.9830, TAYLOR-KAN maintained a competitive performance (KC = 0.9773, MCC = 0.9774, Error Rate = 0.0113), showcasing minimal performance drop.

In the balanced PAD-UFES-20 dataset, the SBTAYLOR-KAN model also achieved the best performance among the three models evaluated. It achieved a KC score of 0.5772, an MCC of 0.5795, and the lowest error rate of 0.3523, with all results being statistically significant (p-value $<$ 0.05). This shows that the Taylor-KAN model consistently outperforms the others in accuracy and reliability for this dataset.

To further assess performance under limited data scenarios, we conducted experiments using progressive data reduction. Since the performance of the KAN models was improved with the balanced PAD-UFES-20 dataset (see \ref{tab:kan_performance}), this dataset was selected for further experimentation.
The results indicate that SBTAYLOR-KAN retained $>$88\% accuracy at 50\% data and remained above 86\% accuracy with only 30\% of training data on the Brain Tumor dataset. On the COVID-19 dataset, it achieved the highest F1-score of 97.33\% and maintained accuracy above 92\% even under severe data scarcity. Additionally, on the TB X-ray dataset, SBTAYLOR-KAN consistently maintained accuracy above 96\% down to 25\% data and preserved a strong F1-score of 92.50\% even at 20\% data. Compared to the other KAN variants, SBTAYLOR-KAN also exhibited smoother degradation curves and less performance fluctuation, particularly on balanced PAD-UFES-20, where its accuracy and F1-score remained stable. 

Additionally, experiments on BreastMNIST and PneumoniaMNIST, both small-scale subsets of the MedMNIST benchmark, reaffirmed the scalability and practical deployability of our models. As seen in Table V, SBTAYLOR-KAN achieved 89.38\% accuracy (AUC = 95.90\%) on BreastMNIST and 95.39\% accuracy (AUC = 98.89\%) on PneumoniaMNIST, outperforming even larger KAN-based architectures like MedKAN-L (48M parameters) and MedKAN-B (24.6M parameters), which achieved 88.50\% and 94.60\%, respectively. Importantly, SBTAYLOR-KAN accomplished this with only 2,872 parameters, demonstrating more than 99.99\% parameter reduction compared to the largest models, without compromising accuracy. Both SBRBF-KAN (93,168 params) and SBWAVELET-KAN (3,300 params) also produced competitive results, further reinforcing the strength of spline-based compact architectures, but neither matched the consistent dominance of SBTAYLOR-KAN.

Overall, the findings of this study empirically validate the superiority of the SBTAYLOR-KAN model, both in terms of predictive performance and practical applicability. As hypothesized, the Taylor series-based KAN architecture delivers the most robust and consistent performance among the evaluated configurations. Its mathematical structure allows for effective local approximation, while the B-spline foundation ensures smooth global learning dynamics, making it particularly well-suited for tasks involving small or highly imbalanced datasets. Although the dataset is small, it is well-balanced, allowing the model to perform effectively. Moreover, the SBTAYLOR-KAN exhibits strong performance without relying on image preprocessing, underscoring its architectural efficiency. Its lightweight design and minimal computational demands render it well-suited for deployment in resource-limited settings, including rural clinics, mobile units, and edge devices. This positions the model as a practical, interpretable, and scalable solution for real-world medical AI applications.

\section{Limitations and Future Work}
\label{limitations_future_work}
The SBTAYLOR-KAN model demonstrates strong potential in terms of efficiency, accuracy, and interpretability, but it also presents several limitations that open up avenues for future research and development. One of the key findings from our experiments with the PAD-UFES-20 dataset is that balancing the data significantly improves model performance. This suggests that the proposed KAN architectures effectively learn from smaller, balanced datasets. However, a practical limitation arises: real-world medical datasets are often inherently imbalanced, and achieving balance is not always feasible. While balancing enhances classification accuracy, future research should explore strategies like advanced data augmentation, cost-sensitive learning, or generative methods to ensure the model remains robust under imbalanced conditions. Besides, integrating Taylor series and B-spline functions enables smooth modelling but may struggle with abrupt discontinuities or irregular patterns in noisy data. While the model’s compact and interpretable design is suited for resource-limited settings, it may not capture the deeper hierarchical representations needed for complex tasks.
To address these challenges, future work could incorporate advanced techniques for handling class imbalance and exploring hybrid KAN architectures. Specifically, combining KAN with lightweight transformer modules could enable the model to extract richer representations from limited data, especially in image classification tasks, without significantly increasing the number of parameters. Additionally, expanding the theoretical framework to encompass broader function spaces could further extend the model’s applicability across various scientific and biomedical domains. Overall, these directions aim to enhance the flexibility, robustness, and scalability of the SBTAYLOR-KAN model, while maintaining its core strengths in efficiency and interpretability.

\section*{CRediT authorship contribution statement}
\noindent
\textbf{Kaniz Fatema:} Conceptualization, Data curation, Resources, Methodology, Software, Writing – original draft, Writing – review and editing \\
\textbf{Emad A. Mohammed:} Funding acquisition, Formal analysis, Investigation, Project administration, Validation, Writing – review and editing
 \\
\textbf{Sukhjit Singh Sehra:} Visualization, Validation, Writing – review and editing

\section*{Declaration of competing interest}
The authors declare that they have no financial interests or personal relationships that could have influenced the work presented in this paper.

\section*{Data Availability}
The datasets used in this study are publicly available for research purposes. The Brain Tumor Kaggle Dataset \citep{Nickparvar-BTMD}, COVID-19 Chest X-Ray Dataset \citep{SK-COVIDX}, TB Chest X-ray Database \citep{TB-Kaggle}, PAD-UFES-20 Skin Cancer Dataset \citep{MendeleyDataset}, BreastMNIST Dataset \citep{MedMNIST}, PneumoniaMNIST Dataset \citep{MedMNIST}, COVID-19 Radiography Dataset \citep{RahmanCOVID19}, Shenzhen \& Montgomery CXR Dataset \citep{TB-CXR-Shenzhen, MC-CXR}, and Fluorescence Skin Cancer (FLUO-SC) Dataset \citep{FLUO-SC2024}.

\bibliography{References}    

\appendix
\label{appendix}
\setcounter{table}{0}  
\renewcommand{\thetable}{A.\arabic{table}}  


\begin{table*}[htbp]
\centering
\caption{Compare the Experimental Results of Each KAN Model on Brain Tumor Dataset with Varying Data Reductions}
\vspace{1em} 
\label{tab:brain_tumor_results}
\fontsize{8}{10}\selectfont  

\renewcommand{\arraystretch}{0.8}  

\resizebox{\textwidth}{!}{  
\begin{tabular}{|c|c|c|c|c|c|c|}
\hline
\multirow{2}{*}{\textbf{Percentage of Data}} & \multirow{2}{*}{\textbf{Model}} & \multicolumn{4}{c|}{\textbf{Performance Metrics}} \\ \cline{3-6} 
 & & \textbf{ACC (\%)} & \textbf{Precision (\%)} & \textbf{Recall (\%)} & \textbf{F1-score (\%)} \\ 
\hline
\multirow{3}{*}{100\% (3,998)} & SBTAYLOR-KAN & 95.09 & 93.75 & 93.75 & 93.75 \\
 & SBRBF-KAN & 93.68 & 93.00 & 93.00 & 93.00 \\
 & SBWAVELET-KAN & 94.91 & 94.75 & 94.50 & 94.62 \\
\hline
\multirow{3}{*}{95\% (3,798)} & SBTAYLOR-KAN & 93.51 & 93.00 & 93.00 & 93.00 \\
 & SBRBF-KAN & 91.67 & 91.50 & 91.00 & 91.25 \\
 & SBWAVELET-KAN & 94.56 & 93.50 & 93.25 & 93.37 \\
\hline
\multirow{3}{*}{90\% (3,598)} & SBTAYLOR-KAN & 93.77 & 92.25 & 92.25 & 92.25 \\
 & SBRBF-KAN & 92.37 & 91.00 & 90.50 & 90.75 \\
 & SBWAVELET-KAN & 94.30 & 93.75 & 93.50 & 93.62 \\
\hline
\multirow{3}{*}{85\% (3,398)} & SBTAYLOR-KAN & 94.74 & 92.00 & 92.00 & 92.00 \\
 & SBRBF-KAN & 91.84 & 90.00 & 90.00 & 90.00 \\
 & SBWAVELET-KAN & 87.54 & 86.00 & 85.00 & 85.50 \\
\hline
\multirow{3}{*}{80\% (3,198)} & SBTAYLOR-KAN & 93.25 & 91.50 & 91.50 & 91.50 \\
 & SBRBF-KAN & 91.67 & 92.25 & 91.75 & 92.00 \\
 & SBWAVELET-KAN & 93.33 & 91.00 & 91.00 & 91.00 \\
\hline
\multirow{3}{*}{75\% (2,998)} & SBTAYLOR-KAN & 91.93 & 91.00 & 90.25 & 90.62 \\
 & SBRBF-KAN & 90.61 & 89.25 & 89.25 & 89.25 \\
 & SBWAVELET-KAN & 89.82 & 90.25 & 89.75 & 90.00 \\
\hline
\multirow{3}{*}{70\% (2,798)} & SBTAYLOR-KAN & 92.98 & 92.00 & 91.50 & 91.75 \\
 & SBRBF-KAN & 90.44 & 91.00 & 90.25 & 90.62 \\
 & SBWAVELET-KAN & 93.07 & 92.00 & 92.00 & 92.00 \\
\hline
\multirow{3}{*}{65\% (2,598)} & SBTAYLOR-KAN & 90.53 & 89.25 & 89.00 & 89.12 \\
 & SBRBF-KAN & 89.39 & 89.00 & 89.00 & 89.00 \\
 & SBWAVELET-KAN & 72.28 & 76.50 & 70.50 & 73.37 \\
\hline
\multirow{3}{*}{60\% (2,398)} & SBTAYLOR-KAN & 89.91 & 88.75 & 87.50 & 88.12 \\
 & SBRBF-KAN & 89.39 & 88.25 & 88.00 & 88.12 \\
 & SBWAVELET-KAN & 92.19 & 92.00 & 92.00 & 92.00 \\
\hline
\multirow{3}{*}{55\% (2,198)} & SBTAYLOR-KAN & 91.84 & 89.50 & 89.50 & 89.50 \\
 & SBRBFLET-KAN & 90.26 & 89.75 & 89.75 & 89.75 \\
 & SBWAVE-KAN & 90.00 & 88.50 & 87.75 & 88.12 \\
\hline
\multirow{3}{*}{50\% (1,999)} & SBTAYLOR-KAN & 89.56 & 87.50 & 87.50 & 87.50 \\
 & SBRBF-KAN & 89.65 & 87.50 & 87.50 & 87.50 \\
 & SBWAVELET-KAN & 90.18 & 90.00 & 90.00 & 90.00 \\
\hline
\multirow{3}{*}{45\% (1,799)} & SBTAYLOR-KAN & 89.74 & 89.50 & 89.00 & 89.25 \\
 & SBRBF-KAN & 89.21 & 87.00 & 87.00 & 87.00 \\
 & SBWAVELET-KAN & 89.82 & 88.75 & 88.25 & 88.50 \\
\hline
\multirow{3}{*}{40\% (1,599)} & SBTAYLOR-KAN & 87.28 & 88.25 & 88.00 & 88.12 \\
 & SBRBF-KAN & 87.63 & 84.50 & 84.50 & 84.50 \\
 & SBWAVELET-KAN & 89.39 & 87.00 & 86.00 & 86.50 \\
\hline
\multirow{3}{*}{35\% (1,399)} & SBTAYLOR-KAN & 89.04 & 87.00 & 87.00 & 87.00 \\
 & SBRBF-KAN & 88.68 & 87.00 & 87.00 & 87.00 \\
 & SBWAVELET-KAN & 89.12 & 90.00 & 90.00 & 90.00 \\
\hline
\multirow{3}{*}{30\% (1,199)} & SBTAYLOR-KAN & 86.49 & 86.00 & 86.00 & 86.00 \\
 & SBRBF-KAN & 87.02 & 83.25 & 83.25 & 83.25 \\
 & SBWAVELET-KAN & 87.98 & 88.00 & 88.00 & 88.00 \\
\hline
\multirow{3}{*}{25\% (999)} & SBTAYLOR-KAN & 80.88 & 86.75 & 80.50 & 83.50 \\
 & SBRBF-KAN & 85.79 & 82.75 & 82.50 & 82.62 \\
 & SBWAVELET-KAN & 76.58 & 80.50 & 80.50 & 80.50 \\
\hline
\multirow{3}{*}{20\% (799)} & SBTAYLOR-KAN & 87.02 & 85.00 & 84.50 & 84.75 \\
 & SBRBF-KAN & 83.60 & 82.00 & 82.00 & 82.00 \\
 & SBWAVELET-KAN & 84.56 & 80.00 & 80.00 & 80.00 \\
\hline
\end{tabular}
}
\end{table*}


\begin{table*}[htbp]
\centering
\caption{Compare the Experimental Results of Each KAN Model on the Chest X-Ray Dataset with Varying Data Reductions}
\vspace{1em} 
\label{tab:COVID_19_Xray}
\fontsize{8}{10}\selectfont  

\renewcommand{\arraystretch}{0.8}

\resizebox{\textwidth}{!}{  
\begin{tabular}{|c|c|c|c|c|c|c|}
\hline
\multirow{2}{*}{\textbf{Percentage of Data}} & \multirow{2}{*}{\textbf{Model}} & \multicolumn{4}{c|}{\textbf{Performance Metrics}} \\ \cline{3-6} 
 & & \textbf{ACC (\%)} & \textbf{Precision (\%)} & \textbf{Recall (\%)} & \textbf{F1-score (\%)} \\ 
\hline
\multirow{3}{*}{100\% (3,660)} & SBTAYLOR-KAN & 96.37 & 97.00 & 97.00 & 97.33 \\
 & SBRBF-KAN & 95.51 & 96.00 & 96.00 & 96.00 \\
 & SBWAVELET-KAN & 95.41 & 94.67 & 94.67 & 94.67 \\
\hline
\multirow{3}{*}{95\% (3,477)} & SBTAYLOR-KAN & 95.80 & 95.00 & 95.00 & 95.00 \\
 & SBRBF-KAN & 95.99 & 96.33 & 96.33 & 96.33 \\
 & SBWAVELET-KAN & 95.41 & 95.33 & 95.67 & 95.67 \\
\hline
\multirow{3}{*}{90\% (3,294)} & SBTAYLOR-KAN & 95.61 & 96.33 & 96.00 & 95.66 \\
 & SBRBF-KAN & 95.13 & 95.33 & 95.00 & 95.33 \\
 & SBWAVELET-KAN & 90.83 & 89.67 & 89.33 & 89.33 \\
\hline
\multirow{3}{*}{85\% (3,111)} & SBTAYLOR-KAN & 94.94 & 95.66 & 95.66 & 95.33 \\
 & SBRBF-KAN & 95.42 & 95.67 & 95.33 & 95.33 \\
 & SBWAVELET-KAN & 95.51 & 95.00 & 95.00 & 95.00 \\
\hline
\multirow{3}{*}{80\% (2,928)} & SBTAYLOR-KAN & 95.42 & 95.33 & 95.33 & 95.66 \\
 & SBRBF-KAN & 95.32 & 95.67 & 95.33 & 95.33 \\
 & SBWAVELET-KAN & 94.65 & 96.00 & 96.00 & 96.00 \\
\hline
\multirow{3}{*}{75\% (2,745)} & SBTAYLOR-KAN & 95.32 & 96.00 & 96.00 & 96.00 \\
 & SBRBF-KAN & 95.61 & 95.00 & 95.00 & 94.67 \\
 & SBWAVELET-KAN & 96.08 & 94.67 & 94.67 & 95.00 \\
\hline
\multirow{3}{*}{70\% (2,562)} & SBTAYLOR-KAN & 94.08 & 94.33 & 94.66 & 94.33 \\
 & SBRBF-KAN & 94.94 & 95.67 & 96.00 & 95.67 \\
 & SBWAVELET-KAN & 95.60 & 94.33 & 94.67 & 94.33 \\
\hline
\multirow{3}{*}{65\% (2,379)} & SBTAYLOR-KAN & 95.13 & 96.00 & 96.00 & 95.66 \\
 & SBRBF-KAN & 94.75 & 95.33 & 95.33 & 95.00 \\
 & SBWAVELET-KAN & 94.55 & 93.67 & 94.00 & 94.00 \\
\hline
\multirow{3}{*}{60\% (2,196)} & SBTAYLOR-KAN & 93.79 & 94.33 & 94.33 & 94.33 \\
 & SBRBF-KAN & 95.32 & 95.33 & 95.00 & 95.00 \\
 & SBWAVELET-KAN & 93.41 & 93.67 & 93.33 & 93.67 \\
\hline
\multirow{3}{*}{55\% (2,013)} & SBTAYLOR-KAN & 95.03 & 94.33 & 94.66 & 94.33 \\
 & SBRBF-KAN & 95.03 & 94.33 & 94.00 & 94.00 \\
 & SBWAVELET-KAN & 94.46 & 94.67 & 95.00 & 94.67 \\
\hline
\multirow{3}{*}{50\% (1,830)} & SBTAYLOR-KAN & 93.79 & 94.00 & 93.67 & 93.67 \\
 & SBRBF-KAN & 94.75 & 94.00 & 93.67 & 94.00 \\
 & SBWAVELET-KAN & 95.03 & 95.00 & 95.00 & 94.67 \\
\hline
\multirow{3}{*}{45\% (1,646)} & SBTAYLOR-KAN & 95.99 & 95.66 & 95.00 & 95.33 \\
 & SBRBF-KAN & 93.89 & 93.33 & 93.33 & 93.33 \\
 & SBWAVELET-KAN & 94.17 & 94.00 & 94.00 & 94.00 \\
\hline
\multirow{3}{*}{40\% (1,464)} & SBTAYLOR-KAN & 93.79 & 94.67 & 94.33 & 94.33 \\
 & SBRBF-KAN & 94.36 & 94.67 & 94.33 & 94.33 \\
 & SBWAVELET-KAN & 93.69 & 94.33 & 94.33 & 94.33 \\
\hline
\multirow{3}{*}{35\% (1,281)} & SBTAYLOR-KAN & 92.93 & 93.33 & 93.33 & 93.33 \\
 & SBRBF-KAN & 93.31 & 95.33 & 95.00 & 95.00 \\
 & SBWAVELET-KAN & 91.59 & 92.00 & 92.33 & 92.00 \\
\hline
\multirow{3}{*}{30\% (1,098)} & SBTAYLOR-KAN & 92.74 & 93.33 & 93.33 & 93.33 \\
 & SBRBF-KAN & 93.89 & 94.67 & 94.67 & 94.67 \\
 & SBWAVELET-KAN & 94.46 & 93.67 & 93.67 & 93.33 \\
\hline
\multirow{3}{*}{25\% (915)} & SBTAYLOR-KAN & 93.60 & 93.67 & 93.67 & 93.33 \\
 & SBRBF-KAN & 92.17 & 92.00 & 92.00 & 92.00 \\
 & SBWAVELET-KAN & 92.26 & 93.67 & 93.33 & 93.67 \\
\hline
\multirow{3}{*}{20\% (731)} & SBTAYLOR-KAN & 92.45 & 93.00 & 93.00 & 93.33 \\
 & SBRBF-KAN & 93.12 & 90.33 & 90.00 & 90.33 \\
 & SBWAVELET-KAN & 90.74 & 90.67 & 90.67 & 91.00 \\
\hline
\end{tabular}
}
\end{table*}

\begin{table*}[htbp]
\centering
\caption{Compare the Experimental Results of Each KAN Model on TB X-ray Dataset with Varying Data Reductions}
\vspace{1em} 
\label{tab:TB_Xray_results}
\fontsize{8}{10}\selectfont  

\renewcommand{\arraystretch}{0.8}  

\resizebox{\textwidth}{!}{  
\begin{tabular}{|c|c|c|c|c|c|c|}
\hline
\multirow{2}{*}{\textbf{Percentage of Data}} & \multirow{2}{*}{\textbf{Model}} & \multicolumn{4}{c|}{\textbf{Performance Metrics}} \\ \cline{3-6} 
 & & \textbf{ACC (\%)} & \textbf{Precision (\%)} & \textbf{Recall (\%)} & \textbf{F1-score (\%)} \\ 
\hline
\multirow{3}{*}{100\% (2,940)} & SBTAYLOR-KAN & 98.93 & 98.00 & 95.00 & 96.50 \\
 & SBRBF-KAN & 98.69 & 99.00 & 96.00 & 97.50 \\
 & SBWAVELET-KAN & 98.45 & 99.50 & 96.50 & 97.50 \\
\hline
\multirow{3}{*}{95\% (2,793)} & SBTAYLOR-KAN & 98.45 & 98.00 & 94.00 & 96.00 \\
 & SBRBF-KAN & 98.69 & 99.00 & 96.00 & 97.50 \\
 & SBWAVELET-KAN & 97.62 & 96.00 & 95.50 & 96.00 \\
\hline
\multirow{3}{*}{90\% (2,646)} & SBTAYLOR-KAN & 98.57 & 99.00 & 94.00 & 96.00 \\
 & SBRBF-KAN & 98.69 & 98.00 & 94.50 & 96.00 \\
 & SBWAVELET-KAN & 98.69 & 99.00 & 95.00 & 97.00 \\
\hline
\multirow{3}{*}{85\% (2,499)} & SBTAYLOR-KAN & 98.69 & 99.00 & 94.00 & 96.00 \\
 & SBRBF-KAN & 98.69 & 98.00 & 95.00 & 96.50 \\
 & SBWAVELET-KAN & 90.24 & 84.50 & 76.50 & 79.50 \\
\hline
\multirow{3}{*}{80\% (2,352)} & SBTAYLOR-KAN & 98.45 & 99.00 & 93.00 & 95.00 \\
 & SBRBF-KAN & 98.33 & 99.00 & 95.00 & 97.00 \\
 & SBWAVELET-KAN & 98.57 & 98.00 & 95.00 & 96.50 \\
\hline
\multirow{3}{*}{75\% (2,205)} & SBTAYLOR-KAN & 97.38 & 98.00 & 91.00 & 94.00 \\
 & SBRBF-KAN & 98.45 & 99.00 & 95.00 & 97.00 \\
 & SBWAVELET-KAN & 98.57 & 98.00 & 95.00 & 97.00 \\
\hline
\multirow{3}{*}{70\% (2,058)} & SBTAYLOR-KAN & 98.33 & 99.00 & 95.00 & 97.00 \\
 & SBRBF-KAN & 96.19 & 98.00 & 88.50 & 92.50 \\
 & SBWAVELET-KAN & 97.74 & 99.00 & 94.50 & 96.50 \\
\hline
\multirow{3}{*}{65\% (1,911)} & SBTAYLOR-KAN & 98.57 & 97.00 & 96.00 & 96.50 \\
 & SBRBF-KAN & 98.10 & 98.50 & 93.50 & 96.00 \\
 & SBWAVELET-KAN & 97.62 & 99.00 & 95.00 & 97.00 \\
\hline
\multirow{3}{*}{60\% (1,764)} & SBTAYLOR-KAN & 98.69 & 98.50 & 93.50 & 96.00 \\
 & SBRBF-KAN & 98.93 & 98.50 & 93.50 & 96.00 \\
 & SBWAVELET-KAN & 95.71 & 94.00 & 92.00 & 93.00 \\
\hline
\multirow{3}{*}{55\% (1,617)} & SBTAYLOR-KAN & 97.98 & 95.00 & 95.00 & 95.00 \\
 & SBRBF-KAN & 98.57 & 98.50 & 92.50 & 95.00 \\
 & SBWAVELET-KAN & 97.74 & 96.50 & 94.00 & 95.00 \\
\hline
\multirow{3}{*}{50\% (1,470)} & SBTAYLOR-KAN & 98.93 & 98.00 & 95.00 & 96.50 \\
 & SBRBF-KAN & 98.57 & 98.00 & 95.00 & 96.50 \\
 & SBWAVELET-KAN & 97.62 & 97.50 & 94.00 & 96.00 \\
\hline
\multirow{3}{*}{45\% (1,323)} & SBTAYLOR-KAN & 97.86 & 97.50 & 94.00 & 96.00 \\
 & SBRBF-KAN & 97.86 & 97.50 & 93.50 & 96.00 \\
 & SBWAVELET-KAN & 95.83 & 94.50 & 94.00 & 94.50 \\
\hline
\multirow{3}{*}{40\% (1,176)} & SBTAYLOR-KAN & 98.45 & 97.00 & 96.00 & 96.50 \\
 & SBRBF-KAN & 98.33 & 98.00 & 94.50 & 96.00 \\
 & SBWAVELET-KAN & 97.38 & 98.00 & 96.00 & 97.00 \\
\hline
\multirow{3}{*}{35\% (1,029)} & SBTAYLOR-KAN & 98.45 & 96.00 & 94.50 & 95.00 \\
 & SBRBF-KAN & 97.50 & 96.00 & 93.00 & 94.50 \\
 & SBWAVELET-KAN & 97.26 & 97.00 & 91.00 & 93.50 \\
\hline
\multirow{3}{*}{30\% (882)} & SBTAYLOR-KAN & 96.31 & 95.00 & 90.00 & 92.00 \\
 & SBRBF-KAN & 97.14 & 96.00 & 93.00 & 94.50 \\
 & SBWAVELET-KAN & 96.31 & 95.50 & 95.50 & 95.00 \\
\hline
\multirow{3}{*}{25\% (735)} & SBTAYLOR-KAN & 96.43 & 96.00 & 92.50 & 94.00 \\
 & SBRBF-KAN & 97.74 & 96.00 & 92.50 & 94.00 \\
 & SBWAVELET-KAN & 96.07 & 97.00 & 92.50 & 94.50 \\
\hline
\multirow{3}{*}{20\% (588)} & SBTAYLOR-KAN & 96.19 & 96.50 & 89.00 & 92.50 \\
 & SBRBF-KAN & 98.10 & 97.50 & 94.00 & 96.00 \\
 & SBWAVELET-KAN & 94.17 & 92.00 & 89.50 & 91.00 \\
\hline
\end{tabular}
}
\end{table*}

\begin{table*}[htbp]
\centering
\caption{Compare the Experimental Results of Each KAN Model on Balanced Skin Cancer (PAD-UFES-20) Dataset with Varying Data Reductions}
\vspace{1em} 
\label{tab:PAD-UFES-20_results}
\fontsize{8}{10}\selectfont  

\renewcommand{\arraystretch}{0.8}  

\resizebox{\textwidth}{!}{  
\begin{tabular}{|c|c|c|c|c|c|c|}
\hline
\multirow{2}{*}{\textbf{Percentage of Data}} & \multirow{2}{*}{\textbf{Model}} & \multicolumn{4}{c|}{\textbf{Performance Metrics}} \\ \cline{3-6} 
 & & \textbf{ACC (\%)} & \textbf{Precision (\%)} & \textbf{Recall (\%)} & \textbf{F1-score (\%)} \\ 
\hline
\multirow{3}{*}{100\% (2,100)} & SBTAYLOR-KAN & 68.22 & 73.29 & 72.77 & 72.51 \\
 & SBRBF-KAN & 67.72 & 64.84 & 65.25 & 64.64 \\
 & SBWAVELET-KAN & 60.52 & 58.50 & 52.60 & 55.39 \\
\hline
\multirow{3}{*}{95\% (1,995)} & SBTAYLOR-KAN & 63.56 & 64.53 & 63.92 & 63.44 \\
 & SBRBF-KAN & 64.39 & 65.50 & 65.18 & 65.04 \\
 & SBWAVELET-KAN & 58.68 & 55.00 & 55.00 & 55.00 \\
\hline
\multirow{3}{*}{90\% (1,890)} & SBTAYLOR-KAN & 64.06 & 67.87 & 67.82 & 67.58 \\
 & SBRBF-KAN & 58.90 & 61.55 & 60.86 & 61.12 \\
 & SBWAVELET-KAN & 59.80 & 58.45 & 58.54 & 50.00 \\
\hline
\multirow{3}{*}{85\% (1,785)} & SBTAYLOR-KAN & 62.06 & 65.47 & 63.88 & 64.38 \\
 & SBRBF-KAN & 66.56 & 66.44 & 66.21 & 66.20 \\
 & SBWAVELET-KAN & 57.99 & 40.50 & 41.83 & 40.50 \\
\hline
\multirow{3}{*}{80\% (1,680)} & SBTAYLOR-KAN & 62.40 & 62.73 & 63.18 & 62.39 \\
 & SBRBF-KAN & 63.56 & 63.27 & 62.92 & 62.54 \\
 & SBWAVELET-KAN & 57.11 & 38.00 & 38.16 & 37.00 \\
\hline
\multirow{3}{*}{75\% (1,575)} & SBTAYLOR-KAN & 61.56 & 61.34 & 61.61 & 60.91 \\
 & SBRBF-KAN & 60.73 & 57.85 & 57.97 & 57.11 \\
 & SBWAVELET-KAN & 57.11 & 38.33 & 40.33 & 38.83 \\
\hline
\multirow{3}{*}{70\% (1,470)} & SBTAYLOR-KAN & 57.40 & 60.20 & 58.30 & 57.42 \\
 & SBRBF-KAN & 60.73 & 61.16 & 60.98 & 59.97 \\
 & SBWAVELET-KAN & 58.86 & 57.16 & 41.00 & 42.66 \\
\hline
\multirow{3}{*}{65\% (1,365)} & SBTAYLOR-KAN & 58.90 & 58.69 & 58.57 & 57.75 \\
 & SBRBF-KAN & 60.07 & 60.82 & 60.25 & 60.14 \\
 & SBWAVELET-KAN & 55.14 & 39.16 & 40.66 & 39.33 \\
\hline
\multirow{3}{*}{60\% (1,260)} & SBTAYLOR-KAN & 58.40 & 60.08 & 59.61 & 58.79 \\
 & SBRBF-KAN & 57.57 & 55.63 & 55.61 & 55.50 \\
 & SBWAVELET-KAN & 53.61 & 31.16 & 30.33 & 30.00 \\
\hline
\multirow{3}{*}{55\% (1,155)} & SBTAYLOR-KAN & 55.57 & 56.65 & 56.23 & 56.05 \\
 & SBRBF-KAN & 58.57 & 56.22 & 55.21 & 54.99 \\
 & SBWAVELET-KAN & 55.80 & 37.33 & 40.00 & 38.16 \\
\hline
\multirow{3}{*}{50\% (1,050)} & SBTAYLOR-KAN & 56.41 & 53.67 & 53.58 & 53.27 \\
 & SBRBF-KAN & 55.91 & 52.07 & 52.30 & 51.24 \\
 & SBWAVELET-KAN & 50.98 & 62.00 & 37.00 & 38.00 \\
\hline
\multirow{3}{*}{45\% (945)} & SBTAYLOR-KAN & 52.58 & 55.23 & 55.05 & 53.92 \\
 & SBRBF-KAN & 59.23 & 51.06 & 51.67 & 51.08 \\
 & SBWAVELET-KAN & 52.08 & 39.16 & 39.50 & 38.83 \\
\hline
\multirow{3}{*}{40\% (840)} & SBTAYLOR-KAN & 48.75 & 53.51 & 51.36 & 49.46 \\
 & SBRBF-KAN & 54.74 & 54.22 & 54.30 & 54.01 \\
 & SBWAVELET-KAN & 54.49 & 33.50 & 33.33 & 32.16 \\
\hline
\multirow{3}{*}{35\% (735)} & SBTAYLOR-KAN & 49.58 & 48.10 & 48.57 & 45.89 \\
 & SBRBF-KAN & 51.75 & 53.84 & 53.28 & 52.86 \\
 & SBWAVELET-KAN & 56.02 & 36.67 & 35.83 & 35.16 \\
\hline
\multirow{3}{*}{30\% (630)} & SBTAYLOR-KAN & 48.92 & 52.96 & 50.02 & 49.92 \\
 & SBRBF-KAN & 46.76 & 45.79 & 46.99 & 45.33 \\
 & SBWAVELET-KAN & 47.92 & 25.50 & 31.00 & 28.00 \\
\hline
\multirow{3}{*}{25\% (525)} & SBTAYLOR-KAN & 50.75 & 52.53 & 51.70 & 50.99 \\
 & SBRBF-KAN & 48.09 & 46.91 & 47.60 & 46.65 \\
 & SBWAVELET-KAN & 45.08 & 25.00 & 28.33 & 26.00 \\
\hline
\multirow{3}{*}{20\% (420)} & SBTAYLOR-KAN & 48.59 & 51.22 & 48.65 & 49.04 \\
 & SBRBF-KAN & 47.59 & 42.55 & 43.69 & 42.83 \\
 & SBWAVELET-KAN & 46.61 & 28.00 & 30.16 & 27.00 \\
\hline
\end{tabular}
}
\end{table*}

\end{document}